\documentclass{article}

\usepackage[square,numbers]{natbib}
\usepackage[preprint]{neurips_2025}

\usepackage[utf8]{inputenc} 
\usepackage[T1]{fontenc}    
\usepackage{hyperref}       
\usepackage{url}            
\usepackage{booktabs}       
\usepackage{amsfonts}       
\usepackage{nicefrac}       
\usepackage{microtype}      
\usepackage{xcolor}         
\usepackage{graphicx}
\usepackage{multirow}

\usepackage[ruled,vlined]{algorithm2e}
\SetKwFor{ForEach}{for each}{do}{endfor}
\usepackage{amsmath} 
\usepackage{tabularx}
\usepackage{makecell}
\usepackage{ragged2e}
\usepackage{utfsym}
\usepackage{pifont}
\usepackage[table]{xcolor}
\definecolor{myred}{rgb}{0.996,0.578,0.574}
\definecolor{myyellow}{rgb}{0.988,0.961,0.898}
\definecolor{mylightred}{rgb}{0.992,0.887,0.883}

\title{Towards Better Dental AI: A Multimodal Benchmark and Instruction Dataset for Panoramic X-ray Analysis}

\author{%
    Jing Hao\textsuperscript{1}$^\ast$$^\spadesuit$ \quad
    Yuxuan Fan\textsuperscript{2}$^\ast$\quad
    Yanpeng Sun\textsuperscript{3}\quad
    Kaixin Guo\textsuperscript{1}\quad
    \textbf{Lizhuo Lin}\textsuperscript{\textbf{1}}\quad 
    \textbf{Jinrong Yang}\textsuperscript{\textbf{4,5}}\quad \\
    \textbf{Qi Yong H. Ai}\textsuperscript{\textbf{6}}\quad 
    \textbf{Lun M. Wong}\textsuperscript{\textbf{7}}\quad 
    \textbf{Hao Tang}\textsuperscript{\textbf{8}}$^\dagger$\quad 
    \textbf{Kuo Feng Hung}\textsuperscript{\textbf{1}}$^\dagger$\quad \\
    \textsuperscript{1}Faculty of Dentistry, The University of Hong Kong\quad \\
    \textsuperscript{2}The Hong Kong University of Science and Technology (GZ)\quad \\
    \textsuperscript{3}National University of Singapore\quad 
    \textsuperscript{4}CVTE\quad
    \textsuperscript{5}Sun Yat-sen University\quad \\
    \textsuperscript{6}Department of Diagnostic Radiology, The University of Hong Kong \quad \\
    \textsuperscript{7}Imaging and Interventional Radiology, Faculty of Medicine, The Chinese University of Hong Kong \quad \\
    \textsuperscript{8}School of Computer Science, Peking University 
}

\begin{document}

\maketitle
\addtocontents{toc}{\protect\setcounter{tocdepth}{-1}}
\begingroup
\renewcommand\thefootnote{}
\footnotetext{$^\ast$ Equal Contribution.}
\footnotetext{$^\spadesuit$ Project Leader.}
\footnotetext{$^\dagger$ Corresponding Authors: hungkfg@hku.hk, haotang@pku.edu.cn}
\addtocounter{footnote}{-2} 
\endgroup

\begin{abstract}
Recent advances in large vision-language models (LVLMs) have demonstrated strong performance on general-purpose medical tasks. However, their effectiveness in specialized domains such as dentistry remains underexplored. In particular, panoramic X-rays, a widely used imaging modality in oral radiology, pose interpretative challenges due to dense anatomical structures and subtle pathological cues, which are not captured by existing medical benchmarks or instruction datasets. 
To this end, we introduce \textbf{MMOral}, the first large-scale multimodal instruction dataset and benchmark tailored for panoramic X-ray interpretation. MMOral consists of 20,563 annotated images paired with 1.3 million instruction-following instances across diverse task types, including attribute extraction, report generation, visual question answering, and image-grounded dialogue. In addition, we present \textbf{MMOral-Bench}, a comprehensive evaluation suite covering five key diagnostic dimensions in dentistry.
We evaluate 64 LVLMs on MMOral-Bench and find that even the best-performing model, \emph{i.e.}, GPT-4o, only achieves 41.45\% accuracy, revealing significant limitations of current models in this domain. 
To promote the progress of this specific domain, we also propose \textbf{OralGPT}, which conducts supervised fine-tuning (SFT) upon Qwen2.5-VL-7B with our meticulously curated MMOral instruction dataset.
Remarkably, a single epoch of SFT yields substantial performance enhancements for LVLMs, \emph{e.g.}, OralGPT demonstrates a 24.73\% improvement.
Both MMOral and OralGPT hold significant potential as a critical foundation for intelligent dentistry and enable more clinically impactful multimodal AI systems in the dental field. 
The dataset, model, benchmark, and evaluation suite are available at \href{https://github.com/isbrycee/OralGPT/}{https://github.com/isbrycee/OralGPT.}

\end{abstract}

\section{Introduction}
Recent advancements in large vision-language models (LVLMs)~\cite{li2024llava,liu2024llavanext, Qwen-VL,abdin2024phi3,abdin2024phi4,lu2024ovis,wang2024emu3,team2025kimi,wang2024cogvlm,dong2023dreamllm,wei2024vary,zhao2023chatspot,yu2024merlin,ye2024mplug,ye2024mplugowl2,lu2024deepseek,young2024yi,deitke2024molmo} have driven significant progress in open-world medical image understanding, supported by benchmarks such as GMAI-MMBench~\cite{ye2024gmai}, RadBench~\cite{wright2016radbench}, and OmniMedVQA~\cite{hu2024omnimedvqa}, as well as models like LLaVA-Med~\cite{li2023llavamed}, HuatuoGPT-Vision~\cite{chen2024huatuogpt}, MedDr~\cite{he2024meddr}, HealthGPT~\cite{lin2025healthgpt}, and so on~\cite{pan2025medvlm,guo2025llava,moor2023med,li2024gmai_model}. These efforts focus on broad, general-purpose medical scenarios, aiming to evaluate and improve LVLMs across diverse modalities and tasks. However, these general-purpose benchmarks overlook the unique requirements of domain-specific medical fields. In particular, oral radiology—a critical specialty relying on dental imaging for diagnosis and treatment planning—remains largely absent from existing medical benchmarks. The panoramic X-ray is one of the most commonly used imaging modalities and has been widely accepted as a primary source of information for assessing oral health~\cite{turosz2023applications,de2015panoramic}. It provides a comprehensive visualization of all teeth and surrounding structures in a single image, enabling a basic evaluation of dentition, periodontal bone loss, and lesions within the jawbones. The omission of this modality leaves a significant gap: the lack of tailored evaluation and instruction resources hampers the adaptation of LVLMs to dentistry-specific tasks.

Unlike other modalities, interpreting panoramic X-rays presents unique challenges, characterized by dense anatomical structures and fine-grained pathological cues. Addressing these challenges requires not only dental-specific instruction data but also a specialized benchmark aligned with the clinical knowledge of dental practitioners. To bridge this gap, we introduce \textbf{MMOral}, the first large-scale multimodal instruction dataset and benchmark tailored for panoramic X-ray understanding. MMOral comprises 20,563 annotated panoramic X-rays paired with 1.3 million instruction-following instances, spanning multiple task formats including attribute extraction, report generation, visual question answering, and image-grounded dialogue. Complementing the dataset, \textbf{MMOral-Bench} offers a curated evaluation suite covering five key diagnostic dimensions, including the condition of teeth, pathological findings, historical treatments, jawbone observations, and clinical summary \& recommendations. 
This benchmark consists of 100 images, paired with 500 closed-ended and 600 open-ended questions.
All cases in MMOral-Bench are manually chosen and checked from the MMOral to ensure their quality and reliability. Together, MMOral and MMOral-Bench lay a critical foundation for advancing intelligent dentistry and enabling clinically meaningful multimodal AI.

\begin{figure}[t!]
  \centering
  \includegraphics[width=\textwidth]{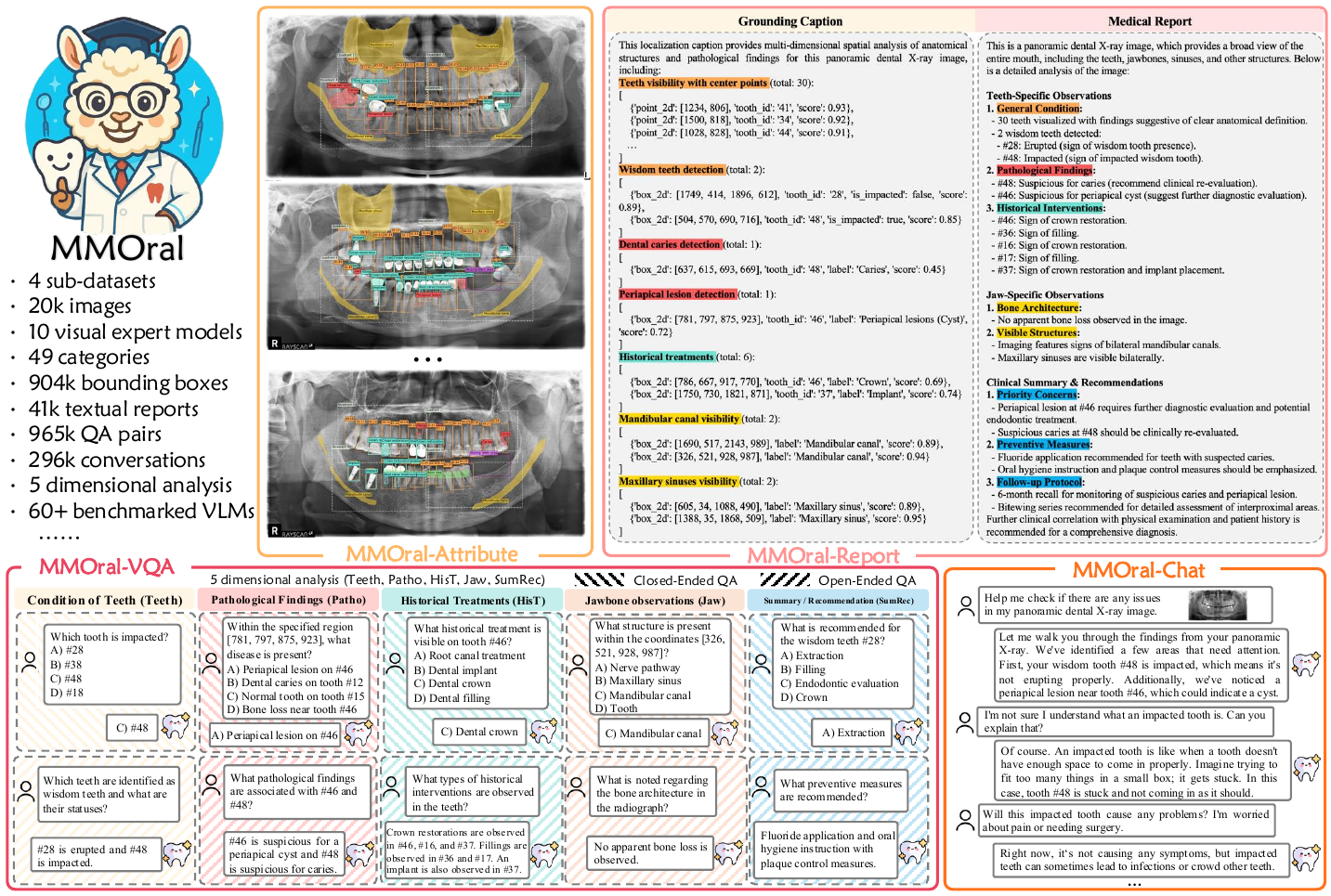}
  \vspace{-0.6cm}
  \caption{Overview of the MMOral. It consists of four sub-datasets: MMOral-Attribute, MMOral-Report, MMOral-VQA, and MMOral-Chat. MMOral-Attribute contains a total of 49 categories of  anatomical structures within panoramic X-rays. MMOral-Report consists of two types of textual descriptions: the grounding caption and the medical report. MMOral-VQA includes closed-ended and open-ended QA pairs spanning five diagnostic dimensions. MMOral-Chat simulates the dialogue process between patients and radiology experts regarding the interpretation of panoramic X-rays.}
  \vspace{-0.48cm}
  \label{fig:mmoral-dataset-overview}
\end{figure}
\footnotetext[1]{https://platform.stepfun.com/}
\footnotetext[2]{https://www.volcengine.com/product/doubao/} 
We assess 53 publicly available LVLMs (44 general-purpose and 9 medical-specific models) as well as 11 advanced proprietary LVLMs such as GPT-4o~\cite{hurst2024gpt}, GPT-4V~\cite{hurst2024gpt}, Claude-3-7-Sonnet~\cite{Claude3S}, Gemini-2.0-Flash~\cite{team2023gemini}, Gemini-2.5-Flash~\cite{team2023gemini}, Qwen-Max-VL~\cite{Qwen-VL}, Step-1o series \footnotemark[1], and Doubao-1.5 series \footnotemark[2] on our Oral-Bench. We summarize five findings according to the evaluation outcomes: \textbf{(1)} MMOral-Bench is a significant challenge for current LVLMs, even for GPT-4o, which achieves only a 41.45\% average score. \textbf{(2)} The performance of existing universal medical LVLMs is suboptimal in the field of dental applications. \textbf{(3)} Existing medical LVLMs show no clear advantage over general-purpose LVLMs for dentistry tasks. \textbf{(4)} Nearly all LVLMs perform worse on open-ended questions compared to closed-ended questions in the MMOral-Bench. \textbf{(5)} LVLMs exhibit a clear performance bias across five diagnostic dimensions (depending on the focus of anatomical structures), and they show relatively limited capability when it comes to fine-grained teeth-related questions.

To further promote the progress of intelligent dentistry, we propose \textbf{OralGPT}, which conducts extensive supervised fine-tuning (SFT) experiments on the Qwen2.5-VL-7B model~\cite{Qwen-VL}. Experimental results show that the average score of OralGPT on MMOral-Bench could improve by 24.73\% when conducting SFT utilizing MMOral instruction data (MMOral-Report, MMOral-VQA, MMOral-Chat) for one epoch. This obvious improvement highlights the value of the MMOral towards intelligent dentistry. 

To summarize, our contributions are threefold: \textbf{(1)} We introduce MMOral, the first large-scale multimodal instruction dataset and benchmark tailored for panoramic X-ray understanding. MMOral-Bench offers a curated evaluation suite covering five key diagnostic dimensions, which could comprehensively reflect the capabilities of current LVLMs in the dental field. \textbf{(2)} 64 existing representative LVLMs are assessed on MMOral-Bench, including 11 proprietary models and 53 open-source models (44 general-purpose models and 9 medical-specific models). Evaluation results pave the way for the next optimization direction to enhance the interpretation of panoramic X-rays. \textbf{(3)} We propose OralGPT, which implements supervised fine-tuning using our MMOral instruction data to enhance the capability of panoramic X-ray analysis. Experiments demonstrate the average score of OralGPT on MMOral-Bench could improve by 24.73\% when conducting SFT for only one epoch.

\begin{figure}[t!]
  \centering
  \includegraphics[width=1\linewidth]{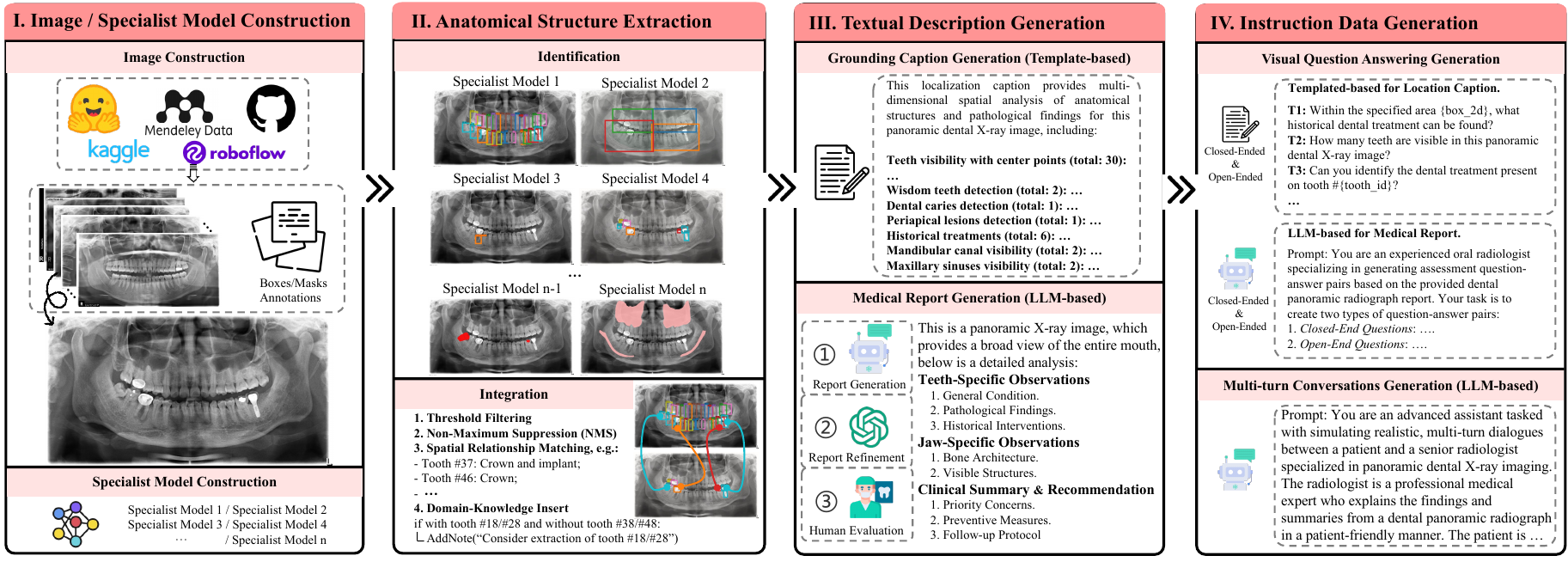}
  \vspace{-0.6cm}
  \caption{The MMOral dataset curation pipeline, which consists of four sequential steps.}
  \vspace{-0.5cm}
  \label{fig:pipeline}
\end{figure}

\section{MMOral Dataset Curation}
The MMOral dataset curation pipeline consists of four sequential steps, which are shown in Figure~\ref{fig:pipeline}.
\subsection{Images and Visual Specialists Construction}
We collect panoramic X-ray images from two publicly available datasets: the TED3 dataset~\cite{ted3} and the dataset proposed by Hoang Viet Do~\cite{do2024dataset}. After filtering out duplicate images, a total of 20,563 images is obtained. 

Subsequently, we build ten visual specialist models to simulate the interpretative process of oral radiology experts. These models are designed to recognize 49 categories of anatomical structures in radiographic images, encompassing visual elements such as tooth numbering (1 to 32 according to the FDI system~\cite{wong2015fdi}), four quadrants, pathological findings, historical treatments, bone loss, and visible bone structures. The category spaces of multiple visual specialist models overlap with each other, ensuring the reliability of anatomical structure extraction. Additionally, we utilize an off-the-shelf OpenOCR model~\cite{Du2024SVTRv2} to detect the acquisition time displayed within the images. \textit{More details of image source information, visual specialist model construction, and the specific category list of detected anatomical structures can be found in the supplementary materials.}

\subsection{Anatomical Structure Extraction}
After obtaining the extracted anatomical structures from all visual specialist models, we further integrate these outcomes for two primary considerations: \textbf{(1) the necessity to deduplicate overlapping categories detected by multiple visual specialist models}, and \textbf{(2) the discrete nature of identified anatomical structures lacking interrelational information.} 
First, we meticulously design the anatomical structures post-processing pipeline to remove redundant information.
Second, since the associations between dental pathological findings, historical treatment, and their corresponding tooth numbering systems remain undefined, we establish these correlations through their spatial relationships.
Third, domain-specific clinical knowledge is inserted. For instance, extraction of maxillary third molars (teeth \#18/28) is recommended when they lack opposing mandibular counterparts (teeth \#48/38). \textit{The developed anatomical structure post-processing and relationship matching pipeline comprises eight systematic steps, as detailed in the supplementary materials.}

\subsection{Report Generation}
We synthesize the discrete anatomical structure information into two coherent textual outputs: a \textbf{grounding caption} and a \textbf{medical report}. The grounding caption contains detailed coordinates, categories, and confidence scores of all anatomical structures, providing a multi-dimensional spatial observation and identification for panoramic X-rays. All anatomical information is systematically organized into structured textual descriptions following manually designed rules. An illustrative example of the grounding caption is shown in the left panel of the MMOral-Report part in Figure~\ref{fig:mmoral-dataset-overview}.

In contrast to grounding captions, medical reports focus on providing a condensed summary of key anatomical structures, abnormal findings, and corresponding diagnostic recommendations. Through extensive consultations with senior dental specialists, we structure the medical report into three principal sections: \textbf{Teeth-Specific Observations}, \textbf{Jaw-Specific Observations}, and \textbf{Clinical Summary \& Recommendations}. The medical report is derived from two-stage LLM-based generation. First, we prompt DeepSeek-R1-Distill-Llama-70B\footnotemark[3] to generate medical reports based on grounding captions. Following that, we manually check the generated medical reports and summarize several common errors. 
According to these errors, we carefully prepare the prompt for report correction and instruct GPT-4-turbo~\cite{achiam2023gpt4_turbo} to simultaneously output both revised reports and corresponding revision logs. By examining these revision logs, we can efficiently identify modified sections of the reports, thereby facilitating quality verification of the revised content. To comprehensively evaluate the quality of the revised reports, two professional dentists are engaged to conduct the human evaluation, which will be discussed in section~\ref{sec:3.2}. \textit{The details of prompts in two-stage generation are provided in the supplementary materials.}
\footnotetext[3]{{https://huggingface.co/deepseek-ai/DeepSeek-R1-Distill-Llama-70B}}

\subsection{Instruction Data Generation}
\label{sec:2.4}
Based on the generated grounding captions and medical reports, we construct two types of single-turn instruction-following QA pairs—closed-ended and open-ended—using template-based and LLM-based approaches. For grounding captions, we generate both closed-ended multiple-choice QA pairs and open-ended QA pairs using manually designed question templates. The incorrect choices in the closed-ended QA are created by introducing random perturbations to the ground truth for enhancing the model's ability to recognize and understand subtle differences. For the medical report, we prompt GPT-4-turbo (see \textit{supplementary materials} for prompt designs) to simultaneously generate both closed-ended and open-ended QA pairs.
To systematically evaluate panoramic X-ray analysis capabilities, we establish a taxonomy across five clinically grounded dimensions: \textbf{condition of teeth (Teeth)}, \textbf{pathological findings (Patho)}, \textbf{historical treatments (HisT)}, \textbf{jawbone observations (Jaw)}, and \textbf{clinical summary \& recommendation (SumRec)}. Each QA pair is mapped to one or more of these diagnostic categories based on its clinical intent, forming a multi-dimensional analysis.

In addition to the single-turn QA pairs mentioned above, we also prompt GPT-4-turbo to generate a multi-turn conversation between the assistant and a person asking questions about the panoramic X-ray. The answers are in a tone as if the assistant is seeing the panoramic radiograph and explaining the findings and summaries in a patient-friendly manner. A diverse set of questions is asked about the visual content of the image and the assistant's explanations, and only questions that have definite answers are considered. \textit{Please see supplementary materials for the detailed prompt.}

\section{MMOral Dataset Analysis}
\subsection{Data Statistics}
\begin{figure}[h!]
  \centering
  \includegraphics[width=\textwidth]{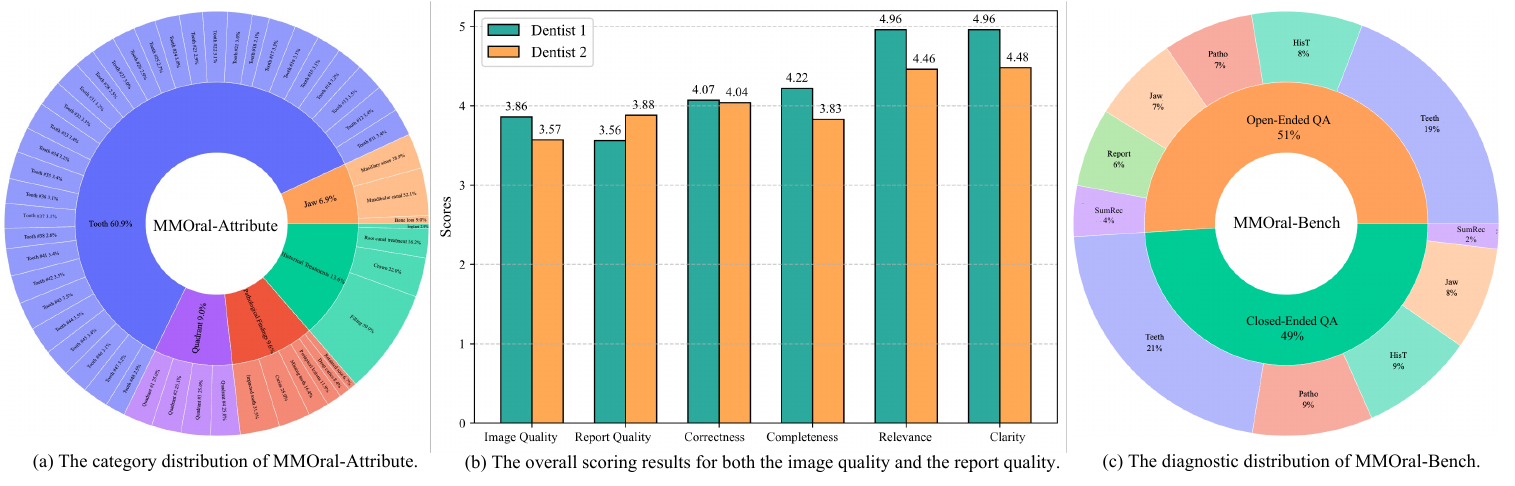}
  \vspace{-0.6cm}
  \caption{The data statistic distribution and human evaluation results.
  }
  \vspace{-0.45cm}
  \label{fig:cate_dist_human_eval}
\end{figure}

The MMOral comprises 20,563 images paired with 1.3 million instruction-following data instances, establishing a comprehensive multimodal resource for human-AI interaction research in the digital dentistry field. It consists of four distinct sub-datasets: MMOral-Attribute, MMOral-Report, MMOral-VQA, and MMOral-Chat. Each sub-dataset corresponds to specific tasks, including visual perception, report generation, visual question answering, and image-grounded dialogue. Table~\ref{tab:MMOral_description} provides detailed information on each component and its corresponding data size. Notably, a single panoramic X-ray contains an average of 44 bounding boxes, reflecting its structural diversity and dense representations, making its interpretation highly complex.
Figure~\ref{fig:cate_dist_human_eval} (a) presents the category distribution of MMOral-Attribute, comprising 5 major categories and 49 subcategories of anatomical structures identified in panoramic X-rays, illustrating its diversity and comprehensive coverage. 
To the best of our knowledge, MMOral is the largest multimodal dataset for panoramic X-rays to date, forming a robust foundation for the development and evaluation of LVLMs.

\begin{table}[]
\caption{The brief description of four sub-datasets in MMOral and their corresponding data size.}
\vspace{-0.2cm}
\label{tab:MMOral_description}
\centering
\resizebox{0.9\linewidth}{!}{
\begin{tabular}{l|c|c|c}
\toprule
Dataset                 & Sub-Dataset      & Description   & Size    \\ 
\midrule
\multirow{8}{*}{\centering MMOral} 
                        & MMOral-Attribute & \makecell[l]{The category, position and correlation of anatomical \\ structures shown in the panoramic X-ray image.} 
                                                                  & 904k  \\ 
                        \cmidrule(lr){2-4}  
                        & MMOral-Report    & \makecell[l]{Two types of textual description for each panoramic \\ X-ray image: grounding caption and medical report.} 
                                                                  & 41k   \\ 
                        \cmidrule(lr){2-4}  
                        & MMOral-VQA       & \makecell[l]{Two types of visual question answering: closed-ended \\ QA and open-ended QA.} 
                                                                  & 965k  \\ 
                        \cmidrule(lr){2-4} 
                        & MMOral-Chat      & \makecell[l]{The multi-turn conversation between the assistant and \\ a person asking questions about the panoramic X-ray.} 
                                                                  & 296k  \\ 
\bottomrule
\end{tabular}
}
\vspace{-0.4cm}
\end{table}

\subsection{Strategies for Ensuring Data Quality}
\label{sec:3.2}
We ensure the quality of the MMOral from the following three perspectives:

\textbf{(1) Collaborative validation of anatomical structures by multiple specialist models.}
The anatomical structures present in panoramic X-rays serve as metadata for subsequent report generation and instruction data construction. If these visual structures contain errors, such inaccuracies will propagate throughout all datasets. To ensure precise detection of anatomical structures, we construct ten visual specialist models with overlapping category spaces. For instance, ten structures (e.g., caries, periapical lesion, impacted teeth, missing teeth area, filling, implant, root canal treatment, crown, mandibular canal, and maxillary sinus) are validated by two or more visual specialists, with the final results obtained through post-processing. This approach ensures the reliability of the extraction of anatomical structures, providing trustworthy metadata for subsequent medical report generation and instruction data construction.

\textbf{(2) Two-stage LLM-based report generation.}
Unlike most existing approaches that rely on a single LLM to generate image captions, we adopt a two-stage scheme of generation followed by correction to ensure report quality. Due to the complexity of generating structured medical reports based on discrete anatomical structures, LLMs inevitably introduce errors in areas such as structural organization, content description, and hallucinations. By manually reviewing the preliminary reports generated by DeepSeek-R1-Distill-Llama-70B, we identify common errors and summarize them into 8 rules for prompting GPT-4-turbo to revise the reports. Analysis of the revision logs revealed that \textbf{95.45\%} of the reports are successfully corrected, significantly improving their overall quality.

\textbf{(3) Professional dentist evaluation.}
We invite two professional dentists to evaluate the medical report and assess whether the associated panoramic X-ray image meets practical imaging quality standards. Both the image quality and the report quality are evaluated using a scoring system with five levels, ranging from 1 to 5, representing a progression from "very dissatisfied" to "very satisfied." For the report evaluation, in addition to assigning an overall score, the evaluators are required to provide further scores in four specific aspects: correctness, completeness, relevance, and clarity. 
Figure~\ref{fig:cate_dist_human_eval} (b) illustrates the overall scoring results for both the image quality and the report quality, showing that the average scores for all aspects exceeded 3.5, highlighting the high quality of MMOral. 
\textit{The evaluation guidelines are detailed in the supplementary materials.}

\section{MMOral-Bench}
\subsection{Benchmark Construction}

We construct MMOral-Bench by curating 500 closed-ended and 600 open-ended QA pairs with 100 images \textbf{through significant manual selection and validation}. To ensure image quality, we select images from the dataset proposed by Hoang Viet Do~\cite{do2024dataset} because its acquisition process is clearer and more reliable. Moreover, we filter out QA pairs that could not be answered with the image, and incorrect answers are identified and re-annotated. MMOral-Bench covers five different clinically grounded dimensions (e.g., Teeth, Patho, HisT, Jaw, SumRec) and thus can comprehensively evaluate the ability of LVLMs to understand and interpret panoramic X-rays. Each QA pair is assigned to one or more diagnostic dimensions based on its clinical intent, enabling a multi-dimensional analysis. 
The distribution of the diagnostic dimension on MMOral-Bench can be seen in Figure~\ref{fig:cate_dist_human_eval}(c), and some QA examples are present in the MMOral-VQA part of Figure~\ref{fig:mmoral-dataset-overview}.

\subsection{Evaluation Metrics}
We adopt two evaluation metrics tailored for closed-ended and open-ended questions, respectively. For closed-ended questions, we use accuracy as the evaluation metric. For open-ended questions, following previous works~\cite{yu2023mmvetv1,yu2024mmvetv2}, we construct a few-shot prompt and leverage GPT-4-turbo to assist with the evaluation. The few-shot prompt incorporates nine in-context examples with open-ended answers, covering fully correct cases, partially correct cases, and incorrect cases. GPT-4-turbo assigns a score ranging from 0 to 1 based on each sample’s input question, ground truth, and model output. We report the evaluation scores for each category as well as the overall performance. \textit{The full details of the few-shot prompt can be found in the supplementary materials.} We integrate the evaluation of MMOral-Bench into the standard VLMEvalKit~\cite{duan2024vlmevalkit} framework, thereby facilitating subsequent capability assessments of newly developed LVLMs\footnotemark[4].
\footnotetext[4]{https://anonymous.4open.science/r/MMOral-2459/README.md}
\section{Experiments}
\subsection{Experimental Setups}
\label{sec:5.1}

\noindent\textbf{Benchmarked LVLMs.} 
We conduct zero-shot evaluations across 64 LVLMs on our MMOral-bench, and we pick 36 out of 64 models for demonstration in the main text; additional results are provided in the \textit{supplementary material}. 
We evaluate 8 proprietary LVLMs via API: GPT-4o~\cite{hurst2024gpt}, GPT-4V \cite{hurst2024gpt}, Claude-3-7-sonnet~\cite{Claude3S}, Gemini-2.5-Flash~\cite{team2023gemini}, Gemini-2.0-Flash~\cite{team2023gemini}, Qwen-Max-VL~\cite{Qwen-VL}, Step-1o-turbo\footnotemark[3], and Doubao-1-5-thinking-vision-pro\footnotemark[4]. For medical-specific LVLMs, we test 5 powerful models, including LLaVA-Med~\cite{li2023llavamed}, HuatuoGPT-V~\cite{chen2024huatuogpt}, MedVLM-R1~\cite{pan2025medvlm}, MedDr~\cite{he2024meddr}, and HealthGPT~\cite{lin2025healthgpt}. We also evaluate 23 representative general-purpose LVLMs: Deepseek-VL-7b~\cite{lu2024deepseek}, Emu3~\cite{wang2024emu3}, Qwen2.5-VL-72B~\cite{bai2025qwen2}, CogVLM2-19B~\cite{wang2024cogvlm}, GLM-4V-9B~\cite{glm2024chatglm}, LLaVA-NeXT-13B-hf~\cite{liu2024llavanext}, LLaVA-OneVision~\cite{li2024llava}, LLaMA-3.2-Vision-11B~\cite{llama3_2}, Cambrian-34B~\cite{tong2024cambrian}, Phi-4-multimodal-instruct~\cite{abdin2024phi4}, InternVL3-38B~\cite{chen2024expanding}, Chameleon-7B~\cite{lu2023chameleon}, PaliGemma-3B~\cite{beyer2024paligemma}, MiniCPM-O2.6~\cite{yao2024minicpm}, Kosmos-2~\cite{peng2023kosmos}, Kimi-VL-A3B-Thinking~\cite{team2025kimi}, Ovis2-34B~\cite{lu2024ovis}, Qwen-QVQ-72B~\cite{qvq-72b-preview}, Gemma3-12B~\cite{team2025gemma}, XComposer2-VL-7B~\cite{zhang2024internlm}, Molmo-72B~\cite{deitke2024molmo}, and Yi-VL-34B~\cite{young2024yi}.

\noindent\textbf{Supervised Fine-Tuning.}
We also implement supervised fine-tuning (SFT) on two popular LVLM models with different scales (Qwen-2.5-VL-7B~\cite{Qwen-VL} and LLaVA-Next-13B~\cite{liu2024llavanext}) using our MMOral instruction data to verify its effectiveness. The extensive experiments are implemented through the LLaMA-Factory framework ~\cite{zheng2024llamafactory} while maintaining default hyperparameters, with all models being trained for a single epoch. The results on Qwen-2.5-VL-7B will be discussed in Sec.~\ref{sec:5.3}, and the results on LLaVA-Next-13B can be found in \textit{supplementary materials}. We refer to the LVLM obtained through supervised fine-tuning of Qwen-2.5-VL-7B as \textbf{OralGPT}, due to its outstanding performance in panoramic X-ray analysis.

\begin{table}[t]
\centering
\caption{Results on MMOral-Bench for existing various LVLMs across both closed-ended and open-ended VQA tasks. The best-performing model in each category is highlighted \textbf{in-bold}, while the second-best is \underline{underlined}. 36 out of 64 models for demonstration in the main text; additional results are provided in the \textit{supplementary material}. 
\vspace{-0.2cm}
}
\label{tab:evaluation_results}
\resizebox{\textwidth}{!}{%
\begin{tabular}{@{}l|cccccc|ccccccc|c@{}}
\toprule
\multirow{2}{*}{\textbf{Model}}
  & \multicolumn{6}{c|}{\textbf{Close‐ended VQA}}
  & \multicolumn{7}{c|}{\textbf{Open‐ended VQA}}  
  & \multirow{2}{*}{\textbf{Avg.}} \\
\cline{2-14} 
\rule{0pt}{2.5ex} 
  & \textbf{Teeth} & \textbf{Patho} & \textbf{His} & \textbf{Jaw} & \textbf{Summ} & \textbf{Overall}
  & \textbf{Teeth} & \textbf{Patho} & \textbf{His} & \textbf{Jaw} & \textbf{Summ} & \textbf{Report} & \textbf{Overall} \\
\hline
\rowcolor{mylightred} \multicolumn{15}{l}{\textit{Proprietary LVLMs}} \\ 
\hline
GPT-4o-2024-11-20~\cite{hurst2024gpt}                  & 39.65 & 40.99 & 46.71 & 55.81 & 56.25 & 45.40 & 31.48 & \underline{26.05} & 37.56 & {57.42} & {30.37} & 42.50 & 37.50 & \underline{41.45} \\
GPT-4V~\cite{hurst2024gpt}                  & 37.88 & 39.13 & 48.50 & 51.69 & 58.33 & 43.40 & 27.76 & 13.47 & 33.50 & \underline{58.95} & {30.84} & 45.00 & 34.83 & 39.12 \\
Claude-3-7-Sonnet-20250219~\cite{Claude3S} & 41.24 & 37.27 & 44.31 & 39.70 & 45.83 & 41.40 & 36.93 & 26.65 & 42.39 & {51.09} & {28.04} & 50.00 & 40.67 & 41.04\\
Gemini-2.5-Flash-preview-04-17~\cite{team2023gemini}     & 24.60 & 16.15 & 27.55 & 20.60 & 10.42 & 22.00 & 35.99 & 22.76 & \underline{40.61} & {51.53} & {32.71} & 45.50 & 39.08 & 30.54 \\
Gemini-2.0-Flash~\cite{team2023gemini}         & 37.17 & 35.40 & 44.31 & 46.82 & 58.33 & 41.20 & \underline{37.27} & 26.05 & {40.36} & 52.40 & {35.05} & \underline{49.00} & \underline{40.67} & {40.94} \\
Qwen-Max-VL-2025-04-08~\cite{Qwen-VL} & 18.41 & 11.18 & 27.55 & 32.96 & 47.92 & 22.00 & 10.22 & 7.30 & 11.12 & 22.88 & 6.86 & 27.00 & 14.33 & 18.15 \\
Step-1o-turbo \footnotemark & 31.86 & 24.22 & 38.92 & 55.81 & 41.67 & 36.00 &33.02&21.56&30.20&51.31&31.31&45.00&36.00&36.00 \\
Doubao-1-5-thinking-vision-pro-250428 \footnotemark & 26.20 & 27.33 & 23.35 & 19.85 & 31.25 & 24.80 & 34.38 & 25.45 & 39.09 & 56.11 & \underline{39.72} & 49.00 & 40.33 & 32.57 \\ \hline 
\rowcolor{mylightred} \multicolumn{15}{l}{\textit{Open-Source LVLMs}} \\ \hline
Deepseek-VL-7b-chat~\cite{lu2024deepseek}      & 22.65 & 17.39 & 28.74 & 59.93 & 52.08 & 31.20 & 12.75 &  8.16 &  8.40 & 30.00 & 13.14 &  9.10 & 13.42 & 22.31\\
Emu3-chat~\cite{wang2024emu3} & 40.89 & 44.72 & 37.73 & 60.67 & 43.75 & 45.8 & 18.02 &  7.02 &  15.50 & 28.53 & 12.44 &  9.60 & 16.05 & 30.93\\
Qwen2.5-VL-72B~\cite{bai2025qwen2}           & 26.55 & 27.95 & 26.35 & 22.47 & 47.92 & 26.80 & 13.05 & 18.44 & 11.66 & 26.88 &  7.44 & 11.50 & 14.77 & 20.79\\
CogVLM2-19B~\cite{wang2024cogvlm}                & 33.63 & 31.68 & 34.13 & 38.95 & 60.42 & 35.20 & 26.11 & 17.09 & 26.86 & 49.24 & 18.14 & 24.50 & 27.63 &31.42 \\
GLM-4V-9B~\cite{glm2024chatglm}                & 29.03 & 35.40 & 41.32 & 62.55 & 64.58 & 40.20 & 17.85 &  8.01 & 17.46 & 24.12 & 15.93 & 19.40 & 17.50 & 28.85\\
LLaVA-NeXT-13B-hf~\cite{liu2024llavanext}         & 30.09 & 32.92 & 30.54 & 38.20 & 60.42 & 33.80 & 14.48 &  10.28 &  9.23 & 22.41 & 14.30 & 21.30 & 15.43 & 24.62\\
LLaVA-OneVision~\cite{li2024llava}          & 14.51 & 18.01 & 35.33 & 42.70 & 31.25 & 24.40 & 22.68 & 13.48 & 17.75 & 38.35 & 18.72 & 11.20 & 20.93 & 22.67\\
LLaMA-3.2-Vision-11B-Instruct~\cite{llama3_2}     &27.96&21.12&40.72&50.94&50.00&34.00     & 24.81 & 20.71 & 25.03 & 33.65 & 17.56 & 19.50 & 23.97 & 28.99\\
Cambrian-34B~\cite{tong2024cambrian}             & 34.87 & 34.16 & 44.31 & 70.04 & 60.42 & 44.40 & 33.10 & 21.42 & 31.83 & 48.24 & 13.60 & 16.00 & 29.63 & 37.02\\
Phi-4-multimodal-instruct~\cite{abdin2024phi4}  &36.28 &36.65 &49.10 &60.30 &54.17 &43.60 &25.52 &20.14 &27.69 &43.29 &13.84 &12.80 &24.57 &34.09 \\
InternVL3-38B~\cite{chen2024expanding}         & 28.67	&21.12	&25.75	&39.33	&31.25	&28.40 &33.69	&22.41	&29.70	&46.11	&20.23	&42.90	&34.15 & 31.28\\
Chameleon-7B~\cite{lu2023chameleon}             & 32.57 & 44.10 & 37.13 & 29.59 & 52.08 & 35.80 &  6.02 &  6.10 &  9.35 &  9.71 &  5.35 &  8.40 &  7.27 & 21.54\\
PaliGemma-3B~\cite{beyer2024paligemma}             & 26.02 & 24.22 & 42.52 & 47.57 & 35.42 & 33.20 &  8.20 &  9.65 &  8.70 & 13.29 &  6.16 &  0.60 &  7.78 & 20.49\\
MiniCPM-O$2.6$~\cite{yao2024minicpm}           & 30.27 & 23.60 & 24.55 & 36.33 & 14.58 & 30.27 & 29.20 & 17.38 & 24.38 & 49.76 & 15.93 & 27.90 & 28.42 & 28.21\\
Kosmos-2~\cite{peng2023kosmos}                 & 15.75 & 18.01 & 28.14 & 10.11 & 25.00 & 17.40 & 13.58 & 10.71 & 11.18 & 19.76 &  8.49 &  3.40 & 11.87 & 14.64\\
mPLUG-Owl3-7B~\cite{ye2024mplug}            & 34.16 & 32.30 & 36.53 & 71.91 & 62.50 & 42.80 & 12.50 &  8.44 &  8.52 & 30.59 &  3.26 & 13.90 & 13.67 & 28.24\\
Gemma3-12B~\cite{team2025gemma}               & 24.78 & 19.88 & 31.74 & 34.08 & 29.17 & 26.60 & 25.21 & 20.00 & 20.65 & 26.88 & 22.33 & 33.20 & 25.32 & 25.96\\
XComposer2-VL-7B~\cite{zhang2024internlm}         & 25.49 & 26.71 & 21.56 & 13.86 & 45.83 & 23.20 &  6.52 & 11.01 & 15.00 &  7.67 &  2.10 &  8.53 &  8.99 & 16.10\\
Molmo-72B-0924~\cite{deitke2024molmo}           & 28.85 & 14.91 & 29.94 & 26.59 & 22.92 & 25.60 &  9.25 &  6.31 &  3.49 & 12.65 &  5.00 &  9.20 &  8.23 & 16.92\\
Yi-VL-34B~\cite{young2024yi} &36.81 &36.64 &43.11 &41.20 &70.83 &40.20 &24.97 &23.40 &20.59 &39.35 &15.23 &9.90 &22.98 &31.59 \\
Qwen-QVQ-72B~\cite{qvq-72b-preview} & \textbf{48.67} & \underline{49.07} & \textbf{59.28} & \underline{74.53} & 72.92 & \underline{56.60} &20.37&12.28&16.75&41.05&22.90&24.5&22.75&39.68 \\ 
Ovis2-34B~\cite{lu2024ovis}                & \underline{45.84} & \textbf{51.55} & \underline{53.89} & \textbf{79.40} & \textbf{79.17} & \textbf{56.80} & 32.48 & 24.33 & 31.60 & 50.88 & 21.05 & 31.70 & 33.02 & \textbf{44.91}\\
Kimi-VL-A3B-Thinking~\cite{team2025kimi} & 25.84	& 27.33	& 25.75	& 29.96	& 27.08	& 26.80 & \textbf{52.53}	& \textbf{37.66}	& \textbf{53.79}	& \textbf{68.59}	& \textbf{50.93}	& \textbf{61.50}	& \textbf{54.55} & 40.68\\
\hline
\rowcolor{mylightred} \multicolumn{15}{l}{\textit{Medical Specific LVLMs}} \\ \hline
LLaVA-Med~\cite{li2023llavamed}                & 25.49 & 26.71 & 21.56 & 13.86 & 45.83 & 23.20 & 23.23 & 18.75 & 11.36 & 32.82 & 26.28 &  5.30 & 19.60 & 21.40\\
HuatuoGPT-V-34B~\cite{chen2024huatuogpt}          & 28.85 & 14.91 & 29.94 & 26.59 & 22.92 & 25.60 & 32.62 & 18.65 & 28.05 & 53.12 & 18.60 & 15.40 & 29.48 & 27.54 \\
HealthGPT-XL32~\cite{lin2025healthgpt} & 39.65 & 44.10 & 51.50 & 76.41 & \underline{79.17} & 52.00 & 29.80 & 22.16  & 24.11 & 47.82 & 24.77 & 10.00 & 27.17 & 39.59 \\
MedVLM-R1~\cite{pan2025medvlm} &28.67 &31.68 &37.72 &65.17 &47.92 &38.60 &22.58 &12.28 &21.57 &40.61 &21.96 &24.50 &24.58 & 31.59\\
MedDr~\cite{he2024meddr} &36.46 &36.02 &41.92 &73.03 &64.58 &46.00 &\underline{27.50} &28.14 &30.20 &49.17 &26.17 &7.50 &26.17 & 36.09\\

\bottomrule
\end{tabular}%
}
\vspace{-0.6cm}
\end{table}

\subsection{Evaluation Results}
 Following a comprehensive review of the evaluation outcomes, which are shown in Table~\ref{tab:evaluation_results}, we have identified 5 key findings regarding the performance of existing LVLMs in the dental domain:

\noindent\textbf{Finding 1. The MMOral-Bench poses significant challenges to ALL LVLMs.} 
 Even the most advanced model, GPT-4o, only achieves 41.45\% overall performance, highlighting persistent challenges and fundamental limitations in current LVLMs' capacity to interpret complex panoramic X-rays, which are characterized by anatomically dense structures and fine-grained pathological patterns. This critical performance gap reveals fundamental limitations of existing LVLMs' capacity in dental-specific images, underscoring substantial room for improvement.

\noindent\textbf{Finding 2. The performance of existing universal medical LVLMs is suboptimal in the field of dental applications.}
Current universal medical LVLMs, which predominantly focus on enhancing capabilities for generalized clinical scenarios across diverse medical imaging modalities, have unsatisfactory performance when it comes to understanding panoramic X-rays—a specialized, fine-grained modality within dental imaging. The results reveal that general medical LVLMs achieve less than 40\% average accuracy on MMOral-Bench, with HealthGPT-XL32~\cite{lin2025healthgpt} attaining peak performance at 39.59\%. This indicates that current universal medical LVLMs still require further exploration and improvement in their ability to interpret panoramic X-rays, which is an imaging modality characterized by complex and numerous anatomical structures.

\noindent\textbf{Finding 3. Existing medical LVLMs show no significant advantage over general LVLMs in the field of dentistry.} 
Existing medical-specific LVLMs, including the LLaVA-Med series, HuatuoGPT series, MedVLM-R1, MedDr, and HealthGPT, fail to outperform general-purpose models in our MMOral-bench. This indicates that current medical LVLMs lack adequate understanding and analytical capabilities specific to the oral region. 
Among all medical LVLMs evaluated, HealthGPT demonstrates the best performance, achieving an average score of 39.59\%. However, this score remains lower than that of general-purpose open-source models such as the Ovis2 series and commercial models like GPT-4o and Claude-3-7-Sonnet. These results highlight the need for further improvements in medical-specific AI models to enhance their understanding of the oral region—an area intrinsically linked to essential human functions such as eating and speaking.

\begin{figure}[!t]
  \centering
  \includegraphics[width=\textwidth]{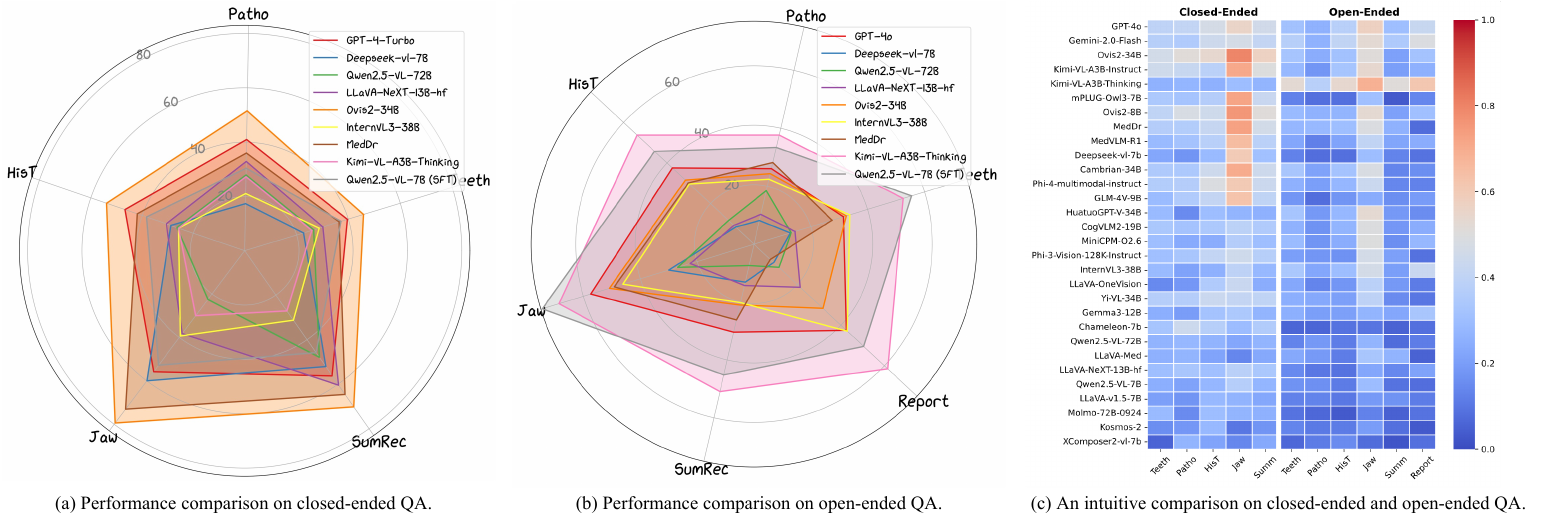}
  \vspace{-0.6cm}
  \caption{Performance comparison on both closed-ended and open-ended QA across multiple LVLMs.}
  \vspace{-0.45cm}
  \label{fig:radar_map}
\end{figure}

\noindent\textbf{Finding 4. Nearly all LVLMs perform worse on open-ended questions compared to closed-ended questions in the MMOral-Bench.} Figure~\ref{fig:radar_map} shows the performance comparison on both closed-ended and open-ended QA tasks. Although some LVLMs (e.g., LLaVA-NeXT-8B-hf, DeepSeek-VL-7B, Ovis2-34B) perform relatively well in closed-ended VQA tasks, they exhibit a significant performance drop in open-ended VQA tasks. Moreover, the proportion of open-sourced models with an overall score below 25\% in open-ended VQA tasks is as high as 62.3\% (33 out of 53 models). This highlights the current limitations of LVLMs in handling open-ended answer generation for dentistry-related questions and the urgent need for targeted optimization. 

\noindent\textbf{Finding 5. LVLMs demonstrate a clear bias across various question categories, depending on the focus of anatomical structures involved.} The questions are categorized into five dimensions based on their focus as mentioned in Sec.~\ref{sec:2.4}: `Teeth', `Patho', `HisT', `Jaw', `SumRec'. We observe that current LVLMs perform relatively well on the `Jaw' category—both in closed-ended and open-ended tasks—where the focus is on larger anatomical structures such as bone loss, mandibular canals, and maxillary sinuses. In contrast, the models generally exhibit poorer performance on categories that require fine-grained visual understanding, such as `Teeth', `HisT', and `Patho'. This suggests that current LVLMs still need significant improvement in their ability to perform fine-grained dental visual understanding and reasoning.

\begin{table}[!t]
\centering
\caption{The effectiveness verification of MMOral instruction data by supervised fine-tuning.}
\vspace{-0.2cm}
\label{tab:sft_results}
\resizebox{\textwidth}{!}{%
\begin{tabular}{@{}l*{17}{c}@{}}
\toprule
\multirow{2}{*}{\textbf{Model}}
  & \multicolumn{3}{c}{\textbf{SFT}} 
  & \multicolumn{6}{c}{\textbf{Close‐ended VQA}}
  & \multicolumn{7}{c}{\textbf{Open‐ended VQA}} 
  & \multirow{2}{*}{\textbf{Avg.}} \\
\cmidrule(lr){2-4} \cmidrule(lr){5-10} \cmidrule(lr){11-17}
  & \textbf{Report} & \textbf{VQA} & \textbf{Chat} & \textbf{Teeth} & \textbf{Patho} & \textbf{His} & \textbf{Jaw} & \textbf{Summ} & \textbf{Overall}
  & \textbf{Teeth} & \textbf{Patho} & \textbf{His} & \textbf{Jaw} & \textbf{Summ} & \textbf{Report} & \textbf{Overall} \\
  
\midrule
Qwen2.5-VL-7B~\cite{bai2025qwen2}   &\ding{55} &\ding{55} &\ding{55}    & 24.96 & 21.12 & 27.54 & 37.08 & 35.42 & 27.00 & 17.01 & 16.10 & 11.18 & 29.41 &  9.07 &  8.20 & 15.92 & 21.46\\

\midrule

\textbf{OralGPT} &\ding{51} &\ding{55} &\ding{55} &26.90 &27.33 &26.35 &45.32 &37.50 &31.00 &27.82 &15.82 &25.92 &63.76 &22.33 &38.00 &32.62 &31.81  \\ 
\textbf{OralGPT} &\ding{55} &\ding{51} &\ding{55} & 39.12 & 36.65 & 37.73 & 62.92 & 43.75 & 43.60 & 36.22 & 31.92 & 32.49 & 78.47 & 40.93 & 4.30 & 35.73 & 39.67 \\ 

\textbf{OralGPT} &\ding{51} &\ding{51} &\ding{55} &\textbf{43.19} &\textbf{40.99} &\textbf{43.11} &\textbf{63.60} &37.50 &\textbf{46.20} & 39.85 & 32.41 &35.20 &\textbf{78.06} &36.98 &36.80 &42.85 &44.53  \\ 
\textbf{OralGPT} &\ding{51} &\ding{51} &\ding{51} & 37.17 &30.43 &38.32 &52.81 &\textbf{45.83} &39.60 & \textbf{55.45} & \textbf{33.40} &\textbf{45.74} &74.47 &\textbf{45.17} &\textbf{50.50} &\textbf{52.77} &\textbf{46.19}  \\ 

\bottomrule
\end{tabular}%
}
\vspace{-0.45cm}
\end{table}

\subsection{Efficacy Validation of MMOral Instruction Data}
\label{sec:5.3}
We implement supervised fine-tuning (SFT) on the Qwen2.5-VL-7B model using our MMOral instruction data, and the results are presented in Table~\ref{tab:sft_results}. When using the MMOral-report or MMOral-VQA dataset individually for SFT, the average score on MMOral-Bench improved by 10.35\% (from 21.46\% to 31.81\%) and 18.21\% (from 21.46\% to 39.67\%), respectively. Furthermore, when both MMOral-report and MMOral-VQA are used together for SFT, the average score achieves a more significant improvement, rising from 21.46\% to 44.53\%. 
Based on this, incorporating the MMOral-Chat into the SFT process results in an additional 1.66\% improvement in the average score. Notably, OralGPT demonstrates significant improvements on open-ended QA tasks when MMOral-Chat is included in SFT, while exhibiting a slight decline in performance on closed-ended QA tasks. We hypothesize that image-grounded conversation data can significantly enhance the model's instruction-following ability for open-ended questions and improve overall user experience.

\subsection{Case Study}

\begin{figure}[t!]
  \centering
  \includegraphics[width=\textwidth]{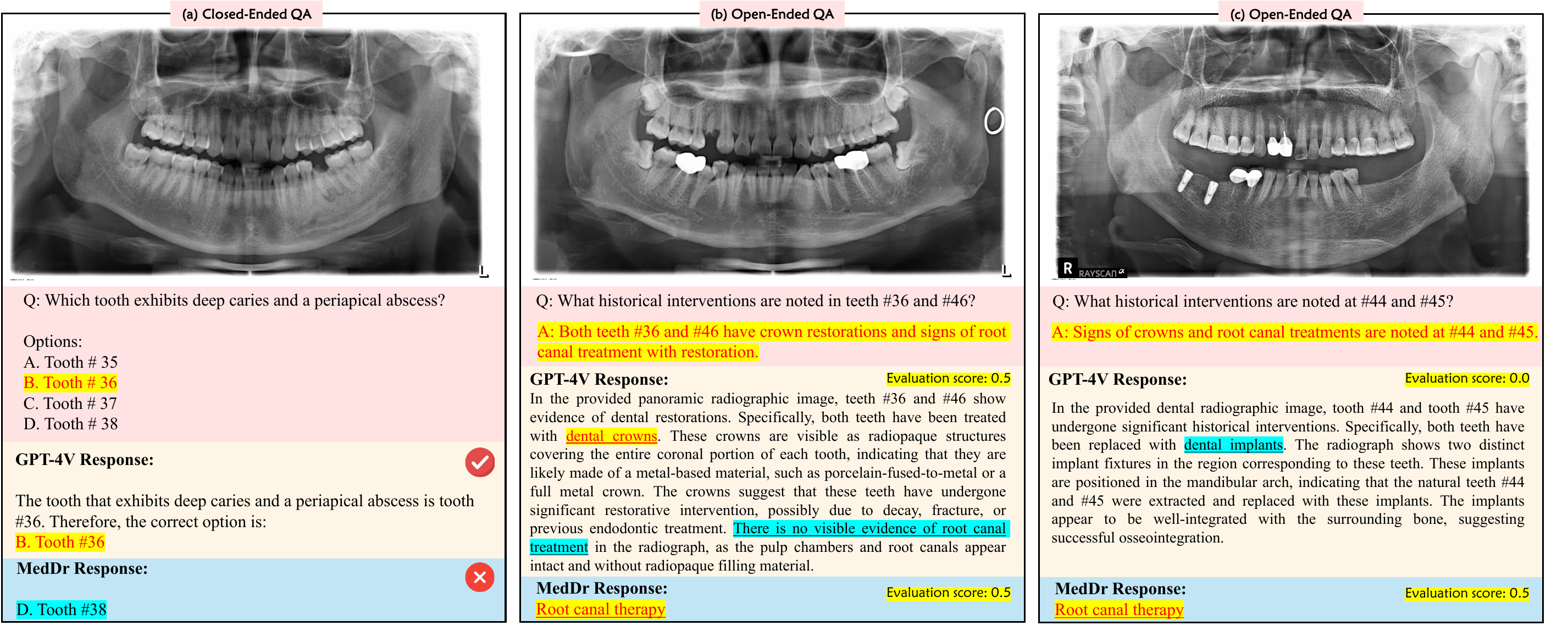}
  \vspace{-0.6cm}
  \caption{Three examples of case studies on closed-ended QA and open-ended QA, respectively. More examples can be found in the \textit{supplementary materials}.}
  \vspace{-0.4cm}
  \label{fig:case_study}
\end{figure}

To provide a more intuitive demonstration of the current capabilities of LVLMs in understanding panoramic X-rays, we conduct a case study on a high-performing proprietary model (GPT-4-Turbo) and a medical-specific LVLM (MedDr~\cite{he2024meddr}) by analyzing their performance on both closed-ended and open-ended question types. Figure~\ref{fig:case_study} (a) illustrates a closed-ended QA case, where the question requires the LVLM to simultaneously understand tooth numbering according to the FDI standard, detect caries, and identify periapical abscesses. As shown, GPT-4V could successfully answer the question, while MedDr provides an incorrect response. For the open-ended case, both GPT-4V and MedDr demonstrate some level of understanding of panoramic X-rays but fall short of providing comprehensive outcomes. For instance, as depicted in Figure~\ref{fig:case_study} (b), when examining teeth \#36 and \#46, which both exhibit crown restoration and root canal treatment, GPT-4-Turbo is able to detect the dental crown but explicitly states that no root canal treatment is identified. Conversely, MedDr detects the root canal treatment but overlooks the clearly visible crown restoration. Despite being among the top-performing models on the MMOral-Bench benchmark, GPT-4-Turbo and MedDr still exhibit significant limitations, highlighting substantial room for improvement in their ability to accurately interpret panoramic X-rays. Figure~\ref{fig:case_study} (c) shows the case that GPT-4V misrecognises the tooth \#44 and \#45, leading to a completely incorrect answer.

Additionally, we observe that some proprietary models, such as Qwen-Max-VL and Qwen-QVQ, commonly refuse to provide answers due to safety concerns stemming from commercial policies. For example, these models often respond with statements like, \textit{``Input data may contain inappropriate content.''} This strict adherence to safety protocols and ethical standards significantly limits their response capabilities in the dental field.

\section{Conclusion}
\label{sec:Conclusion_limitations}
We propose MMOral, the first large-scale multimodal instruction dataset and benchmark for panoramic X-ray understanding, and introduce OralGPT, a powerful multimodal vision-language model specifically designed for panoramic X-ray analysis.
The evaluation outcomes on MMOral-Bench reveal that existing universal medical-specific LVLMs have unsatisfactory performance when it comes to understanding panoramic X-rays characterized by dense anatomical structures and fine-grained pathological cues. We hope that our instruction dataset and benchmark will serve as a pivotal resource for advancing digital dentistry towards more sophisticated and intelligent multimodal AI in oral healthcare.

\textbf{Limitation and Future Work.} MMOral exhibits limitations in imaging modality diversity compared to existing medical instruction datasets. However, panoramic X-rays hold significant clinical value owing to their comprehensive visualization of the entire oral anatomy. Automated interpretation of panoramic X-rays could substantially advance intelligent dental AI. Future efforts will aim to expand coverage to additional oral imaging modalities, including 2D modalities such as periapical X-rays, intraoral photographs and cephalometric radiographs, as well as 3D modalities like cone-beam computed tomography (CBCT) and magnetic resonance imaging (MRI).

{
\small
\bibliography{neurips_2025}

\begin{thebibliography}{78}
\providecommand{\natexlab}[1]{#1}
\providecommand{\url}[1]{\texttt{#1}}
\expandafter\ifx\csname urlstyle\endcsname\relax
  \providecommand{\doi}[1]{doi: #1}\else
  \providecommand{\doi}{doi: \begingroup \urlstyle{rm}\Url}\fi

\bibitem[dat(2023{\natexlab{a}})]{dataset_1}
Dentistry computer vision project, 2023{\natexlab{a}}.
\newblock URL \url{https://universe.roboflow.com/nanyang-technological-university-kdgtt/dentistry-vibir}.

\bibitem[dat(2023{\natexlab{b}})]{dataset_2}
Dental caries detection computer vision project, 2023{\natexlab{b}}.
\newblock URL \url{https://universe.roboflow.com/panoramic-xray-images/dental-caries-detection}.

\bibitem[dat(2023{\natexlab{c}})]{dataset_5}
2023{\natexlab{c}}.
\newblock URL \url{https://www.kaggle.com/datasets/reemsalahshehab/dental?select=data.yaml}.

\bibitem[dat(2024{\natexlab{a}})]{dataset_3}
Dental diseases, 2024{\natexlab{a}}.
\newblock URL \url{https://www.kaggle.com/datasets/ayaalialnozahyy/dental-diseases/data}.

\bibitem[dat(2024{\natexlab{b}})]{dataset_4}
vzrad2 computer vision project, 2024{\natexlab{b}}.
\newblock URL \url{https://universe.roboflow.com/arshs-workspace-radio/vzrad2}.

\bibitem[dat(2024{\natexlab{c}})]{dataset_6}
2024{\natexlab{c}}.
\newblock URL \url{https://universe.roboflow.com/arshs-workspace-radio/vzrad2}.

\bibitem[dat(2024{\natexlab{d}})]{dataset_7}
2024{\natexlab{d}}.
\newblock URL \url{https://www.kaggle.com/datasets/nadaaglan/dental-periapical-x-rayss}.

\bibitem[Abdin et~al.(2024{\natexlab{a}})Abdin, Aneja, Awadalla, Awadallah, Awan, Bach, Bahree, Bakhtiari, Bao, Behl, et~al.]{abdin2024phi3}
M.~Abdin, J.~Aneja, H.~Awadalla, A.~Awadallah, A.~A. Awan, N.~Bach, A.~Bahree, A.~Bakhtiari, J.~Bao, H.~Behl, et~al.
\newblock Phi-3 technical report: A highly capable language model locally on your phone.
\newblock \emph{arXiv preprint arXiv:2404.14219}, 2024{\natexlab{a}}.

\bibitem[Abdin et~al.(2024{\natexlab{b}})Abdin, Aneja, Behl, Bubeck, Eldan, Gunasekar, Harrison, Hewett, Javaheripi, Kauffmann, et~al.]{abdin2024phi4}
M.~Abdin, J.~Aneja, H.~Behl, S.~Bubeck, R.~Eldan, S.~Gunasekar, M.~Harrison, R.~J. Hewett, M.~Javaheripi, P.~Kauffmann, et~al.
\newblock Phi-4 technical report.
\newblock \emph{arXiv preprint arXiv:2412.08905}, 2024{\natexlab{b}}.

\bibitem[Achiam et~al.(2023)Achiam, Adler, Agarwal, Ahmad, Akkaya, Aleman, Almeida, Altenschmidt, Altman, Anadkat, et~al.]{achiam2023gpt4_turbo}
J.~Achiam, S.~Adler, S.~Agarwal, L.~Ahmad, I.~Akkaya, F.~L. Aleman, D.~Almeida, J.~Altenschmidt, S.~Altman, S.~Anadkat, et~al.
\newblock Gpt-4 technical report.
\newblock \emph{arXiv preprint arXiv:2303.08774}, 2023.

\bibitem[Anthropic()]{Claude3S}
Anthropic.
\newblock Claude 3.7 sonnet system card.
\newblock URL \url{https://api.semanticscholar.org/CorpusID:276612236}.

\bibitem[Bai et~al.(2023)Bai, Bai, Yang, Wang, Tan, Wang, Lin, Zhou, and Zhou]{Qwen-VL}
J.~Bai, S.~Bai, S.~Yang, S.~Wang, S.~Tan, P.~Wang, J.~Lin, C.~Zhou, and J.~Zhou.
\newblock Qwen-vl: A versatile vision-language model for understanding, localization, text reading, and beyond.
\newblock \emph{arXiv preprint arXiv:2308.12966}, 2023.

\bibitem[Bai et~al.(2025)Bai, Chen, Liu, Wang, Ge, Song, Dang, Wang, Wang, Tang, et~al.]{bai2025qwen2}
S.~Bai, K.~Chen, X.~Liu, J.~Wang, W.~Ge, S.~Song, K.~Dang, P.~Wang, S.~Wang, J.~Tang, et~al.
\newblock Qwen2. 5-vl technical report.
\newblock \emph{arXiv preprint arXiv:2502.13923}, 2025.

\bibitem[Beyer et~al.(2024)Beyer, Steiner, Pinto, Kolesnikov, Wang, Salz, Neumann, Alabdulmohsin, Tschannen, Bugliarello, et~al.]{beyer2024paligemma}
L.~Beyer, A.~Steiner, A.~S. Pinto, A.~Kolesnikov, X.~Wang, D.~Salz, M.~Neumann, I.~Alabdulmohsin, M.~Tschannen, E.~Bugliarello, et~al.
\newblock Paligemma: A versatile 3b vlm for transfer.
\newblock \emph{arXiv preprint arXiv:2407.07726}, 2024.

\bibitem[Chen et~al.(2024{\natexlab{a}})Chen, Gui, Ouyang, Gao, Chen, Chen, Wang, Zhang, Cai, Ji, et~al.]{chen2024huatuogpt}
J.~Chen, C.~Gui, R.~Ouyang, A.~Gao, S.~Chen, G.~H. Chen, X.~Wang, R.~Zhang, Z.~Cai, K.~Ji, et~al.
\newblock Huatuogpt-vision, towards injecting medical visual knowledge into multimodal llms at scale.
\newblock \emph{arXiv preprint arXiv:2406.19280}, 2024{\natexlab{a}}.

\bibitem[Chen et~al.(2024{\natexlab{b}})Chen, Wang, Cao, Liu, Gao, Cui, Zhu, Ye, Tian, Liu, et~al.]{chen2024expanding}
Z.~Chen, W.~Wang, Y.~Cao, Y.~Liu, Z.~Gao, E.~Cui, J.~Zhu, S.~Ye, H.~Tian, Z.~Liu, et~al.
\newblock Expanding performance boundaries of open-source multimodal models with model, data, and test-time scaling.
\newblock \emph{arXiv preprint arXiv:2412.05271}, 2024{\natexlab{b}}.

\bibitem[de~Oliveira~Capote et~al.(2015)de~Oliveira~Capote, de~Almeida~Gon{\c{c}}alves, Gon{\c{c}}alves, and Gon{\c{c}}alves]{de2015panoramic}
T.~S. de~Oliveira~Capote, M.~de~Almeida~Gon{\c{c}}alves, A.~Gon{\c{c}}alves, and M.~Gon{\c{c}}alves.
\newblock Panoramic radiography—diagnosis of relevant structures that might compromise oral and general health of the patient.
\newblock In \emph{Emerging Trends in Oral Health Sciences and Dentistry}. IntechOpen, 2015.

\bibitem[Deitke et~al.(2024)Deitke, Clark, Lee, Tripathi, Yang, Park, Salehi, Muennighoff, Lo, Soldaini, et~al.]{deitke2024molmo}
M.~Deitke, C.~Clark, S.~Lee, R.~Tripathi, Y.~Yang, J.~S. Park, M.~Salehi, N.~Muennighoff, K.~Lo, L.~Soldaini, et~al.
\newblock Molmo and pixmo: Open weights and open data for state-of-the-art multimodal models.
\newblock \emph{arXiv preprint arXiv:2409.17146}, 2024.

\bibitem[Do et~al.(2024)Do, Vo, Nguyen, Luong, Cu, and Le]{do2024dataset}
H.~V. Do, T.~N.~N. Vo, P.~T. Nguyen, T.~H.~L. Luong, N.~G. Cu, and H.~S. Le.
\newblock A dataset of apical periodontitis lesions in panoramic radiographs for deep-learning-based classification and detection.
\newblock \emph{Data in Brief}, 54:\penalty0 110486, 2024.

\bibitem[Dong et~al.(2024)Dong, Han, Peng, Qi, Ge, Yang, Zhao, Sun, Zhou, Wei, et~al.]{dong2023dreamllm}
R.~Dong, C.~Han, Y.~Peng, Z.~Qi, Z.~Ge, J.~Yang, L.~Zhao, J.~Sun, H.~Zhou, H.~Wei, et~al.
\newblock Dreamllm: Synergistic multimodal comprehension and creation.
\newblock \emph{International Conference on Learning Representations}, 2024.

\bibitem[Du et~al.(2024)Du, Chen, Xie, Jia, and Jiang]{Du2024SVTRv2}
Y.~Du, Z.~Chen, H.~Xie, C.~Jia, and Y.-G. Jiang.
\newblock Svtrv2: Ctc beats encoder-decoder models in scene text recognition.
\newblock \emph{CoRR}, abs/2411.15858, 2024.
\newblock URL \url{https://arxiv.org/abs/2411.15858}.

\bibitem[Duan et~al.(2024)Duan, Yang, Qiao, Fang, Chen, Liu, Dong, Zang, Zhang, Wang, et~al.]{duan2024vlmevalkit}
H.~Duan, J.~Yang, Y.~Qiao, X.~Fang, L.~Chen, Y.~Liu, X.~Dong, Y.~Zang, P.~Zhang, J.~Wang, et~al.
\newblock Vlmevalkit: An open-source toolkit for evaluating large multi-modality models.
\newblock In \emph{Proceedings of the 32nd ACM International Conference on Multimedia}, pages 11198--11201, 2024.

\bibitem[GLM et~al.(2024)GLM, Zeng, Xu, Wang, Zhang, Yin, Rojas, Feng, Zhao, Lai, Yu, Wang, Sun, Zhang, Cheng, Gui, Tang, Zhang, Li, Zhao, Wu, Zhong, Liu, Huang, Zhang, Zheng, Lu, Duan, Zhang, Cao, Yang, Tam, Zhao, Liu, Xia, Zhang, Gu, Lv, Liu, Liu, Yang, Song, Zhang, An, Xu, Niu, Yang, Li, Bai, Dong, Qi, Wang, Yang, Du, Hou, and Wang]{glm2024chatglm}
T.~GLM, A.~Zeng, B.~Xu, B.~Wang, C.~Zhang, D.~Yin, D.~Rojas, G.~Feng, H.~Zhao, H.~Lai, H.~Yu, H.~Wang, J.~Sun, J.~Zhang, J.~Cheng, J.~Gui, J.~Tang, J.~Zhang, J.~Li, L.~Zhao, L.~Wu, L.~Zhong, M.~Liu, M.~Huang, P.~Zhang, Q.~Zheng, R.~Lu, S.~Duan, S.~Zhang, S.~Cao, S.~Yang, W.~L. Tam, W.~Zhao, X.~Liu, X.~Xia, X.~Zhang, X.~Gu, X.~Lv, X.~Liu, X.~Liu, X.~Yang, X.~Song, X.~Zhang, Y.~An, Y.~Xu, Y.~Niu, Y.~Yang, Y.~Li, Y.~Bai, Y.~Dong, Z.~Qi, Z.~Wang, Z.~Yang, Z.~Du, Z.~Hou, and Z.~Wang.
\newblock Chatglm: A family of large language models from glm-130b to glm-4 all tools, 2024.

\bibitem[Grattafiori et~al.(2024)Grattafiori, Dubey, Jauhri, Pandey, Kadian, Al-Dahle, Letman, Mathur, Schelten, Vaughan, et~al.]{llama3_2}
A.~Grattafiori, A.~Dubey, A.~Jauhri, A.~Pandey, A.~Kadian, A.~Al-Dahle, A.~Letman, A.~Mathur, A.~Schelten, A.~Vaughan, et~al.
\newblock The llama 3 herd of models.
\newblock \emph{arXiv preprint arXiv:2407.21783}, 2024.

\bibitem[Guo and Huang(2025)]{guo2025llava}
Y.~Guo and W.~Huang.
\newblock Llava-next-med: Medical multimodal large language model.
\newblock In \emph{2025 Asia-Europe Conference on Cybersecurity, Internet of Things and Soft Computing (CITSC)}, pages 474--477. IEEE, 2025.

\bibitem[Hamamci et~al.(2023)Hamamci, Er, Simsar, Yuksel, Gultekin, Ozdemir, Yang, Li, Pati, Stadlinger, et~al.]{hamamci2023dentex}
I.~E. Hamamci, S.~Er, E.~Simsar, A.~E. Yuksel, S.~Gultekin, S.~D. Ozdemir, K.~Yang, H.~B. Li, S.~Pati, B.~Stadlinger, et~al.
\newblock Dentex: An abnormal tooth detection with dental enumeration and diagnosis benchmark for panoramic x-rays.
\newblock \emph{arXiv preprint arXiv:2305.19112}, 2023.

\bibitem[Hao et~al.(2024)Hao, Zhu, He, Liu, Tsoi, and Hung]{ted3}
J.~Hao, Y.~Zhu, L.~He, M.~Liu, J.~K.~H. Tsoi, and K.~F. Hung.
\newblock T-mamba: A unified framework with long-range dependency in dual-domain for 2d \& 3d tooth segmentation.
\newblock \emph{arXiv preprint arXiv:2404.01065}, 2024.

\bibitem[He et~al.(2024)He, Nie, Chen, Cai, Wang, Yang, and Chen]{he2024meddr}
S.~He, Y.~Nie, Z.~Chen, Z.~Cai, H.~Wang, S.~Yang, and H.~Chen.
\newblock Meddr: Diagnosis-guided bootstrapping for large-scale medical vision-language learning.
\newblock \emph{arXiv e-prints}, pages arXiv--2404, 2024.

\bibitem[He et~al.(2020)He, Zhang, Mou, Xing, and Xie]{he2020pathvqa}
X.~He, Y.~Zhang, L.~Mou, E.~Xing, and P.~Xie.
\newblock Pathvqa: 30000+ questions for medical visual question answering.
\newblock \emph{arXiv preprint arXiv:2003.10286}, 2020.

\bibitem[He et~al.(2023)He, Wang, Yan, Liu, Chang, Gentili, McAuley, and Hsu]{he2023medeval}
Z.~He, Y.~Wang, A.~Yan, Y.~Liu, E.~Y. Chang, A.~Gentili, J.~McAuley, and C.-N. Hsu.
\newblock Medeval: a multi-level, multi-task, and multi-domain medical benchmark for language model evaluation.
\newblock \emph{arXiv preprint arXiv:2310.14088}, 2023.

\bibitem[Hu et~al.(2024)Hu, Li, Lu, Shao, He, Qiao, and Luo]{hu2024omnimedvqa}
Y.~Hu, T.~Li, Q.~Lu, W.~Shao, J.~He, Y.~Qiao, and P.~Luo.
\newblock Omnimedvqa: A new large-scale comprehensive evaluation benchmark for medical lvlm.
\newblock In \emph{Proceedings of the IEEE/CVF Conference on Computer Vision and Pattern Recognition}, pages 22170--22183, 2024.

\bibitem[Hurst et~al.(2024)Hurst, Lerer, Goucher, Perelman, Ramesh, Clark, Ostrow, Welihinda, Hayes, Radford, et~al.]{hurst2024gpt}
A.~Hurst, A.~Lerer, A.~P. Goucher, A.~Perelman, A.~Ramesh, A.~Clark, A.~Ostrow, A.~Welihinda, A.~Hayes, A.~Radford, et~al.
\newblock Gpt-4o system card.
\newblock \emph{arXiv preprint arXiv:2410.21276}, 2024.

\bibitem[Lau et~al.(2018)Lau, Gayen, Ben~Abacha, and Demner-Fushman]{lau2018dataset}
J.~J. Lau, S.~Gayen, A.~Ben~Abacha, and D.~Demner-Fushman.
\newblock A dataset of clinically generated visual questions and answers about radiology images.
\newblock \emph{Scientific data}, 5\penalty0 (1):\penalty0 1--10, 2018.

\bibitem[Li et~al.(2024{\natexlab{a}})Li, Zhang, Guo, Zhang, Li, Zhang, Zhang, Zhang, Li, Liu, et~al.]{li2024llava}
B.~Li, Y.~Zhang, D.~Guo, R.~Zhang, F.~Li, H.~Zhang, K.~Zhang, P.~Zhang, Y.~Li, Z.~Liu, et~al.
\newblock Llava-onevision: Easy visual task transfer.
\newblock \emph{arXiv preprint arXiv:2408.03326}, 2024{\natexlab{a}}.

\bibitem[Li et~al.(2023)Li, Wong, Zhang, Usuyama, Liu, Yang, Naumann, Poon, and Gao]{li2023llavamed}
C.~Li, C.~Wong, S.~Zhang, N.~Usuyama, H.~Liu, J.~Yang, T.~Naumann, H.~Poon, and J.~Gao.
\newblock Llava-med: Training a large language-and-vision assistant for biomedicine in one day.
\newblock \emph{Advances in Neural Information Processing Systems}, 36:\penalty0 28541--28564, 2023.

\bibitem[Li et~al.(2025)Li, Lin, Lin, Zhang, Liu, Yang, Li, He, Song, Xiao, et~al.]{li2025eyecaregpt}
S.~Li, T.~Lin, L.~Lin, W.~Zhang, J.~Liu, X.~Yang, J.~Li, Y.~He, X.~Song, J.~Xiao, et~al.
\newblock Eyecaregpt: Boosting comprehensive ophthalmology understanding with tailored dataset, benchmark and model.
\newblock \emph{arXiv preprint arXiv:2504.13650}, 2025.

\bibitem[Li et~al.(2024{\natexlab{b}})Li, Su, Li, Fu, Chen, Huang, Wang, Ma, Chen, Hu, et~al.]{li2024gmai_model}
T.~Li, Y.~Su, W.~Li, B.~Fu, Z.~Chen, Z.~Huang, G.~Wang, C.~Ma, Y.~Chen, M.~Hu, et~al.
\newblock Gmai-vl \& gmai-vl-5.5 m: A large vision-language model and a comprehensive multimodal dataset towards general medical ai.
\newblock \emph{arXiv preprint arXiv:2411.14522}, 2024{\natexlab{b}}.

\bibitem[Lin et~al.(2025)Lin, Zhang, Li, Yuan, Yu, Li, He, Jiang, Li, Song, et~al.]{lin2025healthgpt}
T.~Lin, W.~Zhang, S.~Li, Y.~Yuan, B.~Yu, H.~Li, W.~He, H.~Jiang, M.~Li, X.~Song, et~al.
\newblock Healthgpt: A medical large vision-language model for unifying comprehension and generation via heterogeneous knowledge adaptation.
\newblock \emph{arXiv preprint arXiv:2502.09838}, 2025.

\bibitem[Liu et~al.(2021)Liu, Zhan, Xu, Ma, Yang, and Wu]{liu2021slake}
B.~Liu, L.-M. Zhan, L.~Xu, L.~Ma, Y.~Yang, and X.-M. Wu.
\newblock Slake: A semantically-labeled knowledge-enhanced dataset for medical visual question answering.
\newblock In \emph{2021 IEEE 18th international symposium on biomedical imaging (ISBI)}, pages 1650--1654. IEEE, 2021.

\bibitem[Liu et~al.(2023)Liu, Li, Wu, and Lee]{liu2023visual}
H.~Liu, C.~Li, Q.~Wu, and Y.~J. Lee.
\newblock Visual instruction tuning.
\newblock \emph{Advances in neural information processing systems}, 36:\penalty0 34892--34916, 2023.

\bibitem[Liu et~al.(2024{\natexlab{a}})Liu, Li, Li, Li, Zhang, Shen, and Lee]{liu2024llavanext}
H.~Liu, C.~Li, Y.~Li, B.~Li, Y.~Zhang, S.~Shen, and Y.~J. Lee.
\newblock Llava-next: Improved reasoning, ocr, and world knowledge, January 2024{\natexlab{a}}.
\newblock URL \url{https://llava-vl.github.io/blog/2024-01-30-llava-next/}.

\bibitem[Liu et~al.(2024{\natexlab{b}})Liu, Hu, Ding, Xu, Li, Zhu, Bai, Shi, Wang, Song, et~al.]{liu2024medbench}
M.~Liu, W.~Hu, J.~Ding, J.~Xu, X.~Li, L.~Zhu, Z.~Bai, X.~Shi, B.~Wang, H.~Song, et~al.
\newblock Medbench: A comprehensive, standardized, and reliable benchmarking system for evaluating chinese medical large language models.
\newblock \emph{Big Data Mining and Analytics}, 7\penalty0 (4):\penalty0 1116--1128, 2024{\natexlab{b}}.

\bibitem[Lu et~al.(2024{\natexlab{a}})Lu, Liu, Zhang, Wang, Dong, Liu, Sun, Ren, Li, Yang, et~al.]{lu2024deepseek}
H.~Lu, W.~Liu, B.~Zhang, B.~Wang, K.~Dong, B.~Liu, J.~Sun, T.~Ren, Z.~Li, H.~Yang, et~al.
\newblock Deepseek-vl: towards real-world vision-language understanding.
\newblock \emph{arXiv preprint arXiv:2403.05525}, 2024{\natexlab{a}}.

\bibitem[Lu et~al.(2023)Lu, Peng, Cheng, Galley, Chang, Wu, Zhu, and Gao]{lu2023chameleon}
P.~Lu, B.~Peng, H.~Cheng, M.~Galley, K.-W. Chang, Y.~N. Wu, S.-C. Zhu, and J.~Gao.
\newblock Chameleon: Plug-and-play compositional reasoning with large language models.
\newblock \emph{Advances in Neural Information Processing Systems}, 36:\penalty0 43447--43478, 2023.

\bibitem[Lu et~al.(2024{\natexlab{b}})Lu, Li, Chen, Xu, Luo, Zhang, and Ye]{lu2024ovis}
S.~Lu, Y.~Li, Q.-G. Chen, Z.~Xu, W.~Luo, K.~Zhang, and H.-J. Ye.
\newblock Ovis: Structural embedding alignment for multimodal large language model.
\newblock \emph{arXiv preprint arXiv:2405.20797}, 2024{\natexlab{b}}.

\bibitem[Moor et~al.(2023)Moor, Huang, Wu, Yasunaga, Dalmia, Leskovec, Zakka, Reis, and Rajpurkar]{moor2023med}
M.~Moor, Q.~Huang, S.~Wu, M.~Yasunaga, Y.~Dalmia, J.~Leskovec, C.~Zakka, E.~P. Reis, and P.~Rajpurkar.
\newblock Med-flamingo: a multimodal medical few-shot learner.
\newblock In \emph{Machine Learning for Health (ML4H)}, pages 353--367. PMLR, 2023.

\bibitem[Pan et~al.(2025)Pan, Liu, Wu, Liu, Zhu, Li, Chen, Ouyang, and Rueckert]{pan2025medvlm}
J.~Pan, C.~Liu, J.~Wu, F.~Liu, J.~Zhu, H.~B. Li, C.~Chen, C.~Ouyang, and D.~Rueckert.
\newblock Medvlm-r1: Incentivizing medical reasoning capability of vision-language models (vlms) via reinforcement learning.
\newblock \emph{arXiv preprint arXiv:2502.19634}, 2025.

\bibitem[Panetta et~al.(2021)Panetta, Rajendran, Ramesh, Rao, and Agaian]{panetta2021tufts}
K.~Panetta, R.~Rajendran, A.~Ramesh, S.~P. Rao, and S.~Agaian.
\newblock Tufts dental database: a multimodal panoramic x-ray dataset for benchmarking diagnostic systems.
\newblock \emph{IEEE journal of biomedical and health informatics}, 26\penalty0 (4):\penalty0 1650--1659, 2021.

\bibitem[Park et~al.(2024)Park, Kim, Yoon, Hyun, and Choi]{park2024m4cxr}
J.~Park, S.~Kim, B.~Yoon, J.~Hyun, and K.~Choi.
\newblock M4cxr: Exploring multi-task potentials of multi-modal large language models for chest x-ray interpretation.
\newblock \emph{arXiv preprint arXiv:2408.16213}, 2024.

\bibitem[Peng et~al.(2023)Peng, Wang, Dong, Hao, Huang, Ma, and Wei]{peng2023kosmos}
Z.~Peng, W.~Wang, L.~Dong, Y.~Hao, S.~Huang, S.~Ma, and F.~Wei.
\newblock Kosmos-2: Grounding multimodal large language models to the world.
\newblock \emph{arXiv preprint arXiv:2306.14824}, 2023.

\bibitem[Qin et~al.(2024)Qin, Yin, Campbell, Wu, Zou, Tham, Liu, Zhang, and Chen]{qin2024lmod}
Z.~Qin, Y.~Yin, D.~Campbell, X.~Wu, K.~Zou, Y.-C. Tham, N.~Liu, X.~Zhang, and Q.~Chen.
\newblock Lmod: A large multimodal ophthalmology dataset and benchmark for large vision-language models.
\newblock \emph{arXiv preprint arXiv:2410.01620}, 2024.

\bibitem[Sun et~al.(2024)Sun, Wu, Zhu, Zheng, Chen, Zhang, Zhang, Wan, Lan, Zheng, et~al.]{sun2024pathmmu}
Y.~Sun, H.~Wu, C.~Zhu, S.~Zheng, Q.~Chen, K.~Zhang, Y.~Zhang, D.~Wan, X.~Lan, M.~Zheng, et~al.
\newblock Pathmmu: A massive multimodal expert-level benchmark for understanding and reasoning in pathology.
\newblock In \emph{European Conference on Computer Vision}, pages 56--73. Springer, 2024.

\bibitem[Team et~al.(2023)Team, Anil, Borgeaud, Alayrac, Yu, Soricut, Schalkwyk, Dai, Hauth, Millican, et~al.]{team2023gemini}
G.~Team, R.~Anil, S.~Borgeaud, J.-B. Alayrac, J.~Yu, R.~Soricut, J.~Schalkwyk, A.~M. Dai, A.~Hauth, K.~Millican, et~al.
\newblock Gemini: a family of highly capable multimodal models.
\newblock \emph{arXiv preprint arXiv:2312.11805}, 2023.

\bibitem[Team et~al.(2025{\natexlab{a}})Team, Kamath, Ferret, Pathak, Vieillard, Merhej, Perrin, Matejovicova, Ram{\'e}, Rivi{\`e}re, et~al.]{team2025gemma}
G.~Team, A.~Kamath, J.~Ferret, S.~Pathak, N.~Vieillard, R.~Merhej, S.~Perrin, T.~Matejovicova, A.~Ram{\'e}, M.~Rivi{\`e}re, et~al.
\newblock Gemma 3 technical report.
\newblock \emph{arXiv preprint arXiv:2503.19786}, 2025{\natexlab{a}}.

\bibitem[Team et~al.(2025{\natexlab{b}})Team, Du, Yin, Xing, Qu, Wang, Chen, Zhang, Du, Wei, et~al.]{team2025kimi}
K.~Team, A.~Du, B.~Yin, B.~Xing, B.~Qu, B.~Wang, C.~Chen, C.~Zhang, C.~Du, C.~Wei, et~al.
\newblock Kimi-vl technical report.
\newblock \emph{arXiv preprint arXiv:2504.07491}, 2025{\natexlab{b}}.

\bibitem[Team(2024)]{qvq-72b-preview}
Q.~Team.
\newblock Qvq: To see the world with wisdom, December 2024.
\newblock URL \url{https://qwenlm.github.io/blog/qvq-72b-preview/}.

\bibitem[Tong et~al.(2024)Tong, Brown, Wu, Woo, IYER, Akula, Yang, Yang, Middepogu, Wang, et~al.]{tong2024cambrian}
P.~Tong, E.~Brown, P.~Wu, S.~Woo, A.~J.~V. IYER, S.~C. Akula, S.~Yang, J.~Yang, M.~Middepogu, Z.~Wang, et~al.
\newblock Cambrian-1: A fully open, vision-centric exploration of multimodal llms.
\newblock \emph{Advances in Neural Information Processing Systems}, 37:\penalty0 87310--87356, 2024.

\bibitem[Turosz et~al.(2023)Turosz, Ch{\k{e}}ci{\'n}ska, Ch{\k{e}}ci{\'n}ski, Brzozowska, Nowak, and Sikora]{turosz2023applications}
N.~Turosz, K.~Ch{\k{e}}ci{\'n}ska, M.~Ch{\k{e}}ci{\'n}ski, A.~Brzozowska, Z.~Nowak, and M.~Sikora.
\newblock Applications of artificial intelligence in the analysis of dental panoramic radiographs: an overview of systematic reviews.
\newblock \emph{Dentomaxillofacial Radiology}, 52\penalty0 (7):\penalty0 20230284, 2023.

\bibitem[Wang et~al.(2024{\natexlab{a}})Wang, Bai, Tan, Wang, Fan, Bai, Chen, Liu, Wang, Ge, et~al.]{wang2024qwen2}
P.~Wang, S.~Bai, S.~Tan, S.~Wang, Z.~Fan, J.~Bai, K.~Chen, X.~Liu, J.~Wang, W.~Ge, et~al.
\newblock Qwen2-vl: Enhancing vision-language model's perception of the world at any resolution.
\newblock \emph{arXiv preprint arXiv:2409.12191}, 2024{\natexlab{a}}.

\bibitem[Wang et~al.(2024{\natexlab{b}})Wang, Lv, Yu, Hong, Qi, Wang, Ji, Yang, Zhao, XiXuan, et~al.]{wang2024cogvlm}
W.~Wang, Q.~Lv, W.~Yu, W.~Hong, J.~Qi, Y.~Wang, J.~Ji, Z.~Yang, L.~Zhao, S.~XiXuan, et~al.
\newblock Cogvlm: Visual expert for pretrained language models.
\newblock \emph{Advances in Neural Information Processing Systems}, 37:\penalty0 121475--121499, 2024{\natexlab{b}}.

\bibitem[Wang et~al.(2024{\natexlab{c}})Wang, Wang, Li, Ma, Wang, Jiang, Li, and Tang]{wang2024cxpmrg}
X.~Wang, F.~Wang, Y.~Li, Q.~Ma, S.~Wang, B.~Jiang, C.~Li, and J.~Tang.
\newblock Cxpmrg-bench: Pre-training and benchmarking for x-ray medical report generation on chexpert plus dataset.
\newblock \emph{arXiv preprint arXiv:2410.00379}, 2024{\natexlab{c}}.

\bibitem[Wang et~al.(2024{\natexlab{d}})Wang, Zhang, Luo, Sun, Cui, Wang, Zhang, Wang, Li, Yu, et~al.]{wang2024emu3}
X.~Wang, X.~Zhang, Z.~Luo, Q.~Sun, Y.~Cui, J.~Wang, F.~Zhang, Y.~Wang, Z.~Li, Q.~Yu, et~al.
\newblock Emu3: Next-token prediction is all you need.
\newblock \emph{arXiv preprint arXiv:2409.18869}, 2024{\natexlab{d}}.

\bibitem[Wei et~al.(2024)Wei, Kong, Chen, Zhao, Ge, Yang, Sun, Han, and Zhang]{wei2024vary}
H.~Wei, L.~Kong, J.~Chen, L.~Zhao, Z.~Ge, J.~Yang, J.~Sun, C.~Han, and X.~Zhang.
\newblock Vary: Scaling up the vision vocabulary for large vision-language model.
\newblock In \emph{European Conference on Computer Vision}, pages 408--424. Springer, 2024.

\bibitem[Wong and Eisel{\'e}(2015)]{wong2015fdi}
T.~Wong and J.-L. Eisel{\'e}.
\newblock Fdi world dental federation: Responding to new realities of oral health, 2015.

\bibitem[Wright and Reeves(2016)]{wright2016radbench}
C.~Wright and P.~Reeves.
\newblock Radbench: benchmarking image interpretation skills.
\newblock \emph{Radiography}, 22\penalty0 (2):\penalty0 e131--e136, 2016.

\bibitem[Yao et~al.(2024)Yao, Yu, Zhang, Wang, Cui, Zhu, Cai, Li, Zhao, He, et~al.]{yao2024minicpm}
Y.~Yao, T.~Yu, A.~Zhang, C.~Wang, J.~Cui, H.~Zhu, T.~Cai, H.~Li, W.~Zhao, Z.~He, et~al.
\newblock Minicpm-v: A gpt-4v level mllm on your phone.
\newblock \emph{arXiv preprint arXiv:2408.01800}, 2024.

\bibitem[Ye et~al.(2024{\natexlab{a}})Ye, Wang, Li, Deng, Li, Li, Duan, Huang, Su, Wang, et~al.]{ye2024gmai}
J.~Ye, G.~Wang, Y.~Li, Z.~Deng, W.~Li, T.~Li, H.~Duan, Z.~Huang, Y.~Su, B.~Wang, et~al.
\newblock Gmai-mmbench: A comprehensive multimodal evaluation benchmark towards general medical ai.
\newblock \emph{Advances in Neural Information Processing Systems}, 37:\penalty0 94327--94427, 2024{\natexlab{a}}.

\bibitem[Ye et~al.(2024{\natexlab{b}})Ye, Xu, Liu, Hu, Yan, Qian, Zhang, Huang, and Zhou]{ye2024mplug}
J.~Ye, H.~Xu, H.~Liu, A.~Hu, M.~Yan, Q.~Qian, J.~Zhang, F.~Huang, and J.~Zhou.
\newblock mplug-owl3: Towards long image-sequence understanding in multi-modal large language models.
\newblock \emph{arXiv preprint arXiv:2408.04840}, 2024{\natexlab{b}}.

\bibitem[Ye et~al.(2024{\natexlab{c}})Ye, Xu, Ye, Yan, Hu, Liu, Qian, Zhang, and Huang]{ye2024mplugowl2}
Q.~Ye, H.~Xu, J.~Ye, M.~Yan, A.~Hu, H.~Liu, Q.~Qian, J.~Zhang, and F.~Huang.
\newblock mplug-owl2: Revolutionizing multi-modal large language model with modality collaboration.
\newblock In \emph{Proceedings of the ieee/cvf conference on computer vision and pattern recognition}, pages 13040--13051, 2024{\natexlab{c}}.

\bibitem[Ying et~al.(2024)Ying, Meng, Wang, Li, Lin, Yang, Zhang, Zhang, Lin, Liu, et~al.]{ying2024mmt}
K.~Ying, F.~Meng, J.~Wang, Z.~Li, H.~Lin, Y.~Yang, H.~Zhang, W.~Zhang, Y.~Lin, S.~Liu, et~al.
\newblock Mmt-bench: A comprehensive multimodal benchmark for evaluating large vision-language models towards multitask agi.
\newblock \emph{arXiv preprint arXiv:2404.16006}, 2024.

\bibitem[Young et~al.(2024)Young, Chen, Li, Huang, Zhang, Zhang, Wang, Li, Zhu, Chen, et~al.]{young2024yi}
A.~Young, B.~Chen, C.~Li, C.~Huang, G.~Zhang, G.~Zhang, G.~Wang, H.~Li, J.~Zhu, J.~Chen, et~al.
\newblock Yi: Open foundation models by 01. ai.
\newblock \emph{arXiv preprint arXiv:2403.04652}, 2024.

\bibitem[Yu et~al.(2024{\natexlab{a}})Yu, Zhao, Wei, Yang, Wu, Kong, Wei, Wang, Ge, Zhang, et~al.]{yu2024merlin}
E.~Yu, L.~Zhao, Y.~Wei, J.~Yang, D.~Wu, L.~Kong, H.~Wei, T.~Wang, Z.~Ge, X.~Zhang, et~al.
\newblock Merlin: Empowering multimodal llms with foresight minds.
\newblock In \emph{European Conference on Computer Vision}, pages 425--443. Springer, 2024{\natexlab{a}}.

\bibitem[Yu et~al.(2023)Yu, Yang, Li, Wang, Lin, Liu, Wang, and Wang]{yu2023mmvetv1}
W.~Yu, Z.~Yang, L.~Li, J.~Wang, K.~Lin, Z.~Liu, X.~Wang, and L.~Wang.
\newblock Mm-vet: Evaluating large multimodal models for integrated capabilities.
\newblock \emph{arXiv preprint arXiv:2308.02490}, 2023.

\bibitem[Yu et~al.(2024{\natexlab{b}})Yu, Yang, Ren, Li, Wang, Lin, Lin, Liu, Wang, and Wang]{yu2024mmvetv2}
W.~Yu, Z.~Yang, L.~Ren, L.~Li, J.~Wang, K.~Lin, C.-C. Lin, Z.~Liu, L.~Wang, and X.~Wang.
\newblock Mm-vet v2: A challenging benchmark to evaluate large multimodal models for integrated capabilities.
\newblock \emph{arXiv preprint arXiv:2408.00765}, 2024{\natexlab{b}}.

\bibitem[Yue et~al.(2024)Yue, Ni, Zhang, Zheng, Liu, Zhang, Stevens, Jiang, Ren, Sun, et~al.]{yue2024mmmu}
X.~Yue, Y.~Ni, K.~Zhang, T.~Zheng, R.~Liu, G.~Zhang, S.~Stevens, D.~Jiang, W.~Ren, Y.~Sun, et~al.
\newblock Mmmu: A massive multi-discipline multimodal understanding and reasoning benchmark for expert agi.
\newblock In \emph{Proceedings of the IEEE/CVF Conference on Computer Vision and Pattern Recognition}, pages 9556--9567, 2024.

\bibitem[Zhang et~al.(2024)Zhang, Dong, Zang, Cao, Qian, Chen, Guo, Duan, Wang, Ouyang, et~al.]{zhang2024internlm}
P.~Zhang, X.~Dong, Y.~Zang, Y.~Cao, R.~Qian, L.~Chen, Q.~Guo, H.~Duan, B.~Wang, L.~Ouyang, et~al.
\newblock Internlm-xcomposer-2.5: A versatile large vision language model supporting long-contextual input and output.
\newblock \emph{arXiv preprint arXiv:2407.03320}, 2024.

\bibitem[Zhao et~al.(2023)Zhao, Yu, Ge, Yang, Wei, Zhou, Sun, Peng, Dong, Han, et~al.]{zhao2023chatspot}
L.~Zhao, E.~Yu, Z.~Ge, J.~Yang, H.~Wei, H.~Zhou, J.~Sun, Y.~Peng, R.~Dong, C.~Han, et~al.
\newblock Chatspot: Bootstrapping multimodal llms via precise referring instruction tuning.
\newblock \emph{arXiv preprint arXiv:2307.09474}, 2023.

\bibitem[Zheng et~al.(2024)Zheng, Zhang, Zhang, Ye, Luo, Feng, and Ma]{zheng2024llamafactory}
Y.~Zheng, R.~Zhang, J.~Zhang, Y.~Ye, Z.~Luo, Z.~Feng, and Y.~Ma.
\newblock Llamafactory: Unified efficient fine-tuning of 100+ language models.
\newblock In \emph{Proceedings of the 62nd Annual Meeting of the Association for Computational Linguistics (Volume 3: System Demonstrations)}, Bangkok, Thailand, 2024. Association for Computational Linguistics.
\newblock URL \url{http://arxiv.org/abs/2403.13372}.

\end{thebibliography}
}



\appendix
\newpage

{\centering \Large \textbf{Towards Better Dental AI: A Multimodal Benchmark and Instruction Dataset for Panoramic X-ray Analysis} \par}

{\centering \Large \textbf{Supplementary Materials} \par}

\addtocontents{toc}{\protect\setcounter{tocdepth}{2}}
\tableofcontents 
\newpage

\section{Related works}
Over the past few years, the evaluation landscape for large vision-language models (LVLMs) has evolved significantly. Benchmarking plays a crucial role in assessing model capabilities, identifying model deficiencies, and guiding future optimization directions. Within the medical domain, existing benchmarks can be classified into two primary categories based on their alignment with the imaging modality coverage: general-purpose benchmarks for broad applicability and specialized benchmarks for discipline-specific evaluation.

There have been numerous efforts toward advancing general medical AI, such as LLaVA-Med~\cite{li2023llavamed}, GMAI-VL~\cite{li2024gmai_model}, MedDr~\cite{he2024meddr}, HealthGPT~\cite{lin2025healthgpt}, and HuatuoGPT~\cite{chen2024huatuogpt}. Alongside these advancements, several general-purpose medical benchmarks spanning diverse imaging modalities and medical domains have been proposed, such as MMMU~\cite{yue2024mmmu}, OminimedVQA~\cite{hu2024omnimedvqa}, MedEval~\cite{he2023medeval}, MedBench~\cite{liu2024medbench}, MMT-Bench~\cite{ying2024mmt}, and GMAI-Bench~\cite{ye2024gmai}. While these medical general-purpose benchmarks enable broader assessments across multiple medical fields, they inevitably fall short in their coverage of imaging modalities and specific medical domains. Therefore, they risk overlooking advancements in specific medical domains not encompassed by the general-purpose benchmarks.

Conversely, specialized benchmarks are concentrated on a particular imaging modality or medical discipline. For instance, PathVQA~\cite{he2020pathvqa}, PathMMU~\cite{sun2024pathmmu}, and PathBench~\footnote{https://smartlab.cse.ust.hk/showcase/PathBench/} focus on pathology analysis, while SLAKE~\cite{liu2021slake}, VQA-RAD~\cite{lau2018dataset}, and RadBench~\cite{wright2016radbench} target radiology understanding. In addition, Eyecare-Bench~\cite{li2025eyecaregpt} and LMOD~\cite{qin2024lmod} assess the overall performance of LVLMs on intelligent ophthalmic diagnosis tasks. CXPMRG-Bench~\cite{wang2024cxpmrg} and M4CXR~\cite{park2024m4cxr} provide an evaluation suite for chest X-ray interpretation tasks. These specialized benchmarks facilitate in-depth evaluations within their respective disciplines, which are better suited to advancing the development of specific medical fields.

Notably, both current medical general-purpose benchmarks and specialized benchmarks overlook the evaluation of MLLMs within oral radiology, particularly in panoramic X-ray understanding—a widely used imaging modality that serves as a primary diagnostic source for assessing oral health. Consequently, there is an urgent demand for more comprehensive and robust benchmarks to address this gap and advance the development of LVLMs in oral healthcare.

\section{MMOral Curation Details}
\subsection{Image and Visual specialists construction}

The method used to identify eligible panoramic X-ray datasets for image curation and visual specialist model construction was adapted from previous studies published in prestigious international peer-reviewed journals~\cite{ye2024gmai, hu2024omnimedvqa, qin2024lmod, liu2024medbench} and globally recognized preprint platforms~\cite{li2025eyecaregpt, he2023medeval}.
Specifically, we collect panoramic X-ray images from two publicly available datasets: the TED3 dataset~\cite{ted3} and the dataset proposed by Hoang Viet Do~\cite{do2024dataset}. 
The TED3 dataset is a large-scale semantic segmentation dataset constructed by aggregating 18 publicly available datasets from various public platforms, including Kaggle, Grand Challenge, and Tianchi. We filter out duplicate images according to the naming rules and ultimately obtain 16,639 unique images. The dataset proposed by Hoang Viet Do~\cite{do2024dataset} is designed for detecting apical periodontitis lesions in panoramic radiographs. This dataset is obtained from the high-quality Dental Treatment Centre, School of Dentistry, Hanoi Medical University, and consists of a total of 3,924 images. Therefore, the final curated dataset comprised 20,563 images, and it exhibits significant diversity across various dental conditions, such as dentate and edentulous dentitions, tooth misalignment, impacted teeth, dental caries, root canal treatment, apical lesions, periodontal bone loss, dental implants, and various types of metallic and non-metallic dental restorations.

The TED3 dataset~\cite{ted3} and the dataset proposed by Hoang Viet Do~\cite{do2024dataset} are licensed under the Apache License 2.0 and CC BY 4.0, respectively. Both licenses allow for the reproduction and distribution of copies of the original datasets with modifications. Therefore, we utilize these two publicly available datasets as the image sources for MMOral.

Subsequently, we leverage visual specialist models to simulate the interpretative process of oral radiology experts. This process aims to recognize as many anatomical structures as possible from the image, covering attributes from teeth to bone structures, historical treatments, and potential existing diseases. Owing to the inherent complexity of anatomical variations and fine-grained pathological cues observed in panoramic X-rays, we build a total of ten specialized visual models that are capable of detecting a total of 49 different anatomical structures. These models are fine-tuned on public datasets specifically related to the panoramic radiograph to extract various visual attributes from the images, and object detection and instance segmentation models are selected for training following the annotation protocols of these public datasets~\cite{hamamci2023dentex,panetta2021tufts,dataset_1,dataset_2,dataset_3,dataset_4,dataset_5,dataset_6,dataset_7,Du2024SVTRv2}. The details of developed visual specialists and the corresponding category list of detected anatomical structures are listed in Table~\ref{tab:visualists details}.

\begin{table}[h!]
\centering
\caption{The details of visual specialists and the corresponding category list of detected anatomical and pathological structures.}
\label{tab:visualists details}
\setlength{\tabcolsep}{5pt}
\small
\begin{tabular}{p{2.0cm}|p{2.8cm}|p{4.6cm}|>{\centering}p{1.7cm}|>{\centering\arraybackslash}p{1.7cm}}
\toprule
\textbf{Dataset Source} & \textbf{Task Type} & \textbf{Category Space} & \textbf{\# Categories} & \textbf{\# Samples} \\ 
\midrule
\cite{hamamci2023dentex,panetta2021tufts,dataset_1} & Object Detection &
1 to 32 tooth numbering following the FDI tooth numbering system & 32 & 2798 \\ 
\midrule
\cite{panetta2021tufts} & Object Detection & 4 Quadrants & 4 & 634 \\ \midrule
\cite{panetta2021tufts} & Object Detection & Caries, Deep Caries, Periapical lesions, Impacted tooth & 4 & 705 \\ \midrule
\cite{dataset_2} & Object Detection & Caries, Filling & 2 & 448 \\ \midrule
\cite{dataset_3} & Object Detection & Caries, Crown, Filling, Implant, Malaligned, Mandibular canal, Missing teeth area, Periapical lesion, Retained root, Root canal treatment, Impacted tooth & 11 & 9206 \\ \midrule
\cite{do2024dataset}& Object Detection & Granuloma, Cyst, Abscess & 3 & 3924 \\ \midrule
\cite{dataset_4} & Instance Segmentation & Caries, Filling & 2 & 448 \\ \midrule
\cite{dataset_5} & Instance Segmentation & Bone loss & 1 & 7986 \\ \midrule
\cite{dataset_6} & Instance Segmentation & Mandibular canal, Maxillary sinus & 2 & 327 \\ \midrule
\cite{dataset_7} & Instance Segmentation & Caries, Crown, Root canal treatment, Badly Decayed, Restoration, Normal & 6 & 1899 \\ \midrule
\cite{Du2024SVTRv2} & Object Detection & Optical character recognition (OCR) & N/A & N/A \\
\bottomrule
\end{tabular}
\end{table}

\subsection{Anatomical structure Extraction}
We construct ten visual specialist models with \textbf{overlapping category spaces} to ensure precise detection of anatomical and pathological structures. For instance, ten structures (e.g., caries, periapical lesion, impacted teeth, missing teeth area, filling, implant, root canal treatment, crown, mandibular canal, and maxillary sinus) are validated by two or more visual specialists.
To process the detected redundant visual elements, we meticulously design a post-processing pipeline that integrates anatomical structures and establishes associations between dental pathological findings, historical treatments, and their corresponding tooth notations based on their spatial relationships. The designed anatomical structure integration and relationship generation pipeline comprises eight systematic steps, as detailed in Algorithm~\ref{alg:xray_pipeline}.

\begin{algorithm}[ht]
\caption{Anatomical Structure Integration and Relationship Generation Pipeline.}
\label{alg:xray_pipeline}
\KwIn{Panoramic X-ray images dataset $I = \{I_1, I_2, \dots, I_n\}$, Expert models $\{M_1, M_2, \dots, M_{10}\}$}
\KwOut{Visual attributes and relationships $\mathcal{A} = \{\mathcal{A}_1, \mathcal{A}_2, \dots, \mathcal{A}_n\}$ for all images}

\ForEach{{panoramic X-ray image $I_i \in I$ }}{

    \tcp{Step 1: Detect imaging timestamp}
    Detect imaging timestamp in $I_i$ (if present) and save as $t_i$\;

    \tcp{Step 2: Detect teeth locations and tooth notations}
    Detect all teeth positions and their corresponding notations following FDI tooth numbering system $\mathcal{T}_i = (\mathcal{P}_i,\mathcal{N}_i)$\;

    \tcp{Step 3: Divide the image into four quadrants}
    Divide the panoramic X-ray image into four quadrants: $Q_i = \{Q_{UR}, Q_{UL}, Q_{LR}, Q_{LL}\}$\;

    \tcp{Step 4: Anatomical identification using specialist models}
    Initialize bounding boxes $\mathcal{B}_i = \emptyset$, $\mathcal{L}_i = \emptyset$, and $\mathcal{S}_i = \emptyset$\;
    \ForEach{$M_j \in \{M_3, ..., M_{10}\}$}{
        $(\mathcal{B}_{j}, \mathcal{C}_{j}, \mathcal{S}_{j}) \gets M_j(I_i)$\;
        $\mathcal{B}_i \gets \mathcal{B}_i \cup \mathcal{B}_{j}$\;
        $\mathcal{L}_i \gets \mathcal{L}_i \cup \mathcal{L}_{j}$\;
        $\mathcal{S}_i \gets \mathcal{S}_i \cup \mathcal{S}_{j}$\;
    }

    \tcp{Step 5: Post-process bounding boxes}
    $\mathcal{B}_i \gets \text{Filter}(\mathcal{B}_i, \tau)$, retaining boxes with $s_k \geq \tau$\;
    $\mathcal{B}_i \gets \text{CategoryIntegration}(\mathcal{B}_i, \mathcal{L}_i)$\;
    $\mathcal{B}_i \gets \text{NMS}(\mathcal{B}_i, \mathcal{S}_i)$\;

    \tcp{Step 6: Assign tooth-related observations to the specific tooth}
    $\mathcal{R}_i \gets  \text{Assign}(\mathcal{T}_i, \mathcal{B}_i, \mathcal{L}_i)$
    
    \tcp{Step 7: Insert domain knowledge rules}
    \If{$\#18 \in \mathcal{N}_i$ and $\#48 \notin \mathcal{N}_i$ \textbf{or} $\#28 \in \mathcal{N}_i$ and $\#38 \notin \mathcal{N}_i$}{
        $\mathcal{O}_i \gets \text{Comment}(\text{``Consider extraction of tooth \#18/\#28''})$\;
    }

    \tcp{Step 8: Generate final visual structures and relationships}
    $\mathcal{A}_i \gets \{t_i, \mathcal{T}_i, \mathcal{Q}_i, \mathcal{B}_i, \mathcal{L}_i, \mathcal{S}_i, \mathcal{R}_i, \mathcal{O}_i\}$\;
}

\Return $\mathcal{A} = \{\mathcal{A}_1, \mathcal{A}_2, \dots, \mathcal{A}_n\}$\;

\end{algorithm}

\subsection{Report generation}

Through extensive consultations with senior dental specialists, we structure the medical report of a panoramic x-ray into three principal sections: Teeth-Specific Observations, Jaw-Specific Observations, and Clinical Summary \& Recommendations. Each section is further subdivided into some subsections, and the specific content covered within each part is outlined in Table~\ref{tab:report_sec}.

\begin{table}[h!]
\caption{The detailed construction of the medical report within three sections and their corresponding content.}
\label{tab:report_sec}
\renewcommand{\arraystretch}{1.2}
\begin{tabular}{>{\Centering}m{3cm}|>{\centering}m{3.3cm}|m{6.6cm}}
\toprule
Section   & Subsection    & \multicolumn{1}{>{\centering}m{6.6cm}}{Content}    \\ 
\midrule
\multirow{3}{*}[-5mm]{\parbox{3cm}{\Centering Teeth-Specific Observations}} & General Condition          & Number of teeth, presence and number of wisdom teeth, and   cases of impaction.      \\ \cline{2-3}
& Pathological   Findings    & Presence of cavities (caries, deep caries) and periapical   lesions (e.g., granuloma, cyst, abscess).  \\
\cline{2-3}
& Historical   Interventions & Past dental treatments, including fillings, crowns, root   canal treatments, and implants.         \\
\hline
\multirow{3}{*}[1mm]{\parbox{3cm}{\Centering Jaw-Specific Observations}} & Bone Architecture          & Assessment of periodontal bone loss.    \\ \cline{2-3}
        & Visible Structures         & Observation of key anatomical features such as mandibular   canals and maxillary sinuses.               \\ \hline
Clinical Summary   \& Recommendations          & -                          & Summary of priority concerns, proposed preventive   measures, and recommended follow-up protocols.  \\
\bottomrule
\end{tabular}
\end{table}

\begin{figure}[!t]
  \centering
  \includegraphics[width=\textwidth]{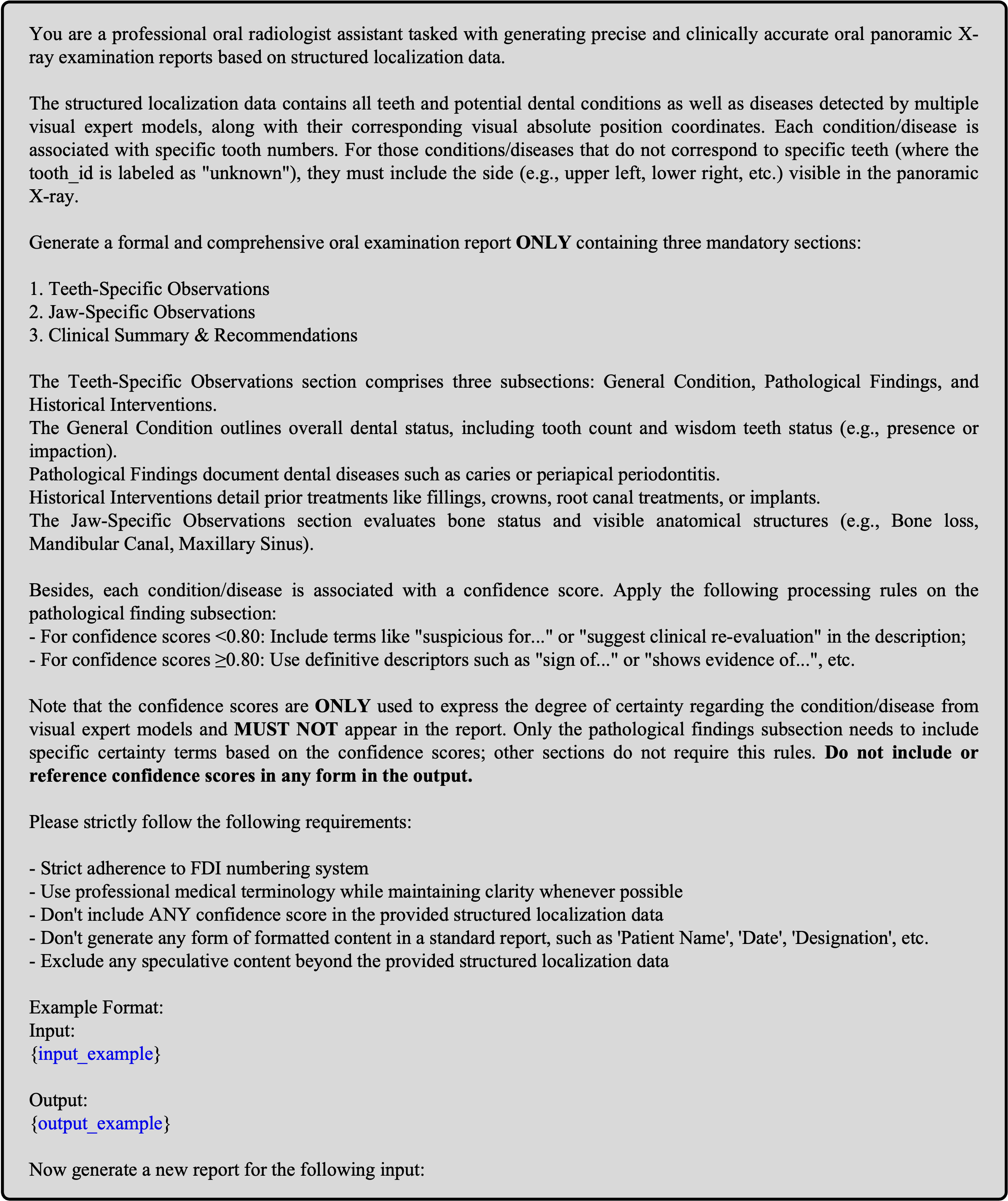}
  \caption{The prompt for DeepSeek-R1-Distill-Llama-70B to generate the medical report of panoramic X-ray images.}
  \label{fig:prompt_report_1st}
\end{figure}

\begin{figure}[!t]
  \centering
  \includegraphics[width=\textwidth]{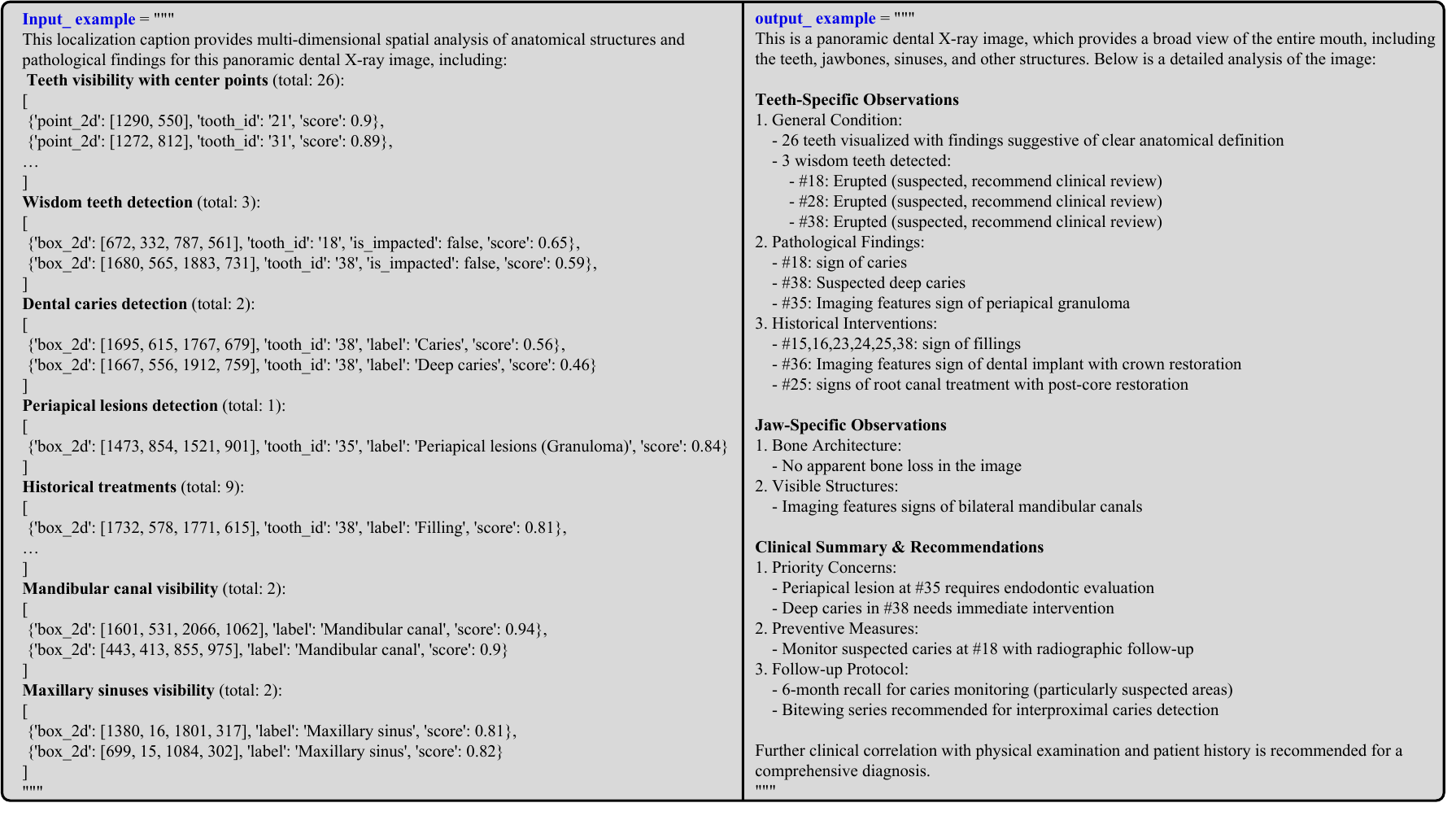}
  \caption{The manually designed in-context examples for medical report generation. Given a grounding caption of panoramic X-rays (left side), the ideal medical report exemplar is shown on the right side. }
  \label{fig:prompt_examplar}
\end{figure}

Thanks to the robust text comprehension and instruction-following capabilities of LLMs, we prompt LLMs to automatically generate medical reports for panoramic X-ray images based on the grounding caption generated by human-designed templates. To ensure precise and structured output, we adopt a two-stage LLM-based generation method to generate the medical report. First, the DeepSeek-R1-Distill-Llama-70B model is selected for medical report generation due to its exceptional performance on text understanding, logical reasoning, and instruction-following abilities. The entire generation process requires approximately 48 hours utilizing 4×NVIDIA A100 80G GPUs.
We meticulously craft a system prompt and manually prepare an example for in-context learning to query the LLM. The details of the system prompt and manually prepared example are shown in the Figure~\ref{fig:prompt_report_1st} and Figure~\ref{fig:prompt_examplar}, respectively.

\begin{figure}[!t]
  \centering
  \includegraphics[width=\textwidth]{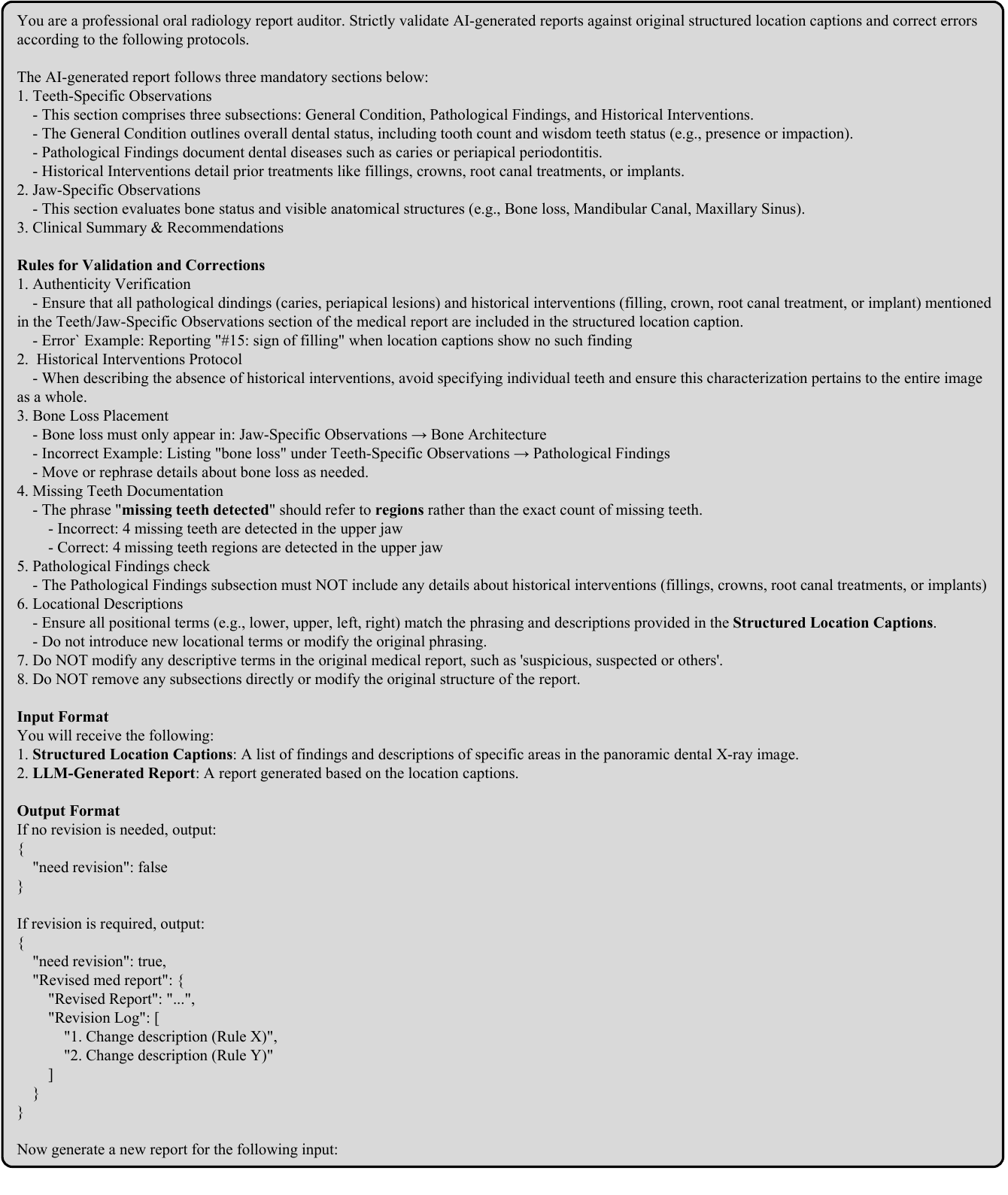}
  \caption{The prompt for GPT-4-Turbo to revise the generated report.
We manually check the generated medical reports from the DeepSeek-R1-Distill-Llama-70B and summarize several rules for validation and correction. We ask the GPT-4-Turbo simultaneously output both revised reports and corresponding revision logs for convenient human verification. }
  \label{fig:prompt_report_2nd}
\end{figure}

\begin{figure}[!t]
  \centering
  \includegraphics[width=\textwidth]{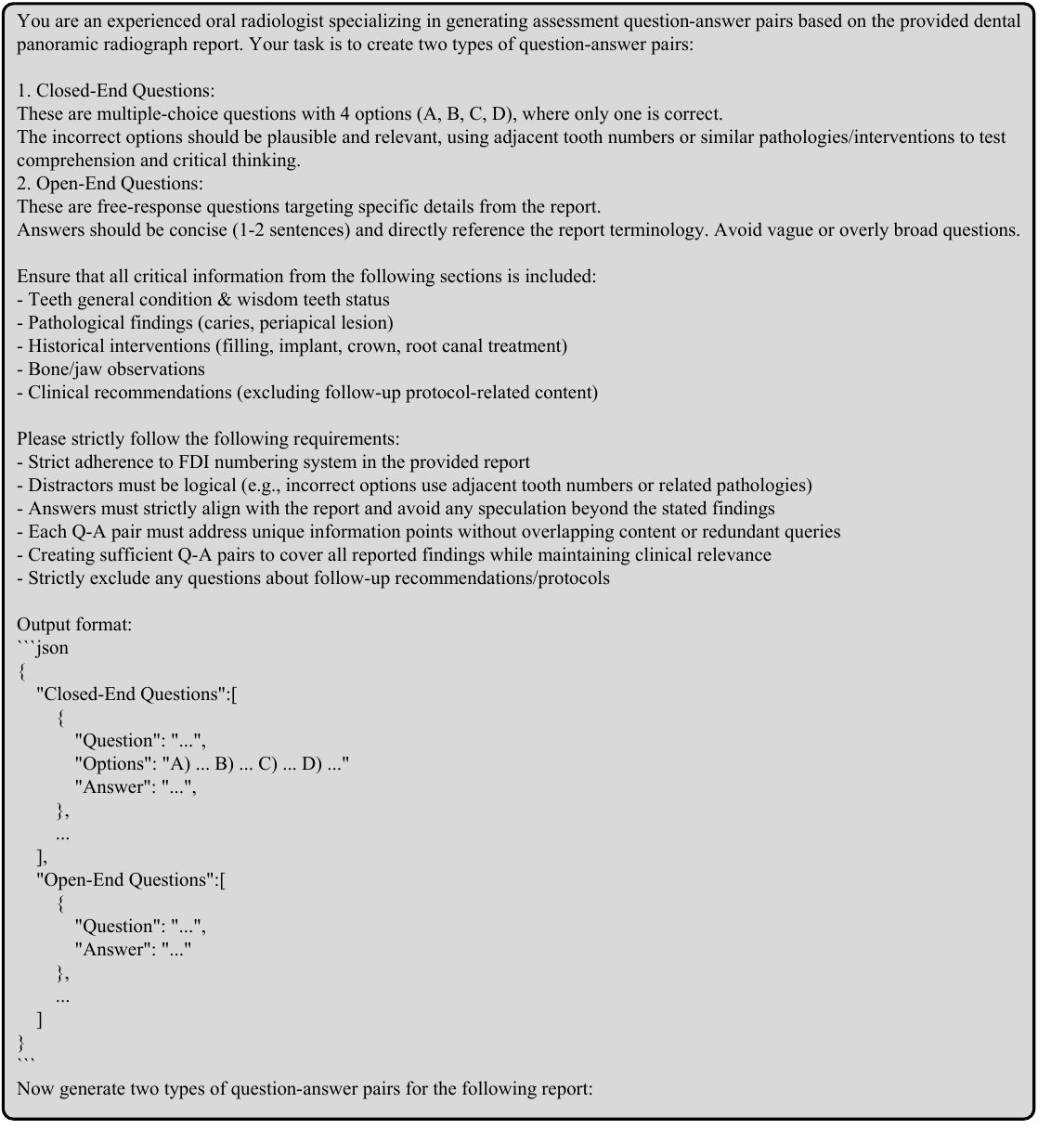}
  \caption{The prompt for GPT-4-Turbo to generate both closed-ended and open-ended question-answering pairs based on the medical report.}
  \label{fig:prompt_vqa}
\end{figure}

\begin{figure}[!t]
  \centering
  \includegraphics[width=\textwidth]{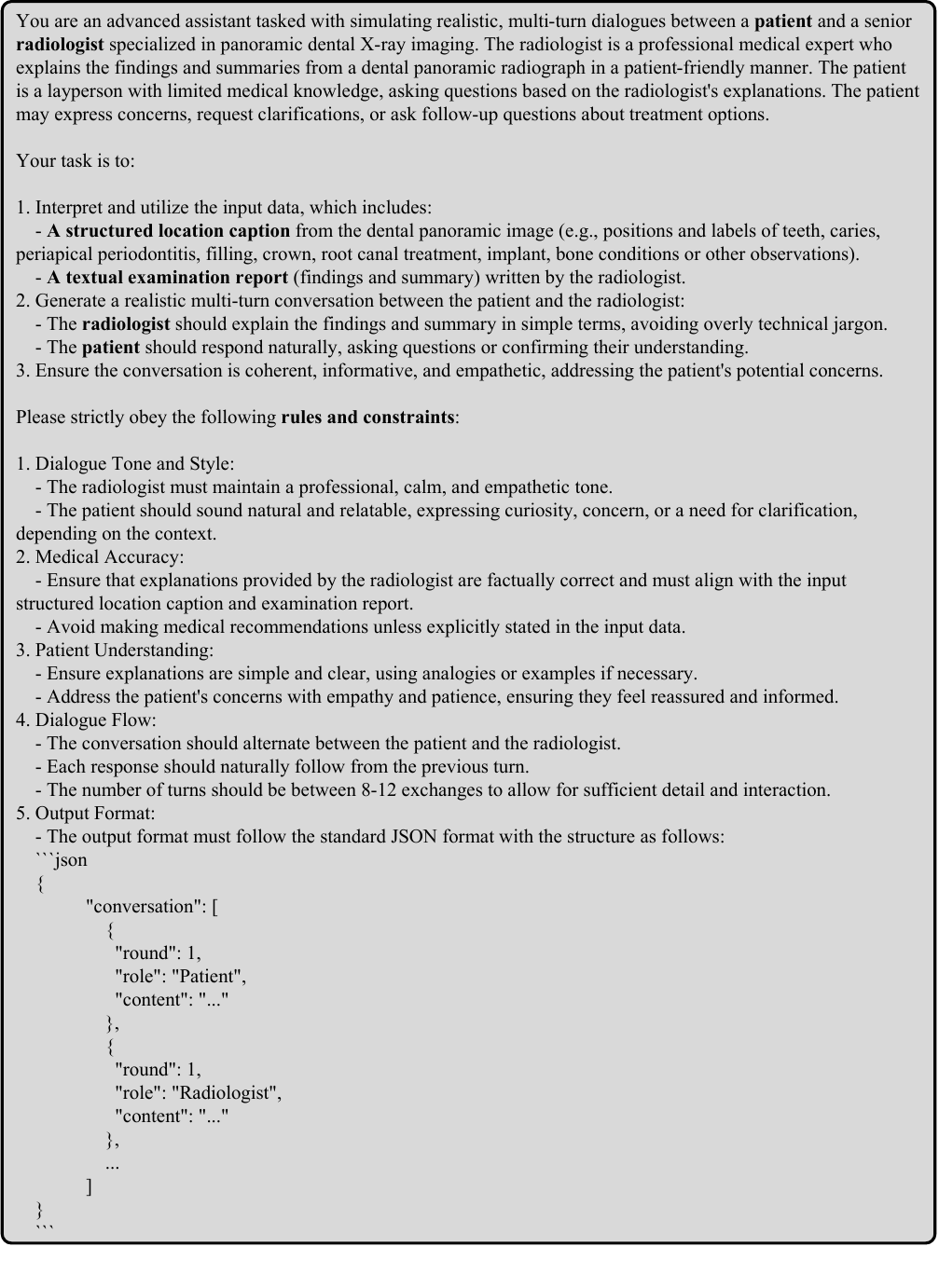}
  \caption{The prompt for GPT-4-Turbo to generate a multi-turn conversation between the assistant and a person asking questions about the panoramic X-ray.}
  \label{fig:prompt_chat}
\end{figure}

During the inspection of generated reports, we identify and summarize several common issues. We hypothesize that these issues are inevitable due to the inherent complexity of the task of generating medical reports based on location captions. This specified task presents significant complexity as it requires LLMs to simultaneously perform multiple cognitive sub-tasks, including text comprehension, organization, classification, structuring, summarization, and extrapolation. As a result, relying solely on the DeepSeek-R1-Distill-Llama-70B model cannot produce high-quality and entirely accurate reports in a single step, despite the model's impressive natural language processing capabilities. To further enhance the quality of reports, we employ GPT-4-Turbo to refine reports generated by DeepSeek. Based on several common issues in the reports, we carefully design rules for validation and corrections and instruct GPT-4-Turbo to simultaneously output both revised reports and corresponding revision logs. By examining these revision logs, we can efficiently identify modified sections of the reports, thereby facilitating quality verification of the revised content. The details of the designed prompt for GPT-4o-Turbo are provided in Figure~\ref{fig:prompt_report_2nd}.

\subsection{Instruction Data Generation}

Based on the refined medical report, we prompt GPT-4-turbo to generate the instruction data, including the visual question-answering data (MMOral-VQA) and the image-grounded conversation data (MMOral-Chat). The designed prompts for these two sub-datasets are shown in Figure~\ref{fig:prompt_vqa} and Figure~\ref{fig:prompt_chat}, respectively. 

\section{MMOral Visualization}
\begin{figure}[!h]
  \centering
  \includegraphics[width=\textwidth]{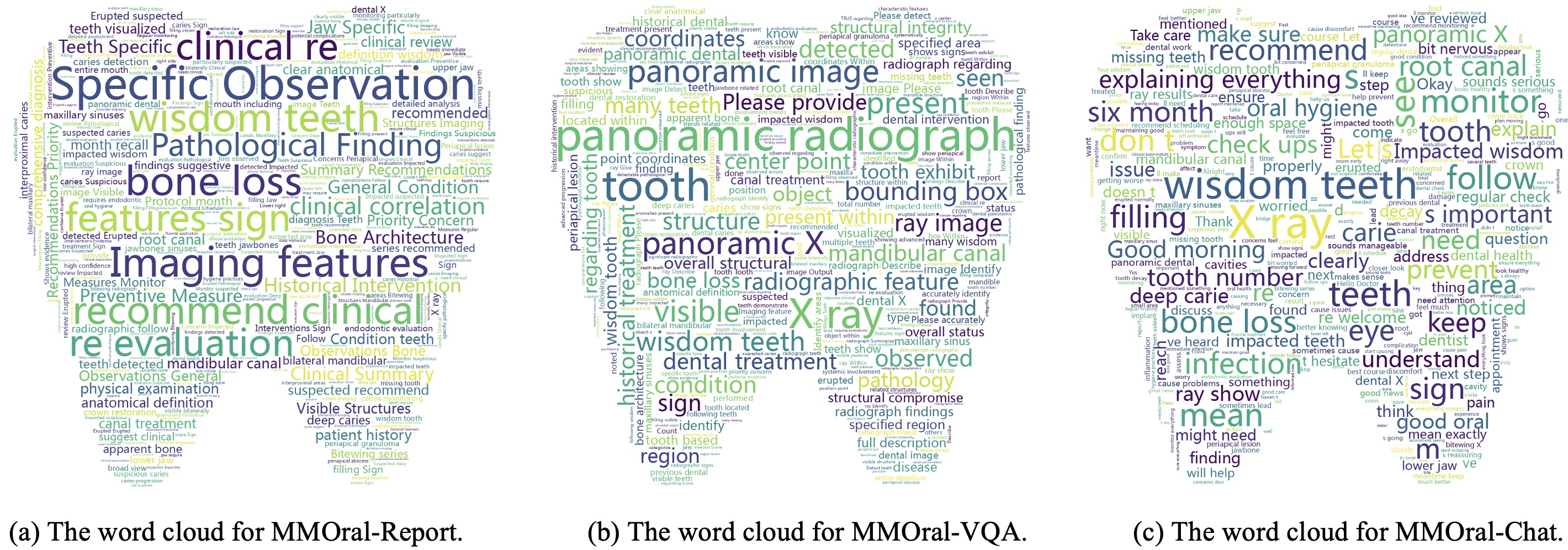}
  \caption{The word cloud maps for MMOral-Report, MMOral-VQA, and MMOral-Chat sub-datasets.}
  \label{fig:cloud_maps}
\end{figure}

\begin{figure}[!h]
  \centering
  \includegraphics[width=\textwidth]{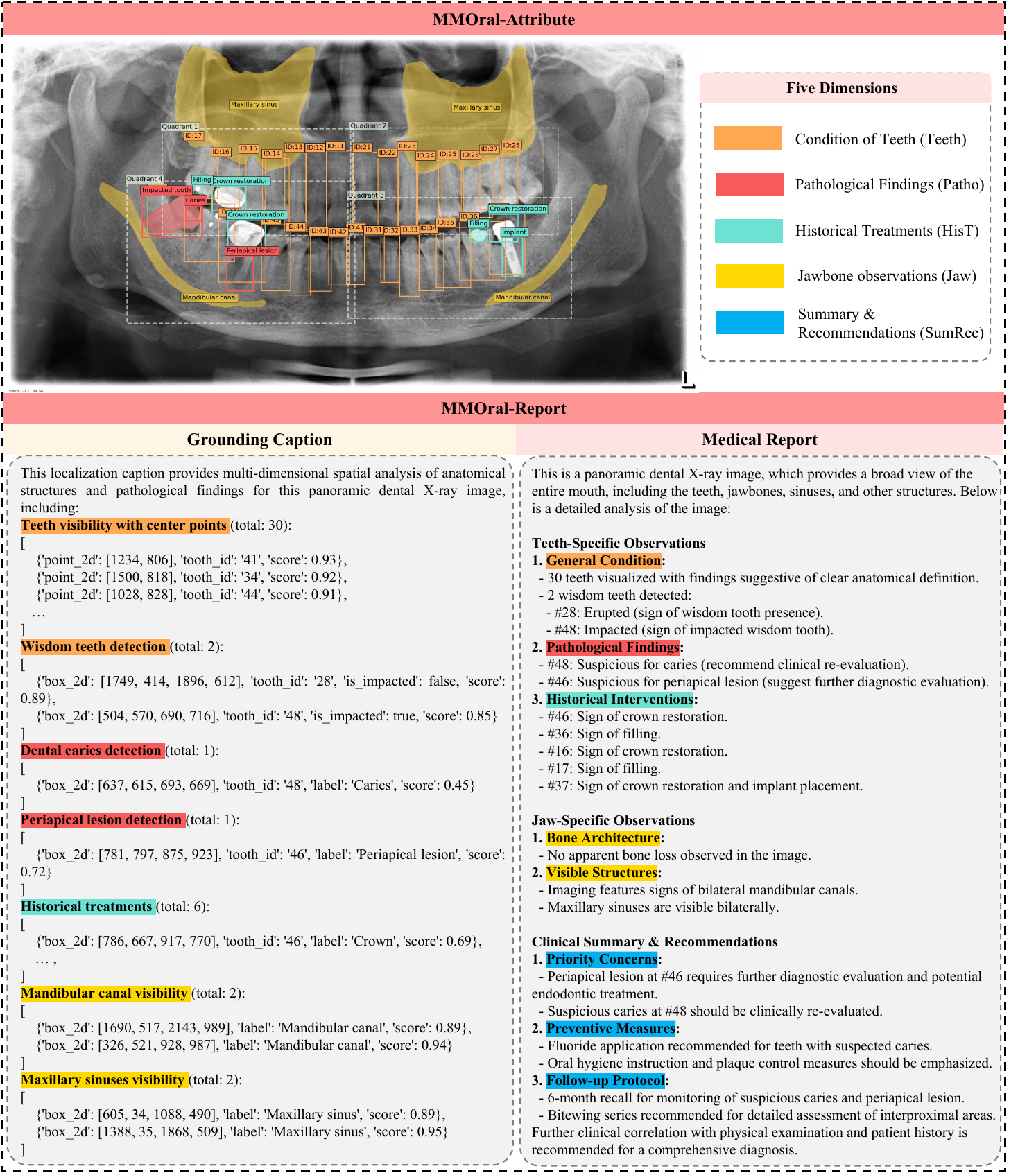}
  \caption{An example of MMOral-Attribute and MMOral-Report.}
  \label{fig:mmoral_report_case_1}
\end{figure}

\begin{figure}[!h]
  \centering
  \includegraphics[width=\textwidth]{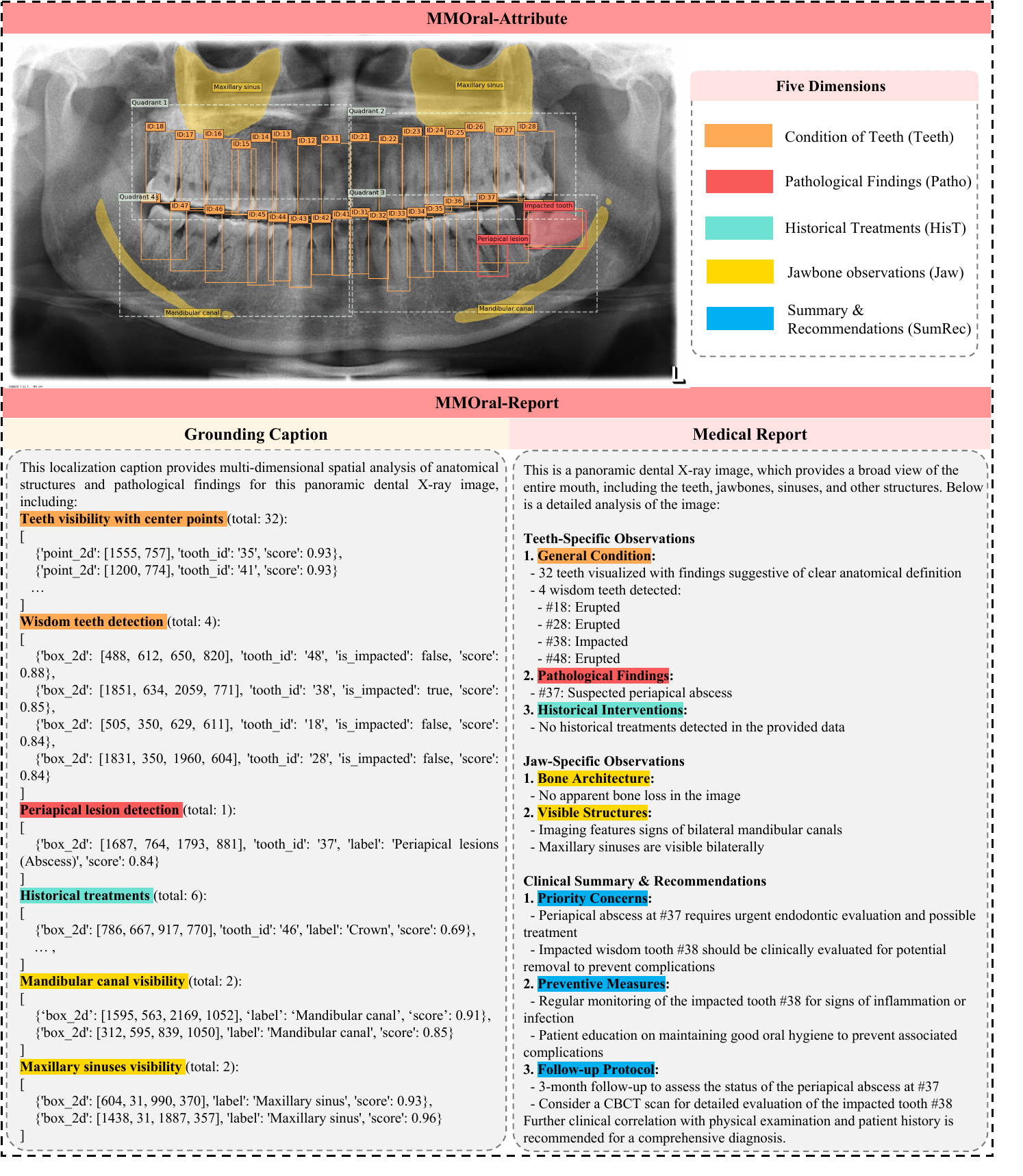}
  \caption{An example of MMOral-Attribute and MMOral-Report.}
  \label{fig:mmoral_report_case_2}
\end{figure}

\begin{figure}[!h]
  \centering
  \includegraphics[width=\textwidth]{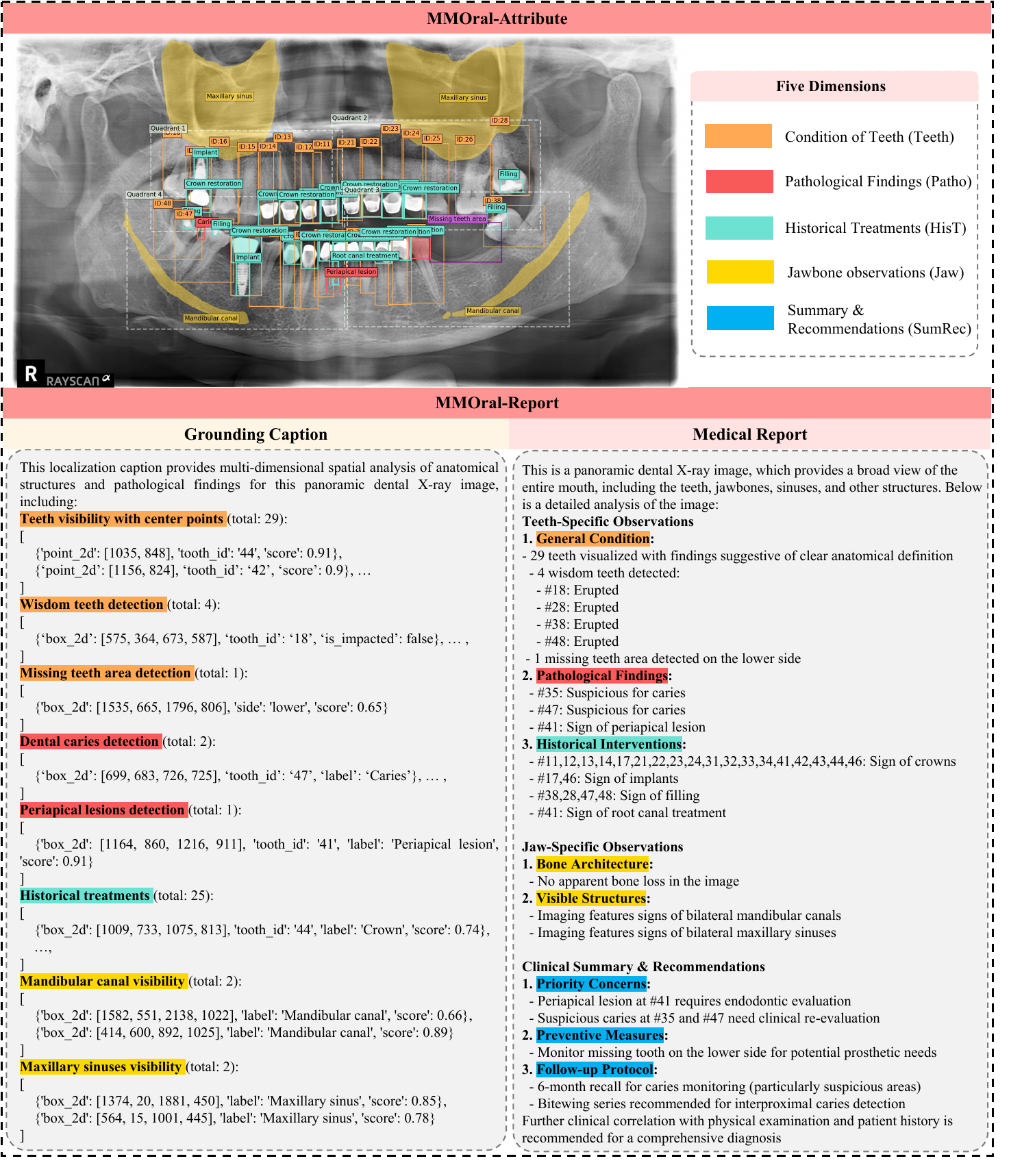}
  \caption{An example of MMOral-Attribute and MMOral-Report.}
  \label{fig:mmoral_report_case_3}
\end{figure}

\begin{figure}[!h]
  \centering
  \includegraphics[width=\textwidth]{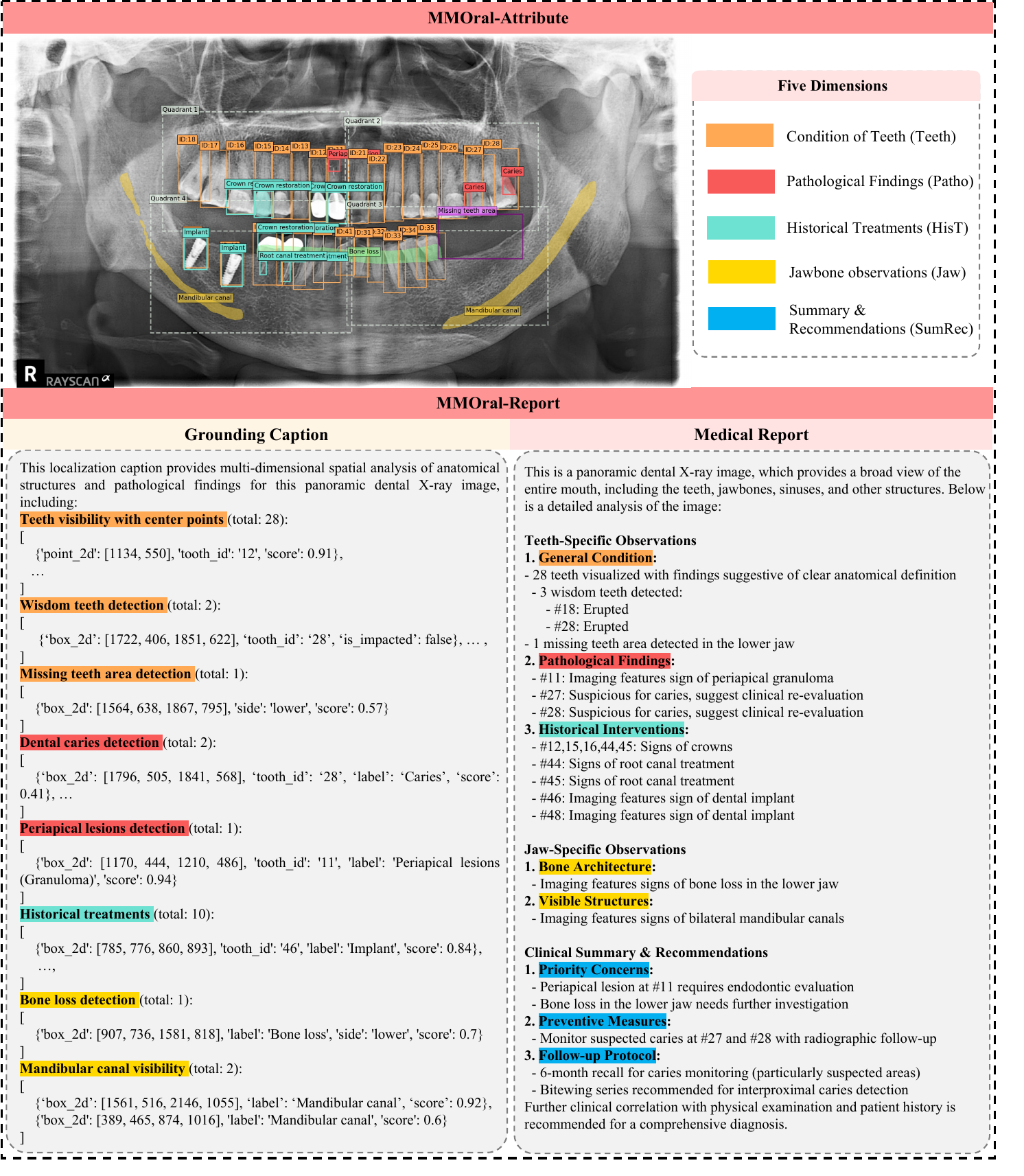}
  \caption{An example of MMOral-Attribute and MMOral-Report.}
  \label{fig:mmoral_report_case_4}
\end{figure}

To provide an intuitive demonstration of the information conveyed in the textual description, we visualize the most frequently occurring words in the MMOral-Report, MMOral-VQA, and MMOral-Chat through word cloud maps, as demonstrated in Figure~\ref{fig:cloud_maps}. Besides, Figures~\ref{fig:mmoral_report_case_1} -~\ref{fig:mmoral_report_case_4} show six examples in MMOral-Attribute and MMOral-Report sub-datasets.

\section{Evaluation}
\subsection{Evaluation Metrics}
In this section, we describe the evaluation process in detail. For closed-ended questions, we adopt the assessment pipeline designed by the CMMMU benchmark. Specifically, we use options as keywords to extract model responses through robust regular expressions, selecting the response with the highest number of options as the final answer. If there
is no valid answer in the model’s response, random selection is performed for multiple-choice questions. We adopt accuracy as the evaluation metric. 

\begin{figure}[!t]
  \centering
  \includegraphics[width=\textwidth]{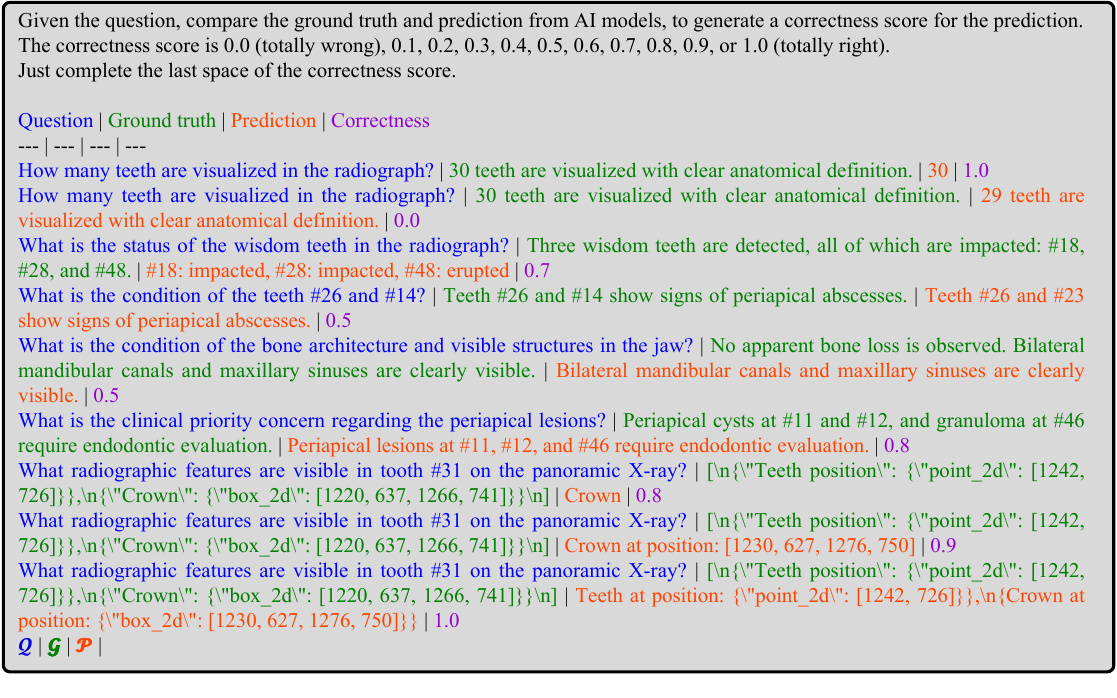}
  \caption{Few-shot prompt for evaluating model predictions using GPT-4-Turbo, where $\mathcal{Q}$ is the question, $\mathcal{Q}$ is the ground truth and $\mathcal{P}$ is the model's prediction for the question. The prompt demonstrates exemplar scoring criteria for diverse open-ended responses.
  Taking the prompt filled with $\mathcal{Q}$,$\mathcal{G}$,and $\mathcal{P}$, GPT-4-Turbo will generate a soft grading score from 0 to 1.}
  \label{fig:prompt_few_shot}
\end{figure}

For open-ended questions, following MM-Vet~\cite{yu2023mmvetv1}and MM-Vetv2~\cite{yu2024mmvetv2}, we construct a few-shot prompt and leverage GPT-4-turbo to assist with the evaluation.
The few-shot prompt is demonstrated in Figure~\ref{fig:prompt_few_shot}. Specifically, our designed prompt includes nine in-context examples with open-ended answers. These examples encompass fully correct responses (i.e., 1.0), entirely incorrect responses (i.e., 0.0), and cases illustrating various types of "partially correct" answers. The LLM-based evaluator enables the evaluation of any style of model's prediction using a unified and consistent metric.

By inputting the prompt, GPT-4 automatically generates scores for each sample based on the input question, ground truth, and model's prediction. Each sample is assigned a score ranging from 0 to 1. The total scores are calculated by 
\begin{equation}\label{eq1}
S = \frac{\sum_{i=1}^N s_i}{N} \times 100\%
\end{equation}
where $s_i$ is the score of sample $i$, and N is the number of samples. The score regarding each sub-category can derived by
\begin{equation}\label{eq1}
S = \frac{\sum s_i}{N_c} \times 100\%, i \in C
\end{equation}
where $C$ is the set of samples belonging to a specific sub-category (e.g., Teeth, Patho, HisT, Jaw, SumRec, Report), and $N_c$ is the number of samples in this set.

The evaluation on MMOral-Bench is conducted using the standard VLMEvalKit framework. We have prepared the automatic evaluation script for MMOral-Bench\footnote{https://github.com/isbrycee/OralGPT}, and it will be supported by the VLMEvalKit framework in the future.  

\subsection{LLM as the Evaluator for Open-ended Questions: A Feasibility Analysis}

\paragraph{Effectiveness.}
To verify the effectiveness of LLM-based evaluation for open-ended tasks, we invite two professional dentists to objectively score the outputs of different LVLMs. We calculate the absolute difference between the evaluators' scores and the human-annotated scores. Specifically, the few-shot prompts designed for LLM-based evaluation are presented to the dentists to determine the evaluation criteria. The two dentists then independently scored the predictions of GPT-4o and HealthGPT-XL32 on 600 cases from the MMOral-Bench open-ended QA task based on these criteria. The absolute differences between human scores and evaluators' scores are shown in Table~\ref{tab:Reliability}, represented as $\Delta$.

Overall, the absolute differences of the ``Overall'' metric given by dentists are approximately 1 point lower than those of the LLM-based evaluation for the predictions of both LVLMs (GPT-4o and HealthGPT-XL32). This indicates that human scoring preferences align with the trends of LLM-based evaluation. However, it also suggests that the dentists' scoring is more stringent compared to the LLM-based evaluation, potentially due to subjective differences in their interpretation of the evaluation criteria provided in the few-shot prompts.

For each subcategory, Dentist A shows smaller differences in scores compared to the LLM-based evaluation for questions in the Teeth, Patho, HisT, and SumRec categories, whereas the differences are larger for the Jaw and Report categories. Although Dentist B exhibits slightly larger differences with LLM-based scoring across all subcategories, their ``overall'' score difference is only 0.37 points. This indicates that LLM-based scoring aligns well with human preferences in reflecting the overall performance of LVLMs on MMOral-Bench open-ended tasks. At the same time, we speculate that the score fluctuations in each subcategory are strongly associated with the subjective perceptions of human evaluators.

\begin{table}[h!]
\centering
\caption{Average absolute differences ($\Delta$) between the evaluation scores of the LLM-based evaluator and the dentist-annotated scores on the open-ended QA task in MMOral-Bench.}
\label{tab:Reliability}
\setlength{\tabcolsep}{5pt}
\small
\begin{tabular}{p{2.8cm}|>{\centering}p{1.7cm}|>{\centering}p{0.9cm}|>{\centering}p{0.8cm}|>{\centering}p{0.8cm}|>{\centering}p{0.7cm}|>{\centering}p{1cm}|>{\centering}p{1cm}|>{\centering\arraybackslash}p{1cm}}
\toprule
\textbf{Model} & \textbf{Evaluators} & \textbf{Teeth} & \textbf{Patho} & \textbf{HisT} & \textbf{Jaw} & \textbf{SumRec} & \textbf{Report} & \textbf{Overall}\\ 
\midrule
 & Dentist A & 30.16 & 27.65 & 40.03 & 67.21 & 33.30 & 33.35 & 35.43  \\ 
GPT-4o~\cite{hurst2024gpt} & GPT-4-Turbo & 31.48 & 26.05 & 37.56 & 57.42 & 30.37 & 42.50 & 37.50  \\ 
 & $\Delta (\downarrow)$ & -1.32 & +1.6 & +2.47 & +9.79 & +2.93 & -9.15 & -2.07  \\ 
\midrule
 & Dentist B & 26.51 & 29.26 & 21.66 & 43.68 & 28.20 & 14.5 & 26.80  \\ 
HealthGPT-XL32~\cite{lin2025healthgpt} & GPT-4-Turbo & 29.80 & 22.16 & 24.11 & 47.82 & 24.77 & 10.00 & 27.17  \\ 
 & $\Delta (\downarrow)$ & -3.29 & +7.1 & -2.45 & -4.14 & +3.43 & +4.50 & -0.37  \\ 
\bottomrule
\end{tabular}
\end{table}

\begin{figure}[!t]
  \centering
  \includegraphics[width=\textwidth]{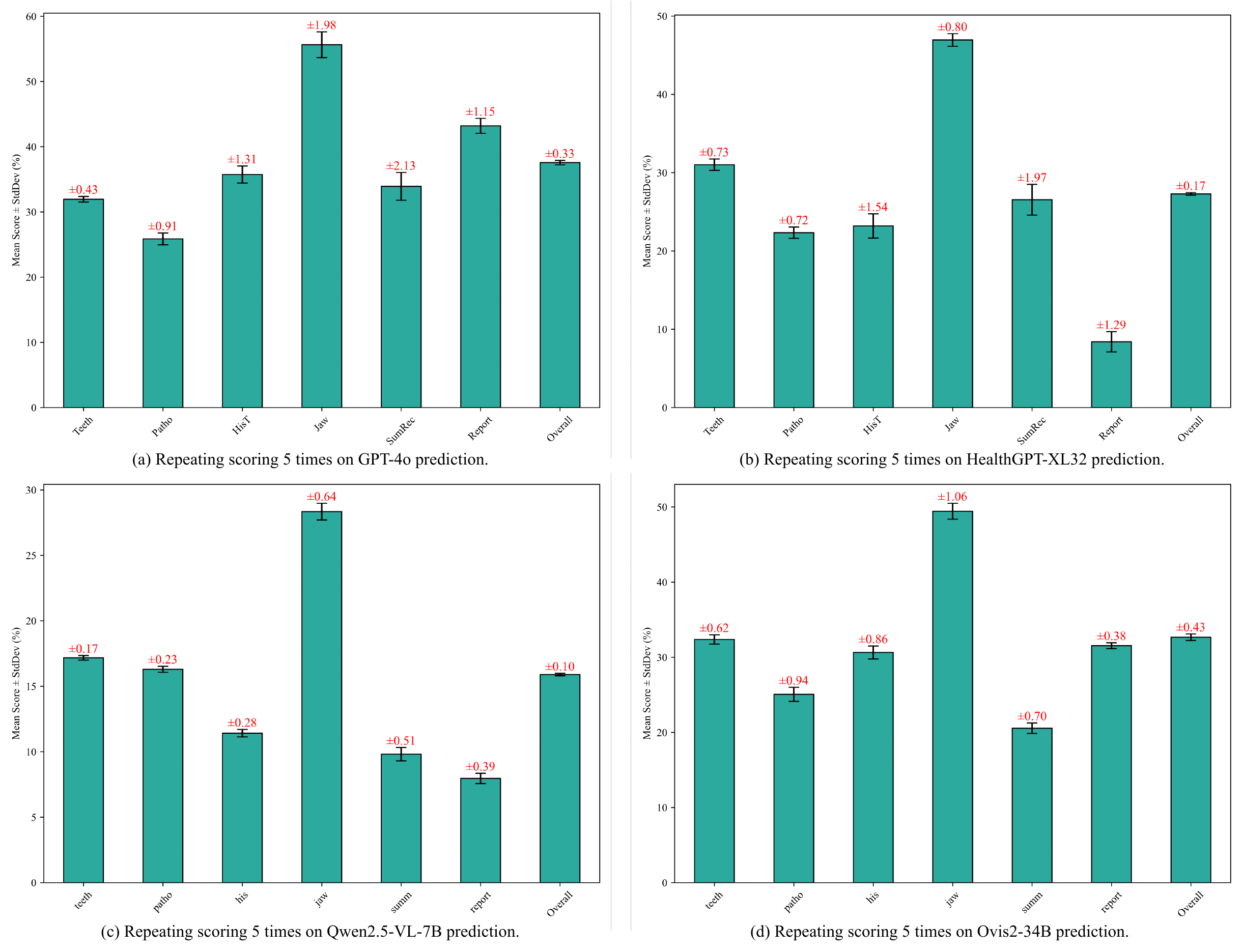}
  \caption{The means and standard deviations of each category on 5 repeated evaluations across four LVLMs' predictions.}
  \label{fig:stability}
\end{figure}

\paragraph{Stability.}
Since using LLMs as evaluators inevitably introduces randomness, even with the temperature hyperparameter set to 0, we conduct multiple repeated experiments to verify the stability of LLMs as evaluators. Specifically, we evaluate the prediction results of GPT-4o~\cite{hurst2024gpt}, HealthGPT-XL32~\cite{lin2025healthgpt}, Qwen2.5-VL-7B~\cite{Qwen-VL}, and Ovis2-34B~\cite{lu2024ovis} on open-ended questions using GPT-4-Turbo~\cite{achiam2023gpt4_turbo} with the same prompt five times. The obtained mean, standard deviation, and coefficient of variation (CV) of the metric ``overall'' are shown in Table~\ref{tab:Stability}. For proprietary models, medical-specific models, and general-purpose LVLMs, the standard deviation of the metric "overall" is within 0.45 when evaluated 5 times using GPT-4-Turbo with our designed few-shot prompt. Specifically, for the prediction results of Ovis2-34B, the standard deviation of the scores is 0.434, while for Qwen2.5-VL-7B, the standard deviation is as low as 0.096. Meanwhile, CV (Coefficient of Variation), as a standardized measure of dispersion of a probability distribution, can be used to assess the stability of scores across multiple experiments. The CV values for the prediction results of these four models, after being scored 5 times, are all around 1\%, which demonstrates the evaluation stability of using LLMs as evaluators. The detailed results across each specific category are demonstrated in Figure~\ref{fig:stability}.

\begin{table}[h!]
\centering
\caption{Stability verification of using LLMs as evaluators: Standard deviation and coefficient of variation (CV) are reported across four LVLMs from five repeated evaluations.}
\label{tab:Stability}
\setlength{\tabcolsep}{5pt}
\small
\begin{tabular}{p{3cm}|>{\centering}p{1.5cm}|>{\centering}p{1.5cm}|>{\centering\arraybackslash}p{4.3cm}}
\toprule
\textbf{Model} & \textbf{Mean} & \textbf{StdDev} & \textbf{CV (Coefficient of Variation, \%)}  \\ 
\midrule
GPT-4o~\cite{hurst2024gpt} & 37.567 & 0.330 & 0.879  \\ 
\midrule
HealthGPT-XL32~\cite{lin2025healthgpt} & 27.284 & 0.172 & 0.631 \\  \midrule
Qwen2.5-VL-7B~\cite{Qwen-VL} & 15.894 & 0.096 & 0.607 \\  \midrule
Ovis2-34B~\cite{lu2024ovis} & 32.671 & 0.434 & 1.329 \\ 
\bottomrule
\end{tabular}
\end{table}

\subsection{Evaluation results}

We conduct zero-shot evaluations across 64 LVLMs on our MMOral-Bench, the results are provided in the Table~\ref{tab:evaluation_results}. 

\begin{table}[t]
\centering
\caption{Results on MMOral-Bench for existing various LVLMs across both closed-ended and open-ended VQA tasks. The best-performing model in each category is highlighted \textbf{in-bold}, while the second-best is \underline{underlined}. 
}
\label{tab:evaluation_results}
\resizebox{\textwidth}{!}{%
\begin{tabular}{@{}l|cccccc|ccccccc|c@{}}
\toprule
\multirow{2}{*}{\textbf{Model}}
  & \multicolumn{6}{c|}{\textbf{Close‐ended VQA}}
  & \multicolumn{7}{c|}{\textbf{Open‐ended VQA}}  
  & \multirow{2}{*}{\textbf{Avg.}} \\
\cline{2-14} 
\rule{0pt}{2.5ex} 
  & \textbf{Teeth} & \textbf{Patho} & \textbf{His} & \textbf{Jaw} & \textbf{Summ} & \textbf{Overall}
  & \textbf{Teeth} & \textbf{Patho} & \textbf{His} & \textbf{Jaw} & \textbf{Summ} & \textbf{Report} & \textbf{Overall} \\
\hline
\rowcolor{mylightred} \multicolumn{15}{l}{\textit{Proprietary LVLMs}} \\ 
\hline
GPT-4o-2024-11-20~\cite{hurst2024gpt}                  & 39.65 & 40.99 & 46.71 & 55.81 & 56.25 & 45.40 & 31.48 & {26.05} & 37.56 & {57.42} & {30.37} & 42.50 & 37.50 & \underline{41.45} \\
GPT-4V~\cite{hurst2024gpt}                  & 37.88 & 39.13 & 48.50 & 51.69 & 58.33 & 43.40 & 27.76 & 13.47 & 33.50 & {58.95} & {30.84} & 45.00 & 34.83 & 39.12 \\
Claude-3-7-Sonnet-20250219~\cite{Claude3S} & 41.24 & 37.27 & 44.31 & 39.70 & 45.83 & 41.40 & 36.93 & 26.65 & 42.39 & {51.09} & {28.04} & \underline{50.00} & 40.67 & 41.04\\
Gemini-2.5-Flash-preview-04-17~\cite{team2023gemini}     & 24.60 & 16.15 & 27.55 & 20.60 & 10.42 & 22.00 & 35.99 & 22.76 & \underline{40.61} & {51.53} & {32.71} & 45.50 & 39.08 & 30.54 \\
Gemini-2.0-Flash~\cite{team2023gemini}         & 37.17 & 35.40 & 44.31 & 46.82 & 58.33 & 41.20 & \underline{37.27} & 26.05 & {40.36} & 52.40 & {35.05} & {49.00} & \underline{40.67} & {40.94} \\
Qwen-Max-VL-2025-04-08~\cite{Qwen-VL} & 18.41 & 11.18 & 27.55 & 32.96 & 47.92 & 22.00 & 10.22 & 7.30 & 11.12 & 22.88 & 6.86 & 27.00 & 14.33 & 18.15 \\
Step-1o-vision-32k \footnotemark & 27.43 & 27.95 & 24.55 & 25.47 & 29.17 & 26.40 &27.93&17.07&25.64&45.85&31.31&26.00&29.08&27.74 \\ 
Step-1o-turbo \footnotemark & 31.86 & 24.22 & 38.92 & 55.81 & 41.67 & 36.00 &33.02&21.56&30.20&51.31&31.31&45.00&36.00&36.00 \\
doubao-1.5-vision-lite-250315~\cite{}  & 21.24 & 14.29 & 28.14 & 27.34 & 33.33 & 21.80 & 32.00 & 12.87 & 29.95 & 60.92 & 28.97 & 48.00 & 36.42 & 29.11 \\
doubao-1.5-vision-pro-250328~\cite{}  & 30.27 & 27.95 & 26.35 & 29.96 & 50.00 & 30.20 & 33.45 & 17.07 & 36.55 & \underline{59.39} & 30.37 & 47.50 & 37.67 & 33.94 \\
Doubao-1-5-thinking-vision-pro-250428 \footnotemark & 26.20 & 27.33 & 23.35 & 19.85 & 31.25 & 24.80 & 34.38 & 25.45 & 39.09 & 56.11 & \underline{39.72} & 49.00 & 40.33 & 32.57 \\ \hline 
\rowcolor{mylightred} \multicolumn{15}{l}{\textit{Open-Source LVLMs}} \\ \hline
Deepseek-VL-7b-chat~\cite{lu2024deepseek}      & 22.65 & 17.39 & 28.74 & 59.93 & 52.08 & 31.20 & 12.75 &  8.16 &  8.40 & 30.00 & 13.14 &  9.10 & 13.42 & 22.31\\
Emu3-chat~\cite{wang2024emu3} & 40.89 & 44.72 & 37.73 & 60.67 & 43.75 & 45.8 & 18.02 &  7.02 &  15.50 & 28.53 & 12.44 &  9.60 & 16.05 & 30.93\\
Qwen2-VL-7B~\cite{wang2024qwen2}              & 30.97 & 27.33 & 26.95 & 35.58 & 29.17 & 30.20 & 18.73 &  9.57 & 13.96 & 40.71 &  4.65 & 15.10 & 18.70 & 22.65\\
Qwen2.5-VL-7B~\cite{bai2025qwen2}            & 24.96 & 21.12 & 27.54 & 37.08 & 35.42 & 27.00 & 17.01 & 16.10 & 11.18 & 29.41 &  9.07 &  8.20 & 15.92 & 21.46\\
Qwen2.5-VL-32B~\cite{bai2025qwen2}           & 25.66 & 26.71 & 23.35 & 22.10 & 14.58 & 24.60 & 16.97 & 10.92 & 11.60 & 23.82 &  9.53 & 13.50 & 15.40 & 20.00 \\
Qwen2.5-VL-72B~\cite{bai2025qwen2}           & 26.55 & 27.95 & 26.35 & 22.47 & 47.92 & 26.80 & 13.05 & 18.44 & 11.66 & 26.88 &  7.44 & 11.50 & 14.77 & 20.79\\
CogVLM-9B~\cite{wang2024cogvlm}                & 26.19 & 21.12 & 34.73 & 32.58 & 64.58 & 29.00 & 25.05 & 14.96 & 26.04 & 42.41 & 13.95 & 14.50 & 24.03 & 26.52\\
CogVLM2-19B~\cite{wang2024cogvlm}                & 33.63 & 31.68 & 34.13 & 38.95 & 60.42 & 35.20 & 26.11 & 17.09 & 26.86 & 49.24 & 18.14 & 24.50 & 27.63 &31.42 \\
GLM-4V-9B~\cite{glm2024chatglm}                & 29.03 & 35.40 & 41.32 & 62.55 & 64.58 & 40.20 & 17.85 &  8.01 & 17.46 & 24.12 & 15.93 & 19.40 & 17.50 & 28.85\\
LLaVA-v1.5-7B~\cite{liu2023visual}            & 20.53 & 17.39 & 28.14 & 26.22 & 47.92 & 23.00 & 12.57 & 13.19 & 10.18 & 18.88 & 17.09 & 11.50 & 13.22 & 18.11\\
LLaVA-v1.5-13B~\cite{liu2023visual}           & 21.24 & 20.50 & 30.54 & 31.46 & 39.58 & 25.20 &  9.71 & 12.13 & 15.62 & 17.29 & 13.60 & 11.40 & 12.23 & 18.72\\ 
LLaVA-NeXT-8B-hf~\cite{liu2024llavanext}         & 36.81 & 33.54 & 41.32 & 51.31 & 64.58 & 41.60 & 15.87 &  5.39 &  9.17 & 27.67 & 14.42 & 24.50 & 16.68 & 29.14\\
LLaVA-NeXT-13B-hf~\cite{liu2024llavanext}         & 30.09 & 32.92 & 30.54 & 38.20 & 60.42 & 33.80 & 14.48 &  10.28 &  9.23 & 22.41 & 14.30 & 21.30 & 15.43 & 24.62\\
LLaVA-OneVision~\cite{li2024llava}          & 14.51 & 18.01 & 35.33 & 42.70 & 31.25 & 24.40 & 22.68 & 13.48 & 17.75 & 38.35 & 18.72 & 11.20 & 20.93 & 22.67\\
LLaMA-3.2-Vision-11B-Instruct~\cite{llama3_2}     &27.96&21.12&40.72&50.94&50.00&34.00     & 24.81 & 20.71 & 25.03 & 33.65 & 17.56 & 19.50 & 23.97 & 28.99\\
Cambrian-8B~\cite{tong2024cambrian}              & 27.26 & 23.60 & 44.31 & 64.04 & 50.00 & 37.00 & 25.96 & 10.78 & 22.37 & 39.41 & 20.58 & 19.50 & 24.28 & 30.64\\
Cambrian-13B~\cite{tong2024cambrian}             & 32.57 & 27.33 & 43.11 & 70.41 & 39.58 & 41.20 & 24.29 & 12.55 & 20.71 & 41.18 & 17.21 & 14.70 & 22.67 & 31.94\\
Cambrian-34B~\cite{tong2024cambrian}             & 34.87 & 34.16 & 44.31 & 70.04 & 60.42 & 44.40 & 33.10 & 21.42 & 31.83 & 48.24 & 13.60 & 16.00 & 29.63 & 37.02\\
Phi-3-Vision-128K-Instruct~\cite{abdin2024phi3} &30.62 &32.92 &40.72 &44.57 &64.58 &37.40 &20.18 &17.80 &16.57 &46.35 &20.35 &8.60 &20.93 &29.17 \\
Phi-$3.5$-Vision-Instruct~\cite{abdin2024phi3} &27.08 &22.36 &40.72 &38.95 & 56.25 &31.60 &28.84 &25.25 &24.08 &42.94 &24.42 &9.50 &25.93 &28.77 \\
Phi-4-multimodal-instruct~\cite{abdin2024phi4}  &36.28 &36.65 &49.10 &60.30 &54.17 &43.60 &25.52 &20.14 &27.69 &43.29 &13.84 &12.80 &24.57 &34.09 \\
InternVL$2.5$-8B~\cite{chen2024expanding}         & 32.21 & 27.95 & 24.55 & 34.83 & 37.50 & 30.20 & 27.85 & 20.50 & 26.75 & 40.06 & 22.33 & 25.00 & 27.50 & 28.85\\
InternVL3-8B~\cite{chen2024expanding}         & 22.12 &	24.22 &	21.56 &	35.96 &	31.25 &	25.40 & 27.89 &	14.47&	25.09&	37.29&	18.14&	23.80&	25.82 & 25.61\\
InternVL3-14B~\cite{chen2024expanding}         & 31.50&	29.81&	26.35&	35.58	&27.08&	31.20& 26.22&	17.59&	27.16&	44.06&	22.91&	36.40&	29.27 & 30.24\\
InternVL3-38B~\cite{chen2024expanding}         & 28.67	&21.12	&25.75	&39.33	&31.25	&28.40 &33.69	&22.41	&29.70	&46.11	&20.23	&42.90	&34.15 & 31.28\\
Chameleon-7B~\cite{lu2023chameleon}             & 32.57 & 44.10 & 37.13 & 29.59 & 52.08 & 35.80 &  6.02 &  6.10 &  9.35 &  9.71 &  5.35 &  8.40 &  7.27 & 21.54\\
PaliGemma-3B~\cite{beyer2024paligemma}             & 26.02 & 24.22 & 42.52 & 47.57 & 35.42 & 33.20 &  8.20 &  9.65 &  8.70 & 13.29 &  6.16 &  0.60 &  7.78 & 20.49\\
MiniCPM-V$2.6$~\cite{yao2024minicpm}           & 28.50 & 25.47 & 29.94 & 34.08 & 14.58 & 28.00 & 27.38 & 17.38 & 24.38 & 49.76 & 15.93 & 27.90 & 28.42 & 28.21\\
MiniCPM-O$2.6$~\cite{yao2024minicpm}           & 30.27 & 23.60 & 24.55 & 36.33 & 14.58 & 30.27 & 29.20 & 17.38 & 24.38 & 49.76 & 15.93 & 27.90 & 28.42 & 28.21\\
Kosmos-2~\cite{peng2023kosmos}                 & 15.75 & 18.01 & 28.14 & 10.11 & 25.00 & 17.40 & 13.58 & 10.71 & 11.18 & 19.76 &  8.49 &  3.40 & 11.87 & 14.64\\
Kimi-VL-A3B-Instruct~\cite{team2025kimi}                 & 44.60 & 42.24 & 38.32 & 70.79 & 66.67 & 49.8 & 28.20	& 17.94	& 32.49	& 53.53	& 19.95	& 25.60	& 30.00 & 39.90\\

Ovis1.5-8B~\cite{lu2024ovis}        & 25.31 & 29.19 & 37.13 & 70.79 & 41.67 & 37.40 & 31.19 & 26.52 & 30.41 & 48.29 & 19.06 & 19.80 & 30.20 & 33.80\\
Ovis2-8B~\cite{lu2024ovis}                 & 40.18 & 47.83 & 44.91 & \underline{75.66} & 70.83 & 50.80 & 28.70 & 26.17 & 26.63 & 53.06 & 20.81 & 30.72 & 30.70 & 40.75\\

mPLUG-Owl2-7B~\cite{ye2024mplugowl2}            & 23.54 & 20.50 & 43.11 & 44.57 & 54.17 & 31.80 & 11.82 & 16.81 &  9.94 & 24.76 & 18.72 & 14.50 & 14.80 & 23.30\\
mPLUG-Owl3-7B~\cite{ye2024mplug}            & 34.16 & 32.30 & 36.53 & 71.91 & 62.50 & 42.80 & 12.50 &  8.44 &  8.52 & 30.59 &  3.26 & 13.90 & 13.67 & 28.24\\
Gemma3-12B~\cite{team2025gemma}               & 24.78 & 19.88 & 31.74 & 34.08 & 29.17 & 26.60 & 25.21 & 20.00 & 20.65 & 26.88 & 22.33 & 33.20 & 25.32 & 25.96\\
XComposer2-VL-7B~\cite{zhang2024internlm}         & 25.49 & 26.71 & 21.56 & 13.86 & 45.83 & 23.20 &  6.52 & 11.01 & 15.00 &  7.67 &  2.10 &  8.53 &  8.99 & 16.10\\
Molmo-7B-O-0924~\cite{deitke2024molmo}          & 25.31 & 21.74 & 26.35 & 25.09 & 12.50 & 24.40 & 11.49 &  6.74 &  8.52 & 10.94 &  3.95 &  6.00 &  9.02 & 16.71\\
Molmo-72B-0924~\cite{deitke2024molmo}           & 28.85 & 14.91 & 29.94 & 26.59 & 22.92 & 25.60 &  9.25 &  6.31 &  3.49 & 12.65 &  5.00 &  9.20 &  8.23 & 16.92\\
Yi-VL-6B~\cite{young2024yi} &28.32 &39.75 &40.11 &58.80 &{77.08} &40.80 &17.43 &13.62 &21.12 &32.24 &16.74 &8.60 &18.00 &29.40 \\
Yi-VL-34B~\cite{young2024yi} &36.81 &36.64 &43.11 &41.20 &70.83 &40.20 &24.97 &23.40 &20.59 &39.35 &15.23 &9.90 &22.98 &31.59 \\

Qwen-QVQ-72B~\cite{qvq-72b-preview} & \textbf{48.67} & \underline{49.07} & \textbf{59.28} & {74.53} & 72.92 & \underline{56.60} &20.37&12.28&16.75&41.05&22.90&24.5&22.75&39.68 \\ 
Ovis2-34B~\cite{lu2024ovis}                & \underline{45.84} & \textbf{51.55} & \underline{53.89} & \textbf{79.40} & \textbf{79.17} & \textbf{56.80} & 32.48 & 24.33 & 31.60 & 50.88 & 21.05 & 31.70 & 33.02 & \textbf{44.91}\\
Kimi-VL-A3B-Thinking~\cite{team2025kimi} & 25.84	& 27.33	& 25.75	& 29.96	& 27.08	& 26.80 & \textbf{52.53}	& \textbf{37.66}	& \textbf{53.79}	& \textbf{68.59}	& \textbf{50.93}	& \textbf{61.50}	& \textbf{54.55} & 40.68\\
\hline
\rowcolor{mylightred} \multicolumn{15}{l}{\textit{Medical Specific LVLMs}} \\ \hline
LLaVA-Med~\cite{li2023llavamed}                & 25.49 & 26.71 & 21.56 & 13.86 & 45.83 & 23.20 & 23.23 & 18.75 & 11.36 & 32.82 & 26.28 &  5.30 & 19.60 & 21.40\\
LLaVA-NeXT-Med~\cite{guo2025llava} &19.12 &22.36 &28.74 &31.09 &43.75 &24.80 &2.51 &1.06 &0.24 &0.94 &2.21 &0 &1.38 &13.09 \\
HuatuoGPT-V-7B~\cite{chen2024huatuogpt}           & 25.31 & 21.74 & 26.35 & 25.09 & 12.50 & 24.40 &  0.00 &  0.00 &  0.30 &  0.29 &  0.00 &  0.70 &  0.20 & 12.21\\
HuatuoGPT-V-34B~\cite{chen2024huatuogpt}          & 28.85 & 14.91 & 29.94 & 26.59 & 22.92 & 25.60 & 32.62 & 18.65 & 28.05 & 53.12 & 18.60 & 15.40 & 29.48 & 27.54 \\
HealthGPT-M3~\cite{lin2025healthgpt} & 45.84 & 44.10 & 46.71 & 72.66 & 75.00 & 52.80 & 30.39 & 18.86  & 26.90 & 46.07 & 21.03 & 19.50 & 28.92 & 40.86 \\
HealthGPT-XL32~\cite{lin2025healthgpt} & 39.65 & 44.10 & 51.50 & 76.41 & \underline{79.17} & 52.00 & 29.80 & 22.16  & 24.11 & 47.82 & 24.77 & 10.00 & 27.17 & 39.59 \\
MedGemma-4B~\cite{team2025gemma} & 34.51 & 29.81 & 43.11 & 59.55 & 39.58 & 40.00 & 29.69 & 11.26  & 28.12 & 41.88 & 22.62 & 39.40 & 30.45 & 35.23 \\
MedVLM-R1~\cite{pan2025medvlm} &28.67 &31.68 &37.72 &65.17 &47.92 &38.60 &22.58 &12.28 &21.57 &40.61 &21.96 &24.50 &24.58 & 31.59\\
MedDr~\cite{he2024meddr} &36.46 &36.02 &41.92 &73.03 &64.58 &46.00 &{27.50} &\underline{28.14} &30.20 &49.17 &26.17 &7.50 &26.17 & 36.09\\
\bottomrule
\end{tabular}%
}
\end{table}
\footnotetext[3]{https://platform.stepfun.com/}
\footnotetext[4]{https://www.volcengine.com/product/doubao/} 
\footnotetext[5]{https://github.com/google-health/medgemma/}

\section{Efficacy Validation of MMOral Instruction Data}
We implement supervised fine-tuning (SFT) on two LVLMs of different scales, Qwen2.5-VL-7B~\cite{Qwen-VL} and LLaVA-Next-13B-hf~\cite{liu2024llavanext}, using our MMOral instruction data, and the results are presented in Table~\ref{tab:sft_results}. We use the LLaMA-Factory~\cite{zheng2024llamafactory} framework to perform SFT for one epoch while maintaining other default hyperparameters.
When fine-tuning Qwen2.5-VL-7B and LLaVA-Next-13B-hf with MMOral-Report, MMOral-VQA, and MMOral-Chat together, the average scores improve by 24.73\% and 18.42\%, respectively. This clearly demonstrates the effectiveness of the MMOral instruction data and its potential value in advancing digital AI applications in the field of oral healthcare.

\begin{table}[!htbp]
\centering
\caption{The effectiveness verification of MMOral instruction data by supervised fine-tuning.}
\label{tab:sft_results}
\resizebox{\textwidth}{!}{%
\begin{tabular}{@{}l*{17}{c}@{}}
\toprule
\multirow{2}{*}{\textbf{Model}}
  & \multicolumn{3}{c}{\textbf{SFT}} 
  & \multicolumn{6}{c}{\textbf{Close‐ended VQA}}
  & \multicolumn{7}{c}{\textbf{Open‐ended VQA}} 
  & \multirow{2}{*}{\textbf{Avg.}} \\
\cmidrule(lr){2-4} \cmidrule(lr){5-10} \cmidrule(lr){11-17}
  & \textbf{Report} & \textbf{VQA} & \textbf{Chat} & \textbf{Teeth} & \textbf{Patho} & \textbf{HisT} & \textbf{Jaw} & \textbf{SumRec} & \textbf{Overall}
  & \textbf{Teeth} & \textbf{Patho} & \textbf{HisT} & \textbf{Jaw} & \textbf{SumRec} & \textbf{Report} & \textbf{Overall} \\
  
\midrule
Qwen2.5-VL-7B~\cite{bai2025qwen2}   &\ding{55} &\ding{55} &\ding{55}    & 24.96 & 21.12 & 27.54 & 37.08 & 35.42 & 27.00 & 17.01 & 16.10 & 11.18 & 29.41 &  9.07 &  8.20 & 15.92 & 21.46\\

\midrule

Qwen2.5-VL-7B~\cite{bai2025qwen2} &\ding{51} &\ding{55} &\ding{55} &26.90 &27.33 &26.35 &45.32 &37.50 &31.00 &27.82 &15.82 &25.92 &63.76 &22.33 &38.00 &32.62 &31.81  \\ 
Qwen2.5-VL-7B~\cite{bai2025qwen2} &\ding{55} &\ding{51} &\ding{55} & 39.12 & 36.65 & 37.73 & 62.92 & 43.75 & 43.60 & 36.22 & 31.92 & 32.49 & 78.47 & 40.93 & 4.30 & 35.73 & 39.67 \\ 

Qwen2.5-VL-7B~\cite{bai2025qwen2} &\ding{51} &\ding{51} &\ding{55} &\textbf{43.19} &\textbf{40.99} &\textbf{43.11} &\textbf{63.60} &37.50 &\textbf{46.20} & 39.85 & 32.41 &35.20 &\textbf{78.06} &36.98 &36.80 &42.85 &44.53  \\ 
Qwen2.5-VL-7B~\cite{bai2025qwen2} &\ding{51} &\ding{51} &\ding{51} & 37.17 &30.43 &38.32 &52.81 &\textbf{45.83} &39.60 & \textbf{55.45} & \textbf{33.40} &\textbf{45.74} &74.47 &\textbf{45.17} &\textbf{50.50} &\textbf{52.77} &\textbf{46.19}  \\ 

\midrule
LLaVA-NeXT-13B-hf~\cite{liu2024llavanext}   &\ding{55} &\ding{55} &\ding{55}       & 30.09 & 32.92 & 30.54 & 38.20 & 60.42 & 33.80 & 14.48 &  10.28 &  9.23 & 22.41 & 14.30 & 21.30 & 15.43 & 24.62\\
\midrule
LLaVA-NeXT-13B-hf~\cite{liu2024llavanext} &\ding{51} &\ding{55} &\ding{55} &27.79 &25.47 &31.74 &\textbf{57.68} &\textbf{62.50} &35.00 &17.23 &10.21 &11.18 &25.41 &16.51 &19.8 &17.07 &26.03  \\
LLaVA-NeXT-13B-hf~\cite{liu2024llavanext} &\ding{55} &\ding{51} &\ding{55} &39.12 &40.37 &31.14 &48.69 &29.17 &39.80 &41.56 &\textbf{26.95} &32.78 &81.94 &\textbf{37.21} &6.20 &37.98 &38.89  \\
LLaVA-NeXT-13B-hf~\cite{liu2024llavanext} &\ding{51} &\ding{51} &\ding{55} &43.19 &48.45 &\textbf{38.92} &45.32 &35.42 &43.20 &40.33 &26.38 &36.57 &76.76 &36.98 &29.30 &41.10 &42.15  \\
LLaVA-NeXT-13B-hf~\cite{liu2024llavanext} &\ding{51} &\ding{51} &\ding{51} &\textbf{59.15} &\textbf{57.14} &37.13 &32.21 &\textbf{35.42} &\textbf{48.20} &\textbf{42.11} &23.48 &\textbf{41.01} &\textbf{88.47} &33.37 &\textbf{33.60} &\textbf{44.18} &\textbf{46.19}  \\

\bottomrule
\end{tabular}%
}
\end{table}

\section{Limitations} 
The ground truth reports generated in this project were based on the ground truth labels provided by previous studies published in esteemed international peer-reviewed journals~\cite{panetta2021tufts, do2024dataset} and globally recognized preprint and dataset platforms~\cite{dataset_1, dataset_2, dataset_3, dataset_4,dataset_5,dataset_6,dataset_7,hamamci2023dentex}. However, the potential inaccuracies in the provided ground truth labels cannot be entirely neglected, as their accuracy has not been validated by independent third-party organizations. Given the considerably large volume of annotated data used to construct the visual specialist model (e.g., 10 datasets comprising 28,375 images), it is not practical for a single centre to manually verify the accuracy of these ground truth labels in a short period of time.

Nevertheless, we have utilized multiple visual specialist models with overlapping category spaces to identify the same anatomical and pathological structures, thereby minimizing the risk of potential inaccuracies in the generated ground truth reports. For instance, ten structures (e.g., caries, periapical lesion, impacted teeth, missing teeth area, filling, implant, root canal treatment, crown, mandibular canal, and maxillary sinus) are validated by two or more visual specialist models, with the final results obtained through post-processing. In addition, we have adopted a two-stage LLM-based scheme of generation followed by correction to ensure the report quality. First, we utilize the DeepSeek-R1-Distill-Llama-70B to generate preliminary reports. Subsequently, through manual review of these preliminary reports, we identify common errors and summarize them into eight key rules (see in Figure~\ref{fig:prompt_report_2nd}) for prompting GPT-4-turbo to revise reports. An analysis of the revision logs shows that 95.45\% of the reports are successfully corrected, leading to a significant improvement in their overall quality. Future efforts should focus on third-party validation of ground truth accuracy in these public datasets to further ensure their reliability.

\section{Experiments Compute Resources}
The experimental section of this paper, involving the construction of the MMOral dataset and the evaluation of MMOral-Bench, requires the use of a proprietary LLMs API. The total cost of the experiments is approximately 1000 USD, with around 600 USD spent on building the MMOral dataset and about 400 USD on evaluating existing LVLMs. Furthermore, the SFT experiments conducted in this paper are performed on 4×NVIDIA A100 80G GPUs.


\section{Case Study}
\begin{figure}[!t]
  \centering
  \includegraphics[width=\textwidth]{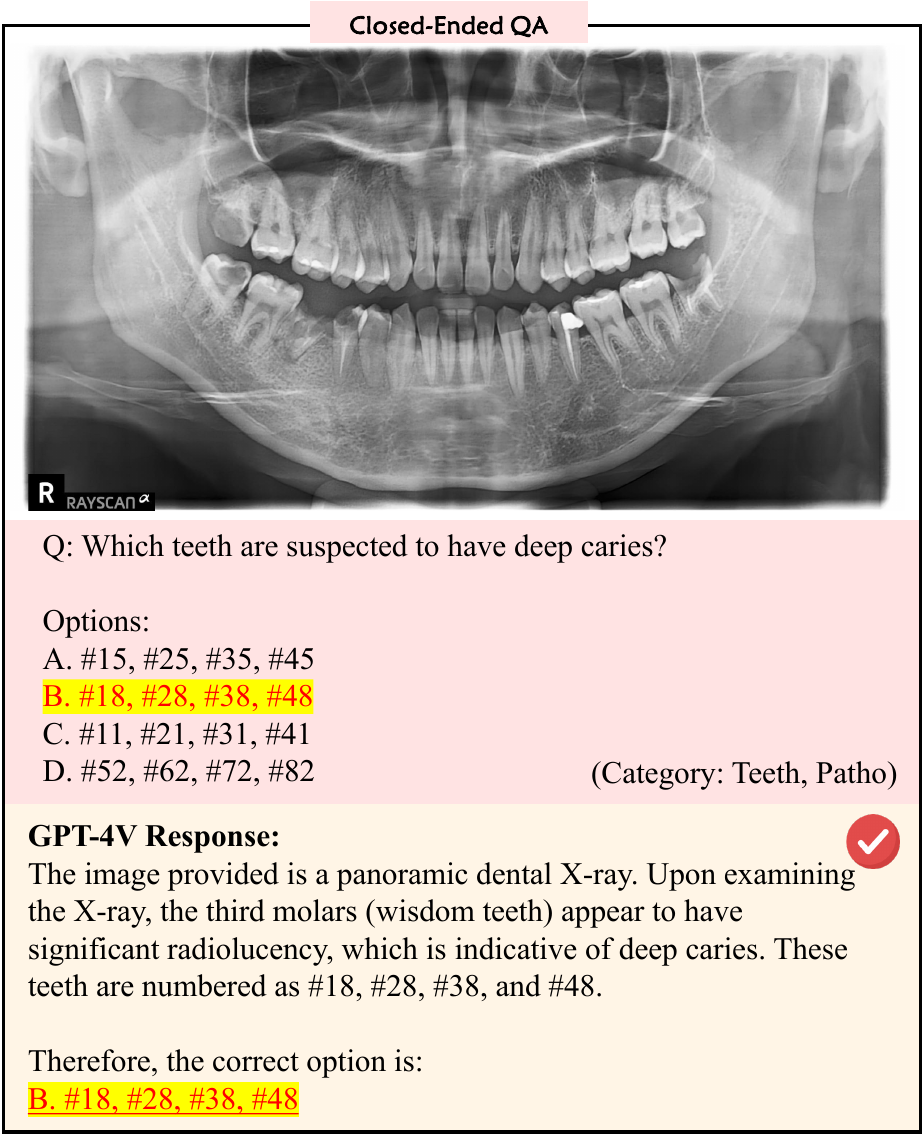}
  \caption{A closed-ended QA example. \colorbox{yellow}{\textcolor{red}{Red}} highlights the right answer. }
  \label{fig:mmoral_cases_close_1}
\end{figure}

\begin{figure}[!t]
  \centering
  \includegraphics[width=\textwidth]{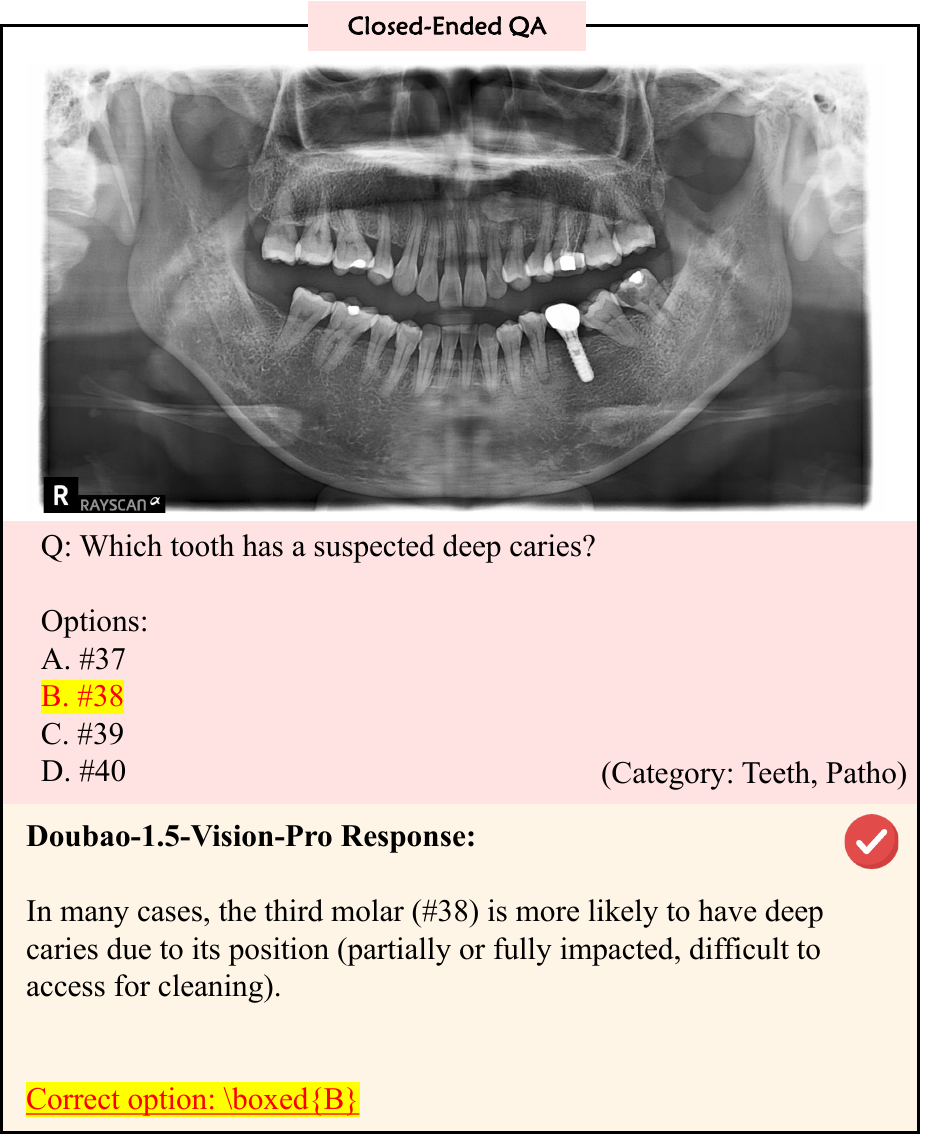}
  \caption{A closed-ended QA example. \colorbox{yellow}{\textcolor{red}{Red}} highlights the right answer. }
  \label{fig:mmoral_cases_close_3}
\end{figure}

\begin{figure}[!t]
  \centering
  \includegraphics[width=\textwidth]{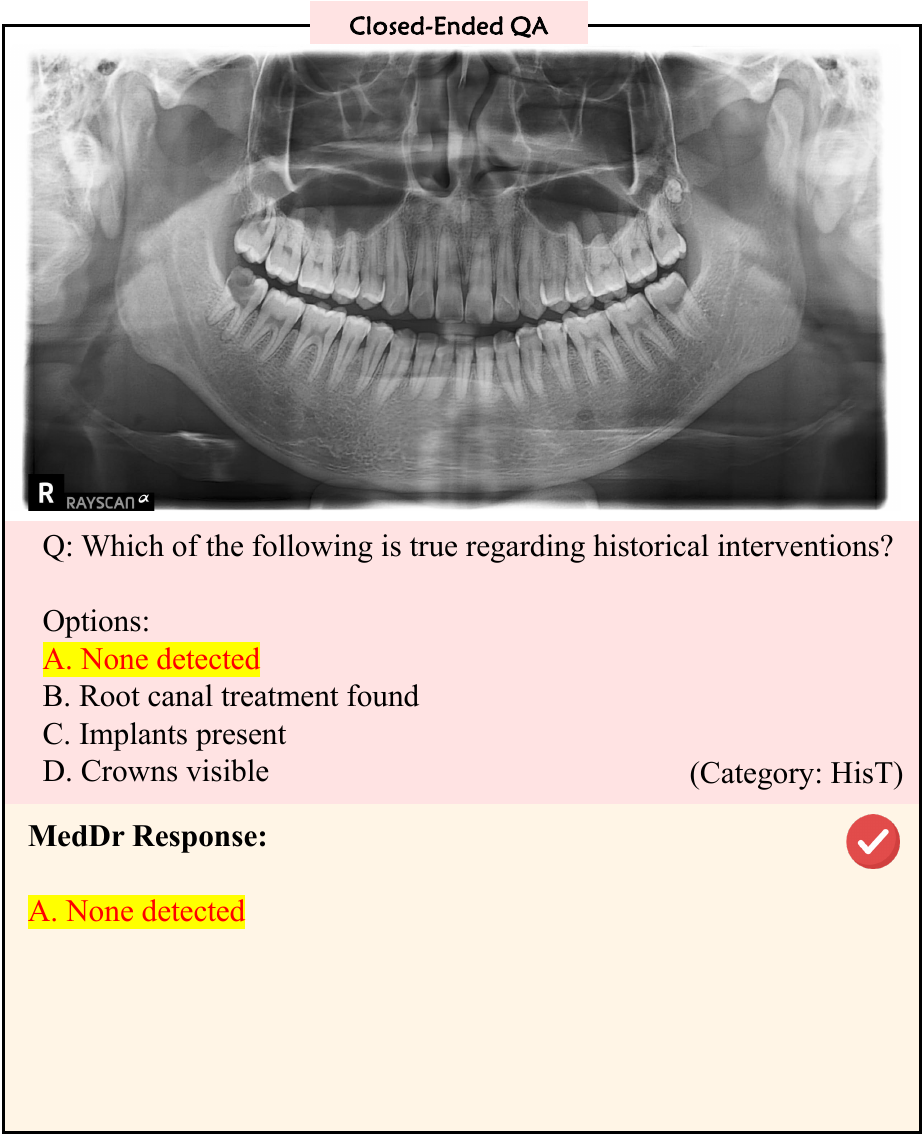}
  \caption{A closed-ended QA example. \colorbox{yellow}{\textcolor{red}{Red}} highlights the right answer. }
  \label{fig:mmoral_cases_close_5}
\end{figure}

\begin{figure}[!t]
  \centering
  \includegraphics[width=\textwidth]{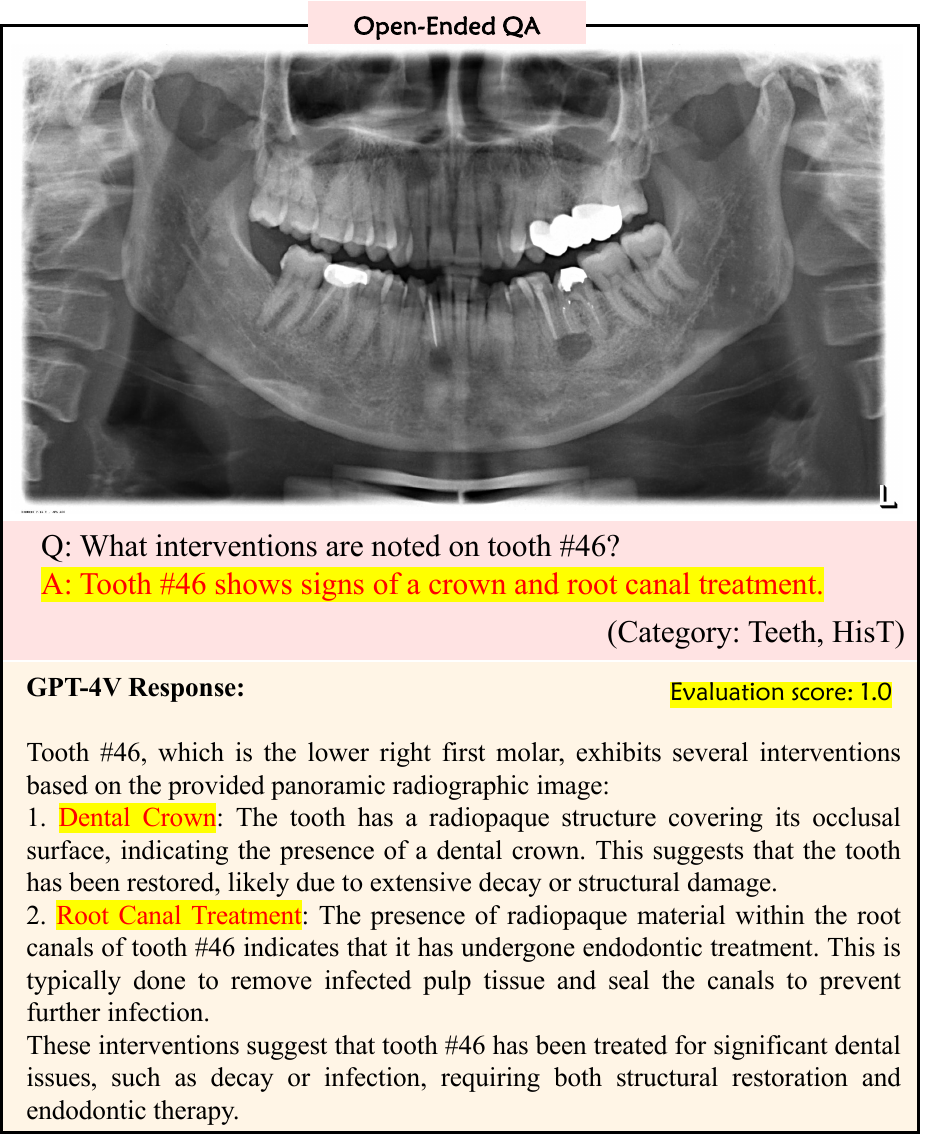}
  \caption{An open-ended QA example. \colorbox{yellow}{\textcolor{red}{Red}} highlights the right description.}
  \label{fig:mmoral_cases_open_1}
\end{figure}

\begin{figure}[!t]
  \centering
  \includegraphics[width=\textwidth]{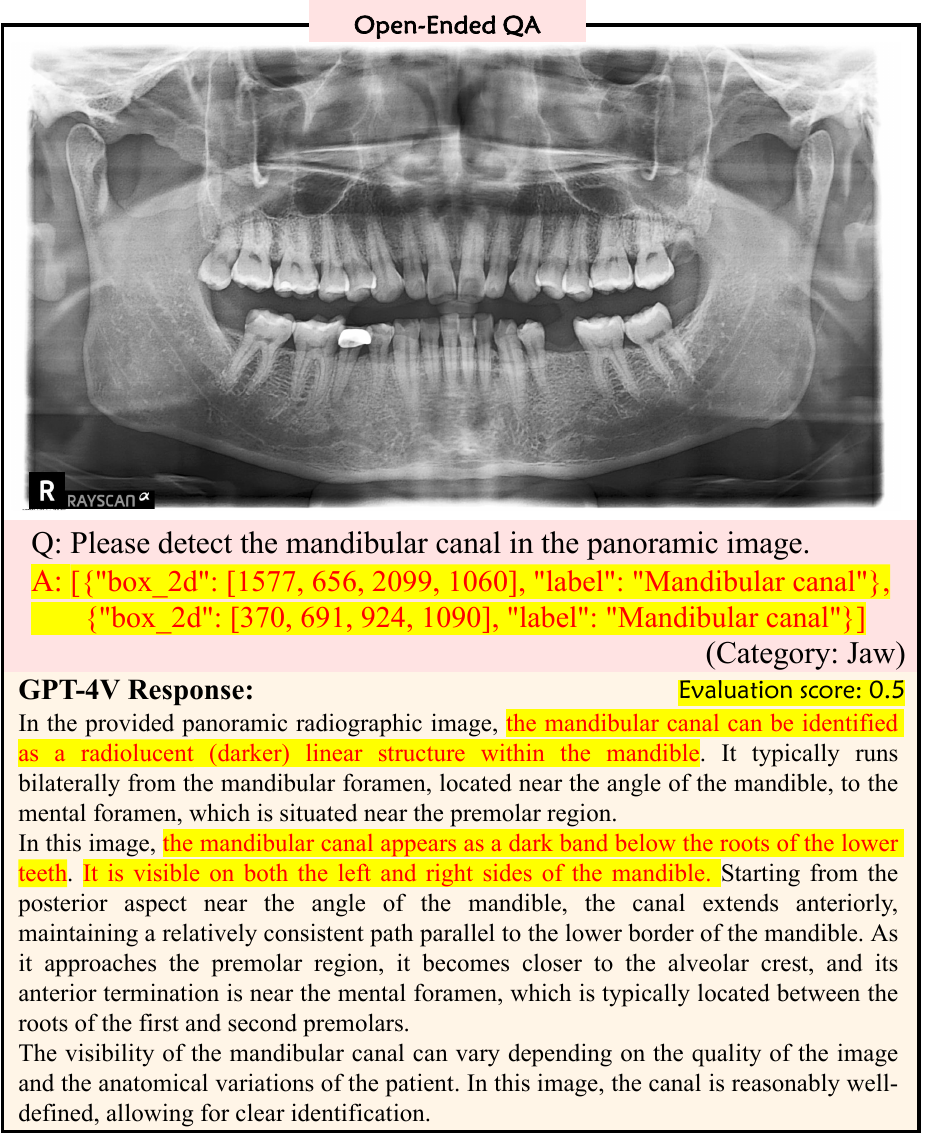}
  \caption{An open-ended QA example. \colorbox{yellow}{\textcolor{red}{Red}} highlights the right description.}
  \label{fig:mmoral_cases_open_3}
\end{figure}

\begin{figure}[!t]
  \centering
  \includegraphics[width=\textwidth]{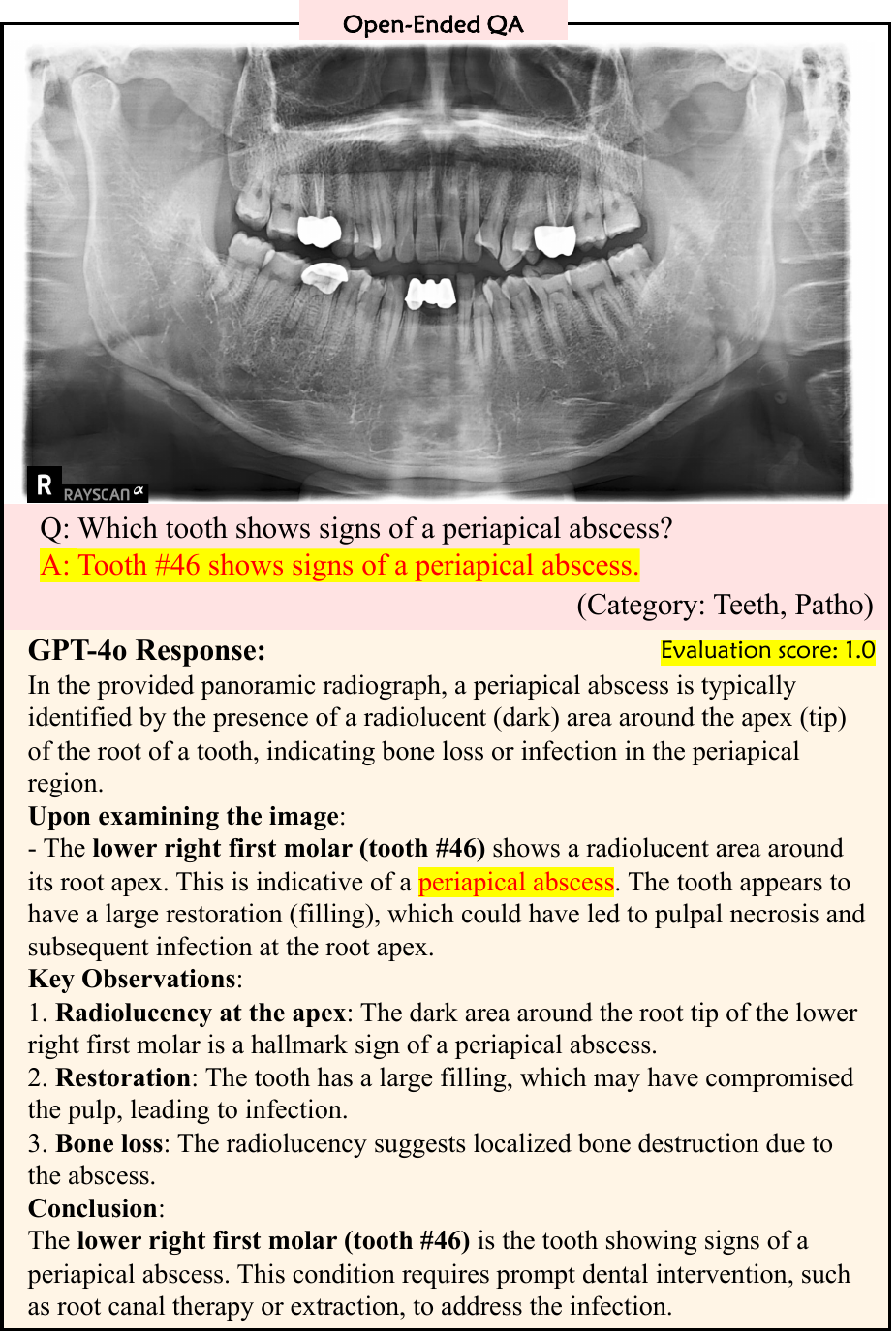}
  \caption{An open-ended QA example. \colorbox{yellow}{\textcolor{red}{Red}} highlights the right description.}
  \label{fig:mmoral_cases_open_4}
\end{figure}

In this section, we provide additional examples of the performance of various models on closed-ended and open-ended QA tasks. Figures~\ref{fig:mmoral_cases_close_1} -~\ref{fig:mmoral_cases_close_6} show examples of closed-ended QA, while Figures~\ref{fig:mmoral_cases_open_1} -~\ref{fig:mmoral_cases_open_4} show examples of open-ended QA.

\definecolor{lightcyan}{RGB}{150,255,255}

\begin{figure}[!t]
  \centering
  \includegraphics[width=\textwidth]{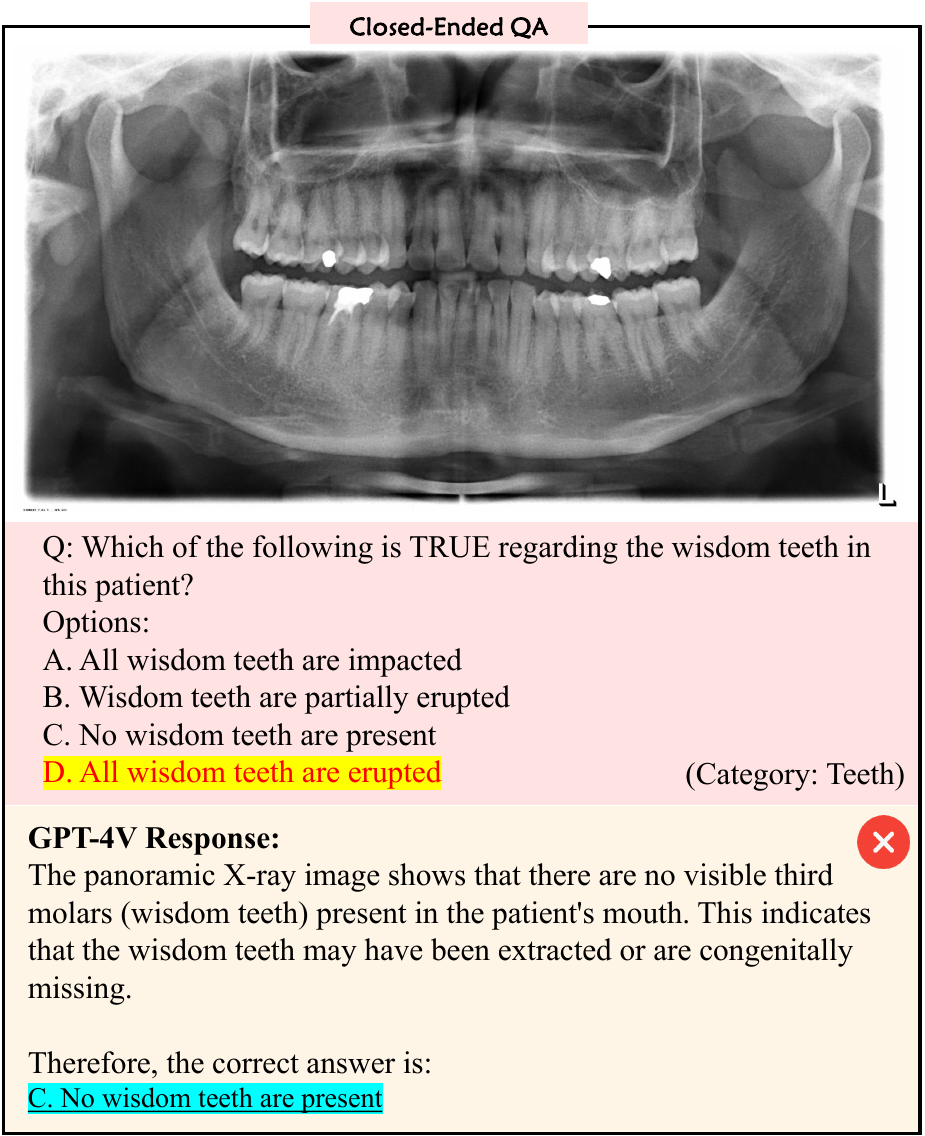}
  \caption{A closed-ended QA example. \colorbox{yellow}{\textcolor{red}{Red}} highlights the right answer. \colorbox{lightcyan}{\underline{Blue}} highlights the wrong answer.}
  \label{fig:mmoral_cases_close_2}
\end{figure}

\begin{figure}[!t]
  \centering
  \includegraphics[width=\textwidth]{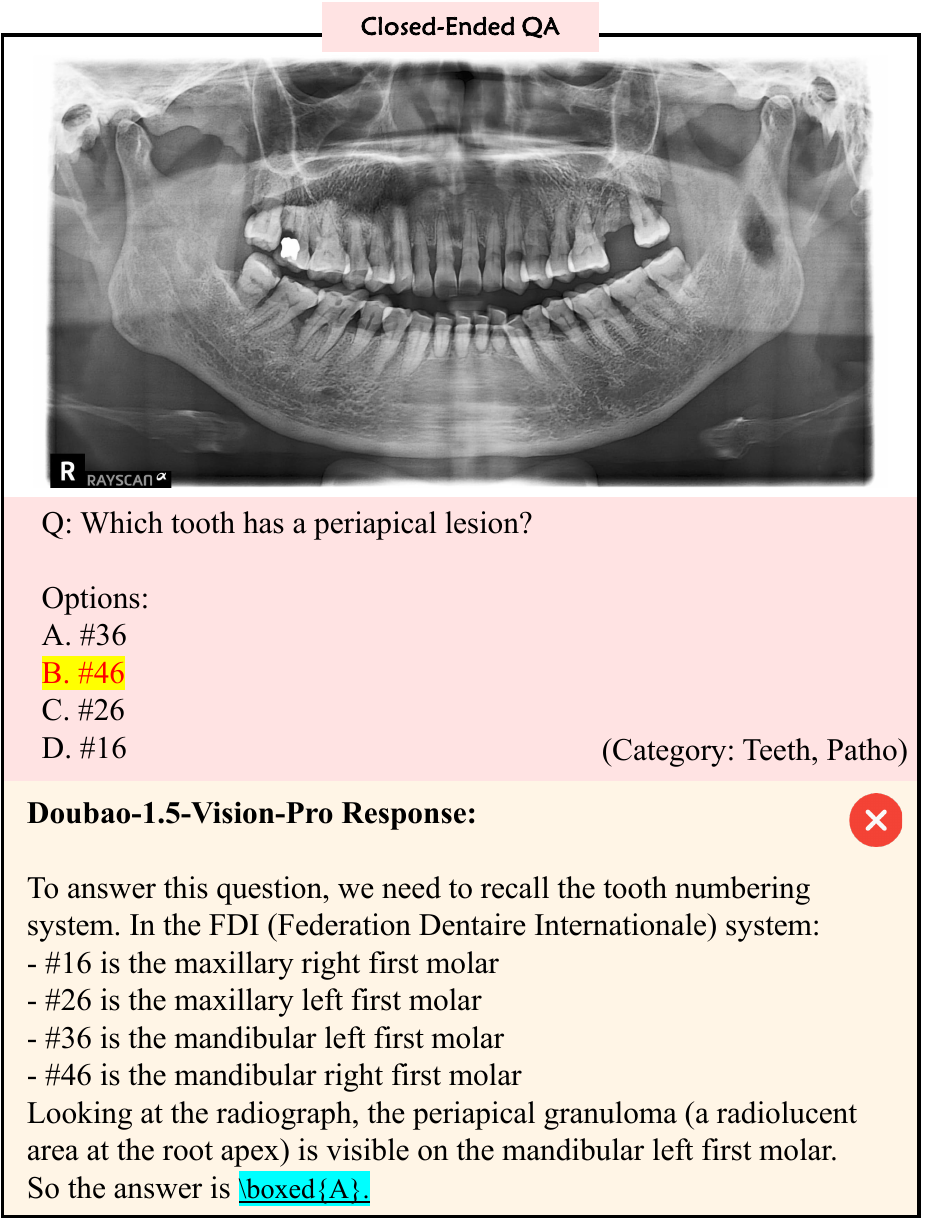}
  \caption{A closed-ended QA example. \colorbox{yellow}{\textcolor{red}{Red}} highlights the right answer. \colorbox{lightcyan}{\underline{Blue}} highlights the wrong answer.}
  \label{fig:mmoral_cases_close_4}
\end{figure}

\begin{figure}[!t]
  \centering
  \includegraphics[width=\textwidth]{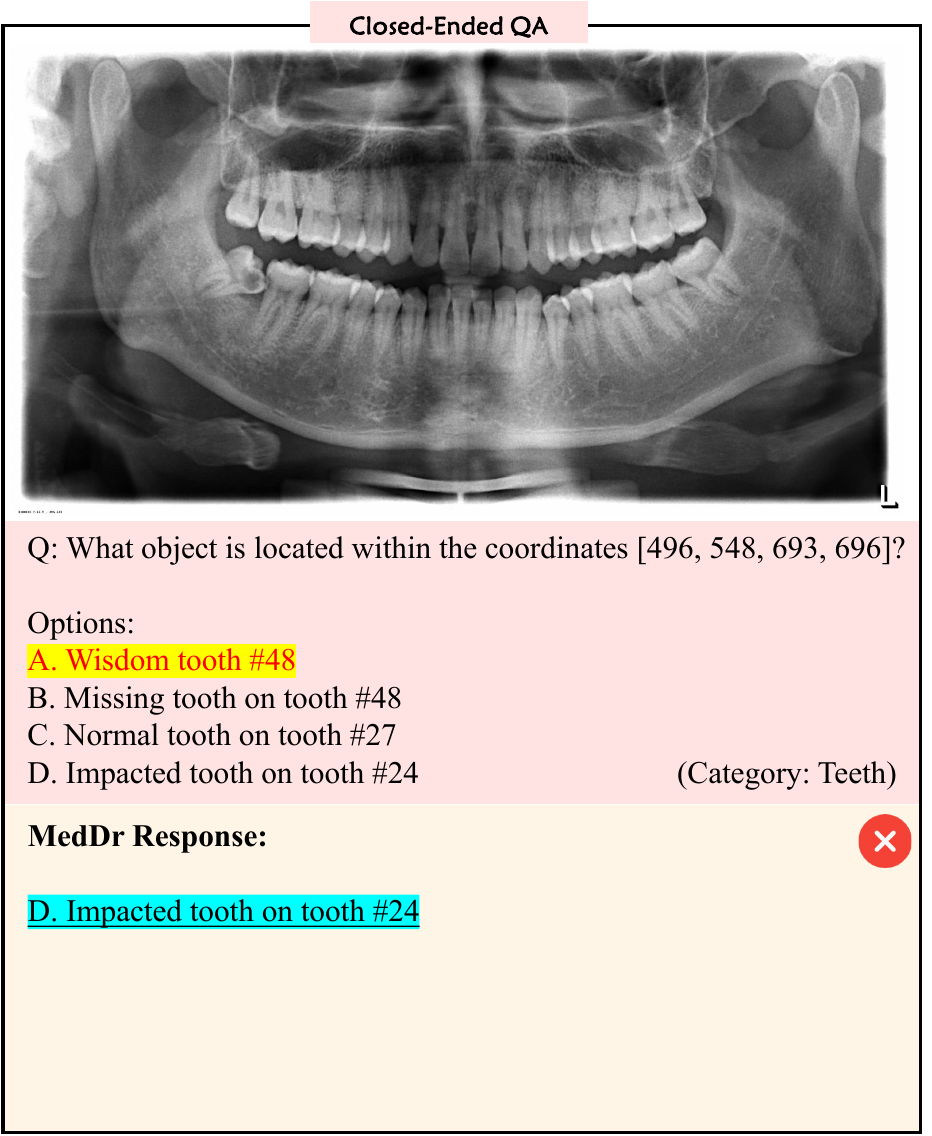}
  \caption{A closed-ended QA example. \colorbox{yellow}{\textcolor{red}{Red}} highlights the right answer. \colorbox{lightcyan}{\underline{Blue}} highlights the wrong answer.}
  \label{fig:mmoral_cases_close_6}
\end{figure}


\end{document}